\documentclass[journal, comsoc]{IEEEtran}

\usepackage[T1]{fontenc}
\usepackage{amsmath,amssymb}
\usepackage{graphicx}
\usepackage{enumitem}
\usepackage{multirow}

\usepackage{tabularx}
\usepackage{booktabs}
\usepackage{makecell}
\usepackage{comment}
\usepackage{microtype}
\usepackage{url}
\usepackage{cite}

\begin{document} 
 
\title{The Three Dimensions of ROS 2 Middleware}
\author{Sanghoon Lee, Taehun Kim, Angelo Corsaro and Kyung-Joon Park
\thanks{Sanghoon~Lee, Taehun~Kim, and Kyung-Joon~Park are with the Daegu Gyeongbuk Institute of Science and Technology (DGIST), Daegu 42988, South Korea. (e-mail: leesh2913@dgist.ac.kr, taehun@dgist.ac.kr, kjp@dgist.ac.kr).}
\thanks{Angelo Corsaro is with Eclipse Zenoh. (e-mail:ac@zenoh.io).}}

\maketitle

\begin{abstract} ROS~2 (Robot Operating System 2) has emerged as the de facto standard for modern robot software development, with middleware implementations such as the Data Distribution Service~(DDS) and Zenoh forming the core infrastructure for distributed robotic communication. 
Despite their architectural flexibility, these middleware systems exhibit structural limitations, particularly under dynamic and resource-constrained wireless environments. 
This paper presents a systematic survey of ROS~2 middleware and introduces a conceptual framework to examine its architectural limits through three structural dimensions required by distributed robotic systems, namely Space, Time, and State. 
We first provide a structured analysis of middleware architecture and operational dynamics, including discovery, data exchange, and state management mechanisms. 
Building on this foundation, we formalize Time as temporal predictability for control loops, Space as spatial abstraction from physical topology to enable modular deployment, and State as contextual continuity despite dynamic node participation and intermittent connectivity. 
Through a comprehensive review of existing implementations and prior studies, we organize middleware research according to the structural trade-offs that arise among these dimensions. 
Under constrained wireless conditions, spatial abstraction can obscure network variability and weaken temporal guarantees, while mechanisms that preserve state continuity introduce computational and network overhead that competes with time-critical communication. 
These interactions reveal structural trade-offs that characterize the practical limits of contemporary robot middleware. 
By synthesizing architectural patterns and identifying gaps in current modeling and analysis approaches, this survey outlines a principled research roadmap for robust and scalable robotic middleware architectures. 
\end{abstract} 

\begin{IEEEkeywords}
Robot Operating System~2~(ROS 2), middleware, Data Distribution Service~(DDS), Zenoh, Distributed robotic systems, Survey, Communication Architecture
\end{IEEEkeywords}

\section{Introduction} 
Robot Operating System 2 (ROS~2) has emerged as the de facto standard for modern robot software development~\cite{ros_metrics_clean,ros_metrics_2025}. 
The growing distribution and modularization of robotic systems has driven this transition, as perception, planning, and control components are deployed across multiple processes, devices, and network domains. 
As robotic platforms expand toward multi-robot coordination, edge-cloud integration, and wireless deployment, communication increasingly spans heterogeneous and resource-constrained environments~\cite{zhang2022distributed, fu2025iort, an2023multi}. 
Within ROS~2, the middleware layer abstracts and manages this complex communication through implementations such as Data Distribution Service~(DDS) and Zenoh~\cite{macenski2022robot}. 
These middleware systems constitute the core infrastructure for discovery, data exchange, and state management, thereby playing a central architectural role in determining system scalability, responsiveness, and robustness~\cite{cruz2012dds, al2025integrating}.

Distributed robotic systems simultaneously require temporal predictability for control loops, spatial abstraction from physical topology to enable modular and scalable deployment, and sustained state continuity despite dynamic node participation and intermittent connectivity. 
When these requirements are exercised in real deployments, distributed systems frequently exhibit performance degradation~\cite{castillo2024novel, agarwal2025scalable}. 
In wireless settings, bandwidth fluctuations and latency variability weaken temporal predictability~\cite{kong2025design, groshev2023edge, al2024ros}. 
In multi-node and multi-robot systems, dynamic topology changes delay or disrupt discovery and synchronization processes, undermining modular coordination across distributed components~\cite{naury2025communication}. 
Furthermore, mechanisms intended to preserve state continuity, such as buffering, replication, and session maintenance, introduce additional computational and network overhead that, in long-running deployments, often results in congestion or session instability~\cite{hietasalo2024overview}. \looseness=-1

A substantial body of research on ROS~2 has concentrated on application utilization and framework development, as well as performance optimization within robotic systems, with particular emphasis on real-time scheduling~\cite{casini2019response, teper2024end, tang2023real, tang2020response, dust2023dynamic}, executor design~\cite{liu2024rtex, staschulat2020rclc, yang2020exploring, enright2024paam, choi2021picas}, and latency control at the execution and transport layers~\cite{xu20223ds, blass2021automatic}. 
Quantitative surveys of ROS~2-specific publications indicate that only a very small fraction of the literature explicitly investigates middleware-layer design and analysis, accounting for approximately 3.5\% of reported studies~\cite{al2024ros}. 
Although interest in ROS~2 middleware has grown in recent years, particularly in areas such as Quality of Service~(QoS) policies, network behavior, and analytical modeling, it remains comparatively limited relative to the broader body of application-level and real-time research~\cite{ros_metrics_2025}. 
Furthermore, existing investigations typically proceed in isolation and lack a unifying architectural framework that systematically captures the cross-dimensional interactions among temporal predictability, spatial abstraction, and state continuity requirements~\cite{qi2025novel, baldoni2021facilitating}. \looseness=-1

These observations reveal that current middleware architectures increasingly constrain the performance envelope of distributed robotic systems. 
As robotics advances toward vision-language-action integration, large-scale multi-robot coordination, and persistent autonomous operation, communication and state management can no longer be treated as secondary concerns~\cite{kong2025design, an2023multi, lee2026harness}. 
The architectural properties of middleware directly influence system-level behavior under realistic deployment conditions, such as dynamic mesh networks and high-payload wireless scenarios. 
This survey therefore undertakes a systematic examination of middleware design to identify architectural gaps and establish a principled research roadmap for robust and scalable robotic middleware architectures.

The contributions of this paper are summarized as follows. 
\begin{itemize}[leftmargin=*] 
\item We provide a structured examination of ROS~2 middleware by analyzing its architectural constructs and operational dynamics, including discovery, data exchange, and state management mechanisms. 
\item Building on this analysis, we formalize three fundamental dimensions required by distributed robotic systems, namely \emph{Space}, \emph{Time}, and \emph{State}, and articulate their architectural implications for middleware design. 
\item We systematically reinterpret existing middleware research through the lens of these dimensions, organizing it according to the structural trade-offs among temporal predictability, spatial abstraction, and state continuity. 
\item Finally, we synthesize the identified structural gaps and propose a research roadmap to guide the development of robust and scalable robotic middleware architectures. \end{itemize}

The remainder of this paper is organized as follows. 
Section~\ref{sec:backgrounds} provides background on ROS~2 and its middleware layer. 
Section~\ref{sec:architecture} presents an architectural analysis of ROS~2 middleware constructs and their operational dynamics, including discovery, data exchange, and state management mechanisms. 
Section~\ref{sec:dimensions} formalizes the three fundamental dimensions of robot middleware, namely \emph{Space}, \emph{Time}, and \emph{State}, and discusses their realization in middleware implementations. 
Section~\ref{sec:tradeoffs} examines the structural trade-offs that arise among these dimensions and organizes related research within this framework.
Section~\ref{sec:roadmap} synthesizes the identified architectural gaps and outlines a research roadmap for robust and scalable robotic middleware design. 
Finally, Section~\ref{sec:conclusion} concludes the paper. \looseness=-1

\section{Backgrounds}
\label{sec:backgrounds}

\subsection{ROS~2} 
ROS~2 is an open-source software framework designed to support the development of distributed robotic systems. 
Unlike monolithic architectures, ROS~2 enables modular composition of perception, planning, and control components across multiple processes and networked devices~\cite{li2022autoware_perf, bonci2023robot}. 
Its primary objective is to provide a standardized software infrastructure that ensures scalability across heterogeneous deployment environments. 
At its core, ROS~2 utilizes a common functional base through the ROS Client Library~(RCL), which provides consistent semantics across language-specific implementations~\cite{ros2_internal_interfaces}. \looseness=-1

ROS 2 architects communication around the ROS Middleware~(RMW) interface, which serves as an abstraction and translation layer between the RCL and the underlying middleware implementation. 
This design enables various external implementations, such as eProsima Fast DDS, Eclipse Cyclone DDS, or Zenoh, to integrate seamlessly beneath a uniform application programming interface~\cite{ros2_diff_middleware_vendors, ros2_middleware_implementations}. 
Unlike ROS~1, which utilized a bespoke and centrally managed protocol, ROS 2 decouples high-level framework logic from low-level mechanisms of discovery, matching, and transport~\cite{ros2_changes_ros1_ros2}. 
ROS 2 translates application-level abstractions through the RMW interface into the primitives of the selected middleware implementation. 
This approach delegates connectivity and reliability management to the middleware while preserving a consistent programming model~\cite{ros2_creating_rmw_implementation}. \looseness=-1

At the application level, users define communication policies such as QoS parameters and callback execution semantics within the ROS 2 programming model. 
The RMW layer propagates these policies, which the underlying middleware implementation then realizes~\cite{ros2_creating_rmw_implementation}. 
The ROS 2 executor coordinates callback dispatch and intra-process scheduling, whereas the middleware handles the enforcement of communication reliability and transport semantics~\cite{ros2_executors}.
Consequently, temporal predictability and QoS behavior emerge from the interaction between application-level execution semantics and their realization within the middleware.

\vspace{-1em}
\subsection{ROS~2 Middleware}
Standard literature defines middleware as an abstraction layer positioned between the application layer and the underlying operating system~(OS), which mediates resource access in an OS-independent manner and provides service coordination across distributed components~\cite{bernstein1996middleware}. 
Among various categories of middleware, communication middleware provides services such as addressing, message routing, reliability control, QoS enforcement, and data serialization, thereby shielding applications from low-level socket programming and transport-specific details~\cite{emmerich2000software}. 
In ROS~2, the middleware implementation corresponds specifically to a communication middleware engine. 
It abstracts QoS policies expressed through the RCL API and interacts with the ROS~2 executor through the RMW WaitSet mechanism, which enables coordinated event notification and data handling across distributed components~\cite{macenski2022robot}. \looseness=-1

To avoid ambiguity, we use the term RMW to collectively denote the RMW interface layer and its corresponding middleware-specific implementation. 
We do not interpret RMW as a passive data transport layer. 
Its internal architectural choices directly influence latency characteristics, discovery convergence, bandwidth utilization, and state persistence across distributed deployments. 
To understand how these architectural choices manifest in practice, we next examine the primary middleware implementations supported in ROS~2. \looseness=-1

ROS~2 designates certain middleware implementations as Tier~one, which indicates the highest level of support and compatibility in its ecosystem~\cite{ros2_kilted_release}. 
These include DDS-based implementations such as eProsima Fast DDS, Eclipse Cyclone DDS, and RTI Connext DDS, as well as Eclipse Zenoh. 
In this paper, we focus on DDS implementations and Zenoh, because they represent the primary communication backends in current ROS~2 deployments. \looseness=-1

\subsubsection{Data Distribution Service (DDS)}
DDS is a widely adopted communication middleware standardized by the Object Management Group~(OMG) and originally developed for mission-critical applications in aerospace, defense, and autonomous systems~\cite{macenski2022robot}. 
DDS serves as the default middleware implementation in ROS~2 and operates based on a data-centric publish-subscribe paradigm in which information is exchanged through named topics. 
In DDS, each ROS~2 context typically creates a DomainParticipant to manage the relevant DDS entities, which establishes an entity-centric hierarchy. 
Explicit QoS policies shape communication behavior within this framework~\cite{pardo2003omg}.
Through these mechanisms, DDS provides configurable parameters for reliability, durability, and deadline, which the RMW interface layer exposes to ROS~2 users.
Of these, only the reliability policy enforces a delivery behavior, while others, such as deadline and transport\_priority, signal or hint rather than enforce. \looseness=-1

\subsubsection{Zenoh}
Zenoh is a communication middleware designed to enable efficient data exchange across distributed, heterogeneous, and resource-constrained environments \cite{zenoh_what_is_zenoh}.
It was developed to operate effectively under unreliable network conditions, including wireless and edge deployments \cite{ros2_kilted_release, ros2_rmw_report_2023}. 
Beginning with the ROS~2 Kilted Kaiju release in 2025, the community officially designated Zenoh as a Tier~one middleware implementation within the ROS~2 ecosystem. 
Architecturally, Zenoh employs a unified publish-subscribe and query protocol organized around a key/value data model, where hierarchical key expressions express interest and drive the resulting data flows.
In Zenoh, the middleware operates through Zenoh sessions, where the specific mode, such as peer, router, and client, determines how they join the network. 
This design follows a session-centric approach that offers significant flexibility in connectivity and topology~\cite{corsaro2023zenoh, zenoh_abstractions}. 
In ROS~2 deployments, users commonly employ a Zenoh router, which facilitates discovery and data forwarding between sessions~\cite{rmw_zenoh_github}. \looseness=-1

\vspace{0.5em}
\subsubsection{Specialized Middleware}
In addition to DDS and Zenoh, the ROS~2 ecosystem includes other middleware-related technologies with specialized scopes. 
For example, iceoryx provides zero-copy shared memory transport to optimize intra-host data exchange~\cite{pohnl2023shared}, while micro-ROS extends ROS~2 to resource-constrained microcontroller environments with limited networking capabilities through an XRCE-DDS-based client-agent architecture~\cite{stadnik2024overview}. 
As these technologies target specific deployment contexts rather than general-purpose distributed communication across networked hosts, this study focuses primarily on DDS and Zenoh. \looseness=-1

\section{ROS~2 Middleware Architecture and Dynamics}
\label{sec:architecture}

\subsection{ROS~2 Middleware Architecture} 
At the application level, ROS~2 software is organized into packages as the fundamental deployment units~\cite{parmar2020syntactic}. 
Each package typically contains executable programs that instantiate one or more nodes at runtime to enable distributed computation. 
A node represents an independent execution entity within a running executable that encapsulates application logic and communicates through standardized communication abstractions~\cite{casini2025survey}. 
Rather than interacting directly with network primitives, nodes create publishers, subscriptions, services, and actions through the RCL. 
The RMW interface layer then maps these abstractions into middleware-level communication constructs to enable typed data exchange~\cite{peng2022scheduling, kouril2024performance}.

ROS~2 provides three communication abstractions at the RCL layer: Topics, Services, and Actions~\cite{ros2_topics_services_actions}. 
Topics implement an asynchronous publish-subscribe communication mechanism, where a publisher sends messages to one or more subscribers without waiting for a request-response exchange.~\cite{ros2_understanding_topics}. 
Services implement synchronous request-response interactions, where a client sends a request and remains blocked until a service server returns a response~\cite{ros2_understanding_services}. 
Actions extend the request-response pattern, where a client sends a goal to an action server that manages execution, publishes feedback, and returns a result. 
This mechanism is internally implemented through a combination of topics and services~\cite{ros2_understanding_actions}. 
All data types are defined in ROS 2 interface files, from which language-specific code is generated at build time and then used in the RCL layer~\cite{ros2_interfaces}
The RMW interface layer does not interpret these abstractions semantically, but maps each RCL entity to middleware-level entities, which operate within specific process and host boundaries~\cite{ros2_internal_interfaces}. \looseness=-1

\begin{figure*}[ht]
\centering
\includegraphics[width=\textwidth]{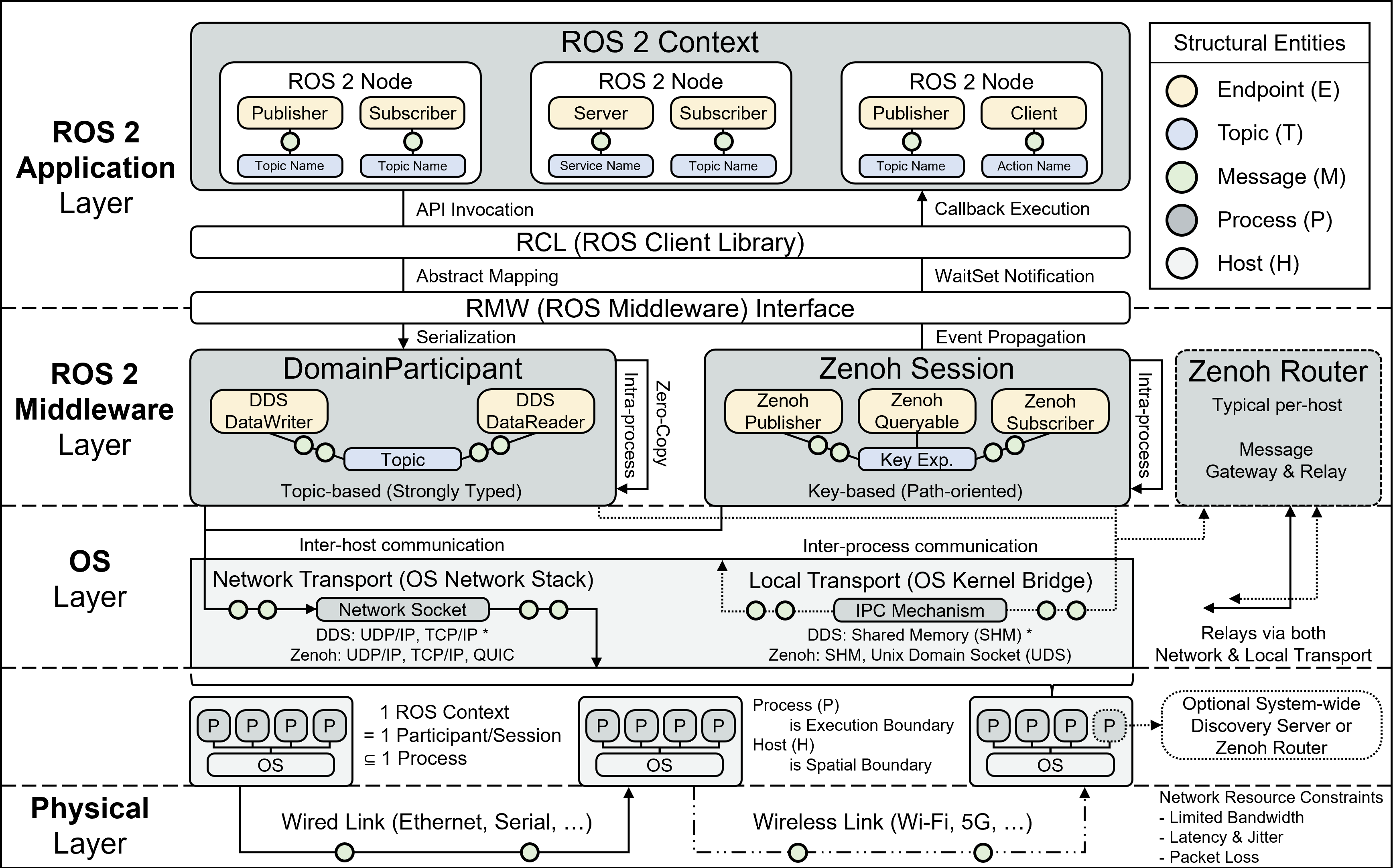}
\caption{Structural abstraction of the ROS~2 communication architecture. For DDS, TCP/IP and SHM (marked with *) are vendor-specific extensions.}
\label{fig:structure}
\vspace{-1em}
\end{figure*}

To provide a unified view across middleware implementations, we abstract the ROS~2 communication architecture into five structural entities: Endpoint~(E), Topic~(T), Message~(M), Process~(P), and Host~(H). 
Although DDS and Zenoh differ in their internal implementation, these entities can be consistently identified across deployment environments.
\begin{itemize}[leftmargin=*]
\item \textbf{Endpoint~(E):} 
An endpoint is the fundamental unit of data production or consumption in the middleware layer. 
At the RCL layer, endpoints correspond to publishers, subscriptions, service clients, and service servers, which the RMW interface layer maps to concrete middleware constructs such as DDS DataWriter/DataReader entities or Zenoh publishers, subscribers, queriers, and queryables.
Through this mapping, endpoints function as event-generating entities within the middleware layer, triggering RCL callbacks upon entity readiness propagated via the WaitSet in the RMW layer and thereby enabling reactive execution in ROS~2 applications. 
Each endpoint is instantiated within a specific OS process and assigned a unique identifier within the middleware domain. 
The middleware forms logical associations between endpoints based on matching topic names and type definitions. 
Each endpoint operates under QoS policies that govern reliability, durability, and delivery semantics, and maintains internal state as required to realize these QoS guarantees.

\item \textbf{Topic~(T):} 
A topic is a logical communication channel identified by a name and an associated data type~\cite{krinkin2018data}, explicitly formed at the RCL layer. 
The RMW layer maps each topic to middleware-specific constructs, such as topic names in DDS or hierarchical key expressions in Zenoh.
Topics provide the naming and typing context required for endpoint matching across distributed contexts, thereby enabling communication between matching endpoints across process and host boundaries.

\item \textbf{Message~(M):} 
A message is a concrete data instance conforming to a type defined in ROS~2 interface. 
At the RCL layer, message structures are generated from interface definitions and instantiated by application logic. 
At the RMW interface, message instances are serialized into middleware-specific representations for transmission. 
In DDS, this representation corresponds to a serialized sample, whereas in Zenoh it corresponds to a value payload associated with a key expression. 
Messages are exchanged between matched endpoints within the scope defined by a topic and may be delivered across process and host boundaries via middleware-managed transport mechanisms. 
Endpoint-level QoS policies govern the lifetime and delivery behavior of message instances. 

\item \textbf{Process~(P):} 
A process is an OS-level execution unit that serves as the direct counterpart to a participant: each process corresponds to one middleware participant, realized as a DDS DomainParticipant or a Zenoh session. 
Within a process, one or more ROS~2 nodes are instantiated and share a single ROS~2 context, which is mapped to a single participant. 
While most deployments run a single node per process, the node composition mechanism allows multiple nodes to be loaded into the same process, sharing a ROS~2 context and therefore a single participant. 
Each participant acts as the entry point through which nodes and their endpoints participate in middleware communication, managing the lifecycle of publishers, subscriptions, service clients, and service servers declared within it. 
Processes on the same host share hardware resources but are allocated distinct OS resources, including network ports and socket descriptors, thereby defining the execution boundary within which middleware participation is maintained and callbacks are scheduled. 
In advanced configurations, a process may maintain multiple contexts, each mapped to an independent participant, to support scenarios such as domain bridging or multi-domain operation.

\item \textbf{Host (H):} 
A host is a distinct computing unit equipped with one or more network interfaces.
It provides the physical execution environment within which one or more processes are deployed. 
Communication between processes on the same host may occur via shared memory or local transport mechanisms, bypassing network contention. 
In contrast, communication across hosts relies on network interfaces and shared communication media. 
Although a host is often associated with a single robot, complex robotic platforms commonly comprise multiple networked hosts interconnected through heterogeneous communication links, including Ethernet, wireless, and serial buses~\cite{blancoenhancing, jiang2020message}.
The host, therefore, defines the spatial boundary across which middleware communication is routed and over which network-level resource constraints are manifested. \looseness=-1
\end{itemize}

Middleware implementations can introduce additional components for routing and discovery. 
For example, Zenoh may employ one or more Zenoh routers, which can be deployed per host or as shared infrastructure within a local network to facilitate discovery and data forwarding. 
Similarly, certain DDS deployments may utilize discovery servers or routing services, instantiated as dedicated processes on specific hosts~\cite{eprosima_discovery_server, rti_routing_service}. 
These components introduce intermediary processes that alter the communication topology and influence network behavior, yet they do not redefine the fundamental structural entities introduced above.
Instead, they mediate how processes and hosts are interconnected within a given deployment. \enlargethispage{\baselineskip}

Fig.~\ref{fig:structure} illustrates the unified structural abstraction of the ROS~2 communication architecture, organized into a four-tier hierarchy of application, middleware, OS, and physical layers. 
Within the application layer, nodes instantiate endpoints that exchange messages over topics.
These entities are hosted within a ROS~2 context mapped to a single participant in the middleware layer.
Two fundamental boundaries govern the system. 
The process defines the execution boundary within which participation is maintained and callbacks are scheduled, while the host defines the spatial boundary across which processes communicate via local or network transport, including Ethernet, wireless, and serial buses.
Together, these structural entities provide a consistent foundation for connecting heterogeneous distributed robotic systems across both high-level abstractions and physical hardware constraints. 

\subsection{Middleware Operational Dynamics} 
In practice, middleware communication is governed by three interrelated operational dynamics, namely discovery, data exchange, and state management, that operate over these structural entities during execution. 
Discovery mechanisms determine how participants and endpoints are discovered and logically interconnected across processes and hosts. 
Data exchange mechanisms regulate how messages are serialized, transmitted, and received under reliability and QoS constraints. 
State management mechanisms maintain and evolve the internal middleware state associated with these entities to preserve continuity under dynamic participation and long-running deployment. 
These dynamics are continuously active rather than sequential stages, and their interactions shape the performance, scalability, and robustness of distributed robotic systems. 
The following subsections examine how DDS and Zenoh realize these mechanisms and analyze the structural implications of their respective design choices. 

\vspace{0.5em} 
\subsubsection{Discovery} 
Discovery mechanisms establish the initial and ongoing spatial structure of a distributed ROS~2 system. 
Through discovery, middleware participants become aware of one another, exchange metadata, and construct the logical graph over which communication subsequently occurs. 
Unlike a one-time initialization step, discovery is continuously active, reacting to dynamic participation, restarts, and network reconfigurations~\cite{putra2015node}. 
The discovery process comprises (i) participant discovery and (ii) endpoint matching.

\vspace{0.5em}
\textit{(i) Participant Discovery:}
Participant discovery refers to the mechanism by which DDS DomainParticipants or Zenoh sessions become mutually aware of one another. 
In DDS, participant discovery is, by default, performed using the Real-Time Publish-Subscribe~(RTPS) protocol~\cite{omg2022rtps}. 
More specifically, DDS employs the Simple Participant Discovery Protocol~(SPDP), a distributed peer-to-peer~(P2P) mechanism in which DomainParticipants periodically announce their presence within a domain. 
These announcements allow other participants to detect newly joined participants and establish mutual awareness. 
By default, this discovery behavior relies on multicast communication within the local network scope, enabling direct P2P visibility without requiring centralized coordination~\cite{maggi2022security}.

In Zenoh, participant discovery can be implemented through multiple independently configurable mechanisms.
One mechanism is multicast scouting, in which a session announces its presence on the local network so that nearby sessions can detect it directly. 
Another mechanism is gossip-based discovery, in which a session propagates discovery information about already known sessions and routers to other connected sessions~\cite{zanni2024efficient}. 
These mechanisms are not inherently coupled and may be enabled or disabled independently. 
In the default configuration of a Zenoh-based RMW interface, multicast scouting is disabled while gossip-based discovery is enabled. 
In this setting, Zenoh routers propagate discovery information received from locally attached sessions and redistribute it to other routers and sessions through gossip messages~\cite{caruso2024autonomous}. \looseness=-1

\textit{(ii) Endpoint Matching:}
Following participant discovery, middleware systems advertise endpoint metadata and determine compatibility within the established communication context. 
In DDS, endpoint matching is performed through the Simple Endpoint Discovery Protocol~(SEDP), which disseminates endpoint metadata via built-in discovery topics~\cite{michaud2018attacking}. 
These topics convey information such as topic names, data types, and QoS policies. 
Matching occurs when DomainParticipants detect compatible name--type pairs and verify that their QoS policies satisfy the DDS compatibility rules. 
Only after this compatibility check do the corresponding entities establish a logical association.

In Zenoh, endpoint advertisement is performed by announcing resource key expressions and subscription interests to directly connected sessions and routers.
Matching occurs when a published key expression satisfies a corresponding subscription interest~\cite{baron2025performance}.
Unlike DDS, Zenoh does not perform endpoint-level QoS compatibility checks during the matching phase; instead, QoS settings influence behavior at the data exchange stage. 
In router-mediated configurations, a router collects the key expressions and subscription interests of its connected sessions and shares them with other routers or with sessions not directly connected to one another. 
Based on this aggregated state, the router constructs and maintains visibility relationships among its connected sessions, thereby enabling endpoint matching among otherwise isolated sessions.

\vspace{0.5em} 
\subsubsection{Data Exchange} 
Data exchange mechanisms determine how message instances are represented, routed, and delivered across processes or hosts under the underlying transports, routing topology, and required QoS. 
Data exchange governs the runtime behavior of communication after endpoint matching, shaping the latency, jitter, and loss characteristics observed by ROS~2 applications. 
In ROS~2, these behaviors emerge from the interaction among (i) serialization and encoding, (ii) transport mechanisms, and (iii) reliability and delivery control.

\vspace{0.5em}
\textit{(i) Serialization and Encoding:}
Before transmission, ROS~2 message instances are serialized into middleware-specific wire representations. ROS~2 adopts Common Data Representation~(CDR)~\cite{tijero2012schedulability} as its standard serialization format at the RCL layer, applied consistently regardless of the underlying middleware implementation.

In DDS, the serialized output constitutes a data sample, which serves as the atomic unit for transmission and caching~\cite{scordino2022hardware}. 
DDS may attach inline metadata that carries RTPS-level information and selected QoS-related parameters required for interpreting or managing the sample at the receiver side. 
The resulting representation combines the application payload and protocol metadata into a form suitable for RTPS-level encapsulation and transport. 
In Zenoh, the serialized output is conveyed as a byte payload associated with a key expression~\cite{zanni2024efficient}. 
The Zenoh wire protocol encapsulates the payload with protocol-level metadata, including the wire key expression, encoding information, and optional timestamp.
Application-level metadata may additionally be attached on a per-message basis as key-value pairs via the attachment API, without requiring re-serialization of the payload.

\textit{(ii) Transport Mechanisms:}
Following serialization, the middleware selects transport mechanisms to deliver payloads across process and host boundaries. 
Both DDS and Zenoh support shared-memory transport for intra-host communication, avoiding kernel-network overhead for co-located processes.
DDS relies on the RTPS protocol, which defines UDP-based unicast and multicast as its primary transport. 
Modern DDS implementations support TCP to accommodate network environments where UDP multicast is restricted or unreliable.

Zenoh employs a pluggable transport layer that supports multiple protocols, including TCP, UDP, and QUIC~\cite{liang2023performance}. 
Communication can proceed over direct P2P links or through router-mediated forwarding, depending on deployment configuration. 
Zenoh routers relay data based on link-state information and network visibility, enabling operation across heterogeneous network topologies~\cite{desbiens2021zenoh}. 

\textit{(iii) Reliability and Delivery Control:}
To ensure delivery semantics under loss and contention, middleware can maintain protocol state and apply reliability, congestion, and flow control mechanisms. 
In DDS, reliable communication is typically realized through stateful RTPS-level mechanisms, such as AckNack exchanges, periodic Heartbeats, and retransmission windows maintained per endpoint~\cite{hakiri2013supporting}. 
These mechanisms operate in conjunction with reliability and history QoS policies, which define send-queue depth and the persistence of unacknowledged samples.
Per-endpoint cache limits and timing parameters that bound Heartbeat and retransmission rates further govern congestion and flow behavior. 

In Zenoh, each session provides a best-effort channel and a reliable channel by default, with the reliable channel maintaining sequence numbers and acknowledgment state for loss detection and retransmission~\cite{corsaro2021reliability}. 
When connection-oriented transports such as TCP or QUIC are used, their native ordering and loss recovery supplement these session-level guarantees~\cite{fu2025iort}. 
Publishers control congestion behavior by selecting a per-message drop or block policy applied along the entire routing path, including Zenoh routers~\cite{corsaro2021reliability}.

\begin{figure*}[t]
\centering
\includegraphics[width=\textwidth]{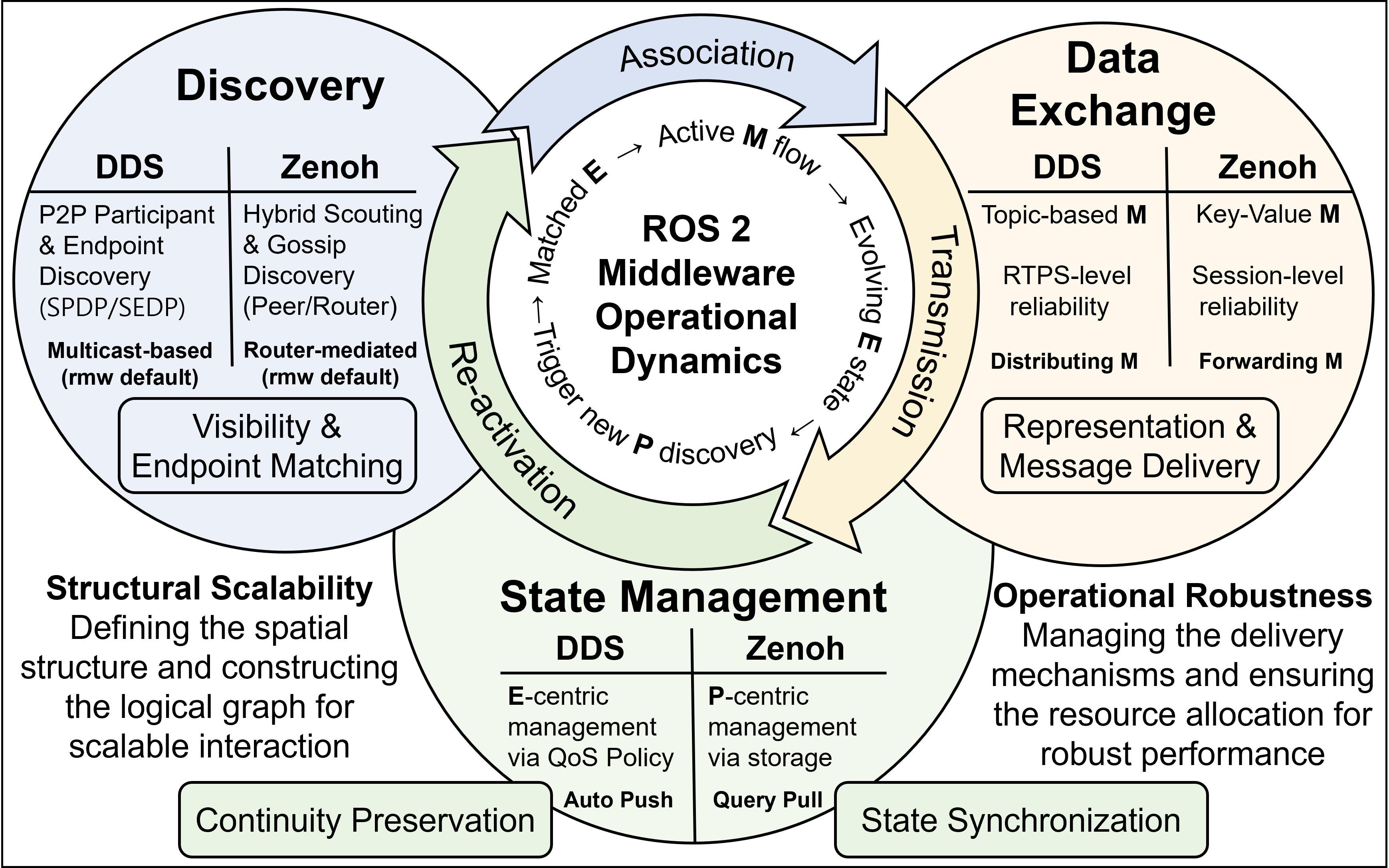}
\caption{ROS~2 Middleware Operational Dynamics.}
\label{fig:dynamics}
\vspace{-0.5em}
\end{figure*}

\vspace{0.5em} 
\subsubsection{State Management} 
State management mechanisms determine how middleware maintains communication state and preserves continuity across process and host boundaries under dynamic conditions and intermittent connectivity. 
State management governs what information is tracked, cached, retained, and recovered. 
In ROS~2, these behaviors emerge from the interaction among three aspects: (i) liveliness maintenance, (ii) history restoration, and (iii) resource control.
\vspace{0.5em}

\textit{(i) Liveliness Maintenance:}
Liveliness maintenance determines how middleware monitors the presence and health of contexts across processes and hosts.
In DDS, liveliness is maintained through Heartbeat and lease mechanisms at both the participant and endpoint levels. 
These mechanisms are continuously active, with periodic liveliness assertions refreshing leases while missed assertions trigger liveliness-loss detection. 
This creates a decentralized monitoring structure in which each DomainParticipant observes the liveliness of others. \looseness=-1

In Zenoh, liveliness is governed by a session lease and keep-alive mechanism~\cite{baron2025performance}.
Any data or protocol message exchanged on a session refreshes the lease, and keep-alive messages are sent only when the session would otherwise be idle.
If the lease expires, the session is closed and the associated resources are removed. 
Zenoh also provides liveliness tokens for application-level monitoring~\cite{zettascale2023charmander}. 
These are explicitly declared resources whose presence is observed by peers, and whose loss is signaled when the declaring session is disconnected or the token is undeclared~\cite{zenoh_rust_liveliness}. 
This design provides session-centric monitoring and avoids continuous P2P polling.

\textit{(ii) History Restoration:}
Middleware governs the synchronization of state for entities that rejoin after disruption or appear after messages have already been produced. 
In DDS, this behavior is localized within the HistoryCache of an endpoint~\cite{fernandez2020performance}. 
Under durability QoS settings, a DataWriter replays messages from its runtime cache to newly matched DataReaders~\cite{barcis2020information}. 
Consequently, in standard ROS~2 deployments, history restoration is limited to data retained in active memory and does not persist across entity restarts without specialized persistent storage services~\cite{pardo2005introduction}.

Zenoh decouples history from the producing endpoint by introducing independent storage managers~\cite{zettascale_zenoh_storage}. 
These managers act as dedicated peers that store values for specific key expressions, enabling late-joining endpoints to pull historical state via query mechanisms even after the original endpoint is no longer available~\cite{zanni2024efficient}. 
This design complements endpoint-local buffering by providing a process-level storage service.

\textit{(iii) Resource Control:}
Resource control mechanisms define the constraints and policies that prevent unbounded memory consumption and buffer exhaustion within the middleware. 
In DDS, these constraints are enforced through the resource limits QoS policy, which bounds HistoryCaches at the individual endpoint level~\cite{corsaro2014data}. 
By defining parameters such as the maximum number of stored samples per endpoint, DDS provides explicit control over the memory allocated to each entity. 
However, the resulting behavior depends on the preconfigured limits, and insufficient limits may cause message drops or blocking.

Zenoh manages resources through transmission queues configurable on sessions and routers~\cite{zenoh_rust_session}. 
Users can explicitly control queue sizes and the memory allocated to each queue. 
This makes resource management more path-based than endpoint-based, because available buffering depends on configured queues along the routing path~\cite{zettascale_zenoh_deployment}. 
Compared with DDS, Zenoh distributes resource control across sessions and routers rather than enforcing it at the individual endpoint level.

\vspace{0.5em} 
Fig.~\ref{fig:dynamics} illustrates a visual synthesis of the three primary operational dynamics in ROS~2 middleware: discovery, data exchange, and state management. 
These dynamics function as interdependent loops to collectively govern the distributed system. 
The spatial structure established by discovery directly shapes the routing paths available for data exchange, while the internal state managed by the middleware ensures the continuity of these interactions. 
This tight coupling implies that any modification to one dynamic inevitably propagates performance implications to the temporal and state-related aspects of the communication stack.

To systematically evaluate these inherent trade-offs, Section~\ref{sec:dimensions} formalizes the three structural dimensions required by ROS~2 middleware: \emph{Space}, \emph{Time}, and \emph{State}. 
We transition to a unified framework that defines the objectives pursued by ROS~2 middleware within its operational dynamics, grounding each dimension in the previously defined structural entities.

\section{Three Dimensions of Robot Middleware} 
\label{sec:dimensions} 
This section formalizes the three structural dimensions essential to distributed robotic systems: \emph{Space}, \emph{Time}, and \emph{State}.
These dimensions are not design choices unique to a particular middleware; they are universal structural requirements imposed by the nature of distributed robotic systems. 
ROS~2 middleware must provide (i)~location transparency so that applications can interact without knowledge of physical placement, (ii)~temporal predictability so that time-sensitive robotic tasks receive consistent and bounded communication behavior, and (iii)~contextual continuity so that system state is preserved across dynamic node participation. 
DDS and Zenoh pursue identical objectives along each of these dimensions, even though their mechanisms for realizing them differ structurally. 
This section formalizes each dimension as a shared objective and establishes how it maps to the operational dynamics and structural entities defined in the preceding section. \looseness=-1

Table~\ref{tab:dimensions} summarizes this mapping by associating each dimension with a key question that captures its operational concern and a primary objective that defines the goal middleware must pursue to satisfy it.
The key question for each dimension is not merely descriptive but diagnostic: it identifies the class of failure that arises when the dimension is insufficiently realized.
When \emph{Space} is inadequately supported, the system cannot answer where an entity is located, and modular deployment breaks down.
When \emph{Time} is inadequately supported, the system cannot guarantee when a message is delivered, and control loops become unstable.
When \emph{State} is inadequately supported, the system cannot preserve how state is maintained across participation changes, and contextual continuity is lost.
By mapping these dimensions to specific key questions and primary objectives, we establish a unified framework for evaluating how middleware supports the complex requirements of modern robotics. \looseness=-1
\begin{table}[ht]
\centering
\caption{Structural Dimensions of Robot Middleware}
\label{tab:dimensions}
\begin{tabularx}{\columnwidth}{l X l}
  \hline
  \textbf{Dimension} & \textbf{Key Question} & \textbf{Primary Objective} \\ \hline
  \textbf{Space} & Where is the entity located? & Location Transparency \\
  \textbf{Time} & When is the message delivered? & Temporal Predictability \\
  \textbf{State} & How is state preserved? & Contextual Continuity \\ \hline
\end{tabularx}
\vspace{-1em}
\end{table}

\subsection{Space: Location Transparency} 
\emph{Space} denotes the structural objective of maintaining a coherent logical communication structure that remains independent of the underlying network and deployment topology. 
The core requirement is location transparency, meaning that applications must be able to express interaction through data identifiers and semantic roles rather than physical network addresses or fixed placements. 
Without this property, system components cannot be composed, relocated, or scaled across processes and hosts, as doing so would require rewriting the application-level logic.
This paper treats location transparency at the architectural level, requiring the middleware to sustain a coherent communication structure without operator-specified routing, intermediary configuration, or protocol bridges.

Both DDS and Zenoh pursue this objective through their discovery mechanisms, which transform entity participation events into logical visibility relationships. 
In each case, the middleware resolves the question of where an entity is located into an answer that is invisible to the application: a topic subscription matches a publisher not because the application is aware of its address, but because the middleware has constructed and maintained the appropriate logical association. 
The specific discovery mechanisms used, such as multicast communication, router-mediated scouting, represent distinct implementations of the same spatial abstraction objective.
Consequently, the discovery scope, routing topology, and endpoint matching policies determine which remote entities are visible and reachable from the application’s perspective.

\subsection{Time: Temporal Predictability} 
\emph{Time} denotes the structural objective of ensuring that middleware communication behavior satisfies the temporal requirements of distributed robotic tasks.
The core requirement is temporal predictability, meaning that the middleware must regulate when messages are delivered to provide predictable and bounded latency under varying network conditions. 
Without this property, time-sensitive task loops cannot achieve stable operation across process and host boundaries.

Both DDS and Zenoh pursue this objective through their data exchange mechanisms, which govern how messages are serialized, transported, and delivered.
In each case, the middleware must ensure that the application receives data within a useful time window while internally handling the complexity of transport heterogeneity and congestion. 
The specific transport layers and queue management strategies represent distinct implementations of the same temporal objective.
These mechanisms also differ in how they treat message priority, with the DDS standard defining the transport\_priority policy as a hint whose realization is left to each implementation~\cite{omgdds1.4}, and Zenoh providing priority-driven delivery as part of its data exchange path.
Any modification to these mechanisms, even if it does not affect the network topology, shapes the temporal envelope available to the robotic application. \looseness=-1

\subsection{State: Contextual Continuity} 
\emph{State} denotes the structural objective of preserving state coherence across the system lifecycle despite dynamic participation and intermittent connectivity of entities.
The core requirement is contextual continuity, meaning that the middleware must track, retain, and recover internal information so that late-joining or recovering components can synchronize their state without disrupting the application context. 
Without this property, distributed robotic systems cannot sustain consistent operation across node restarts, network disturbances, or prolonged deployments. \looseness=-1

Both DDS and Zenoh pursue this objective through their state management mechanisms, which maintain the internal state associated with endpoints, participants, and their routing paths. 
In each case, the middleware must address the question of how state is preserved across boundaries where continuous connectivity can not be assumed. 
State management involves a fundamental trade-off between continuity and resource utilization. 
Mechanisms that extend the temporal reach of retained data inevitably consume memory and introduce synchronization overhead.
The two designs differ in where this cost is located, with DDS placing it in HistoryCaches bound to the producing endpoint, and Zenoh assigning it to separate storage managers that need not be bound to the application.
Bounding these costs through resource control policies necessarily limits the degree of continuity available to rejoining components.

\subsection{Structural Trade-offs of Middleware Design} 
The three structural dimensions, \emph{Space}, \emph{Time}, and \emph{State}, do not operate as independent design concerns. 
Each dimension is realized through operational dynamics that share the same execution and spatial boundaries, namely processes and hosts. 
As a result, the mechanisms that serve one dimension inevitably draw upon resources and infrastructure that are also required by the others. 
This shared dependency creates structural coupling among the dimensions, meaning that optimizing one dimension under constrained conditions imposes pressure on the remaining two. \looseness=-1

The first coupling connects \emph{Space} and \emph{Time}. 
Location transparency relies on discovery mechanisms that continuously maintain logical visibility across dynamic participants. 
However, this abstraction conceals the physical network state that temporal performance depends upon, and the concealment itself becomes a source of degradation.
In wireless or bandwidth-constrained environments, this degradation is amplified, because the overhead necessary to sustain a coherent spatial structure competes directly with the temporal budget available for latency-sensitive control and sensing communication, and the middleware possesses no architectural mechanism to detect or interrupt the escalation. \looseness=-1

The second coupling connects \emph{State} and \emph{Time}.
Contextual continuity requires liveliness maintenance, history restoration, and resource control that generate persistent background traffic and buffer overhead. 
In real-time deployments, these mechanisms do not merely consume residual capacity.
They interact with the data exchange path through shared queues, lock-contended dispatch threads, and retransmission pathways.
When these dynamics are active simultaneously, latency increases and jitter becomes less predictable, directly compromising the temporal predictability. \looseness=-1

The third coupling connects \emph{Space} and \emph{State}. 
As the spatial scale of a deployment grows, the quantity of metadata that must be retained and propagated to maintain location transparency increases. 
Node anonymity withholds the physical identity information required for lifecycle state tracking, and path-blind forwarding withholds the network condition information required to prevent buffer saturation. 
Architectural responses to this pressure, such as introducing router intermediaries or centralized discovery servers, reduce the per-participant state cost but concentrate the system's visibility graph at a single point of failure for external connectivity.

Taken together, these three couplings reveal structural trade-offs inherent to ROS~2 middleware. 
Because the mechanisms required by each dimension compete for the same bounded execution capacity, network bandwidth, and buffer memory, fully satisfying \emph{Space}, \emph{Time}, and \emph{State} simultaneously under constrained deployment conditions remains structurally constrained under shared resource bounds. 
Any middleware design must therefore navigate the trade-off surface defined by these three dimensions, accepting partial compromises in at least one dimension to preserve the requirements of the others.

These structural trade-offs motivate the systematic review presented in the remainder of this paper. Section~\ref{sec:tradeoffs} maps 94 middleware-focused studies on ROS~2 middleware onto the trade-off space defined by the three dimensions. 
Rather than categorizing these works by implementation details or individual performance metrics, we organize them according to the primary dimensional conflict each study addresses and the dimension it prioritizes for optimization. 
This classification reveals the structural boundaries that contemporary middleware research has encountered and exposes the architectural gaps that remain unresolved in the design space of distributed robotic middleware. \looseness=-1

\section{Dimensional Conflict Analysis} 
\label{sec:tradeoffs} 
The three dimensions in Section~\ref{sec:dimensions}, namely \emph{Space}, \emph{Time}, and \emph{State}, are not independent axes but represent structurally coupled mechanisms that inherently compete for shared computational and communication resources.
This coupling produces three structural conflict pairs.
First, spatial abstraction imposes unavoidable costs on temporal predictability.
Second, state continuity mechanisms contend for the same resources required by time-critical transmissions.
Third, the anonymity and full-mesh visibility that location transparency requires withhold the identity information that state continuity demands.
Crucially, these conflicts do not arise within a single layer but follow escalation patterns across the system.

In this section, we map 94 middleware-focused studies onto the three conflict pairs identified above. 
We organize them according to the primary dimensional conflict each study addresses and the structural mechanism through which that conflict manifests, rather than cataloging these works by implementation detail or individual performance metric. 
For each conflict pair, we examine the underlying architectural assumptions that give rise to the tension, trace how proposed approaches attempt to relieve it, and identify the structural commonality that recurs across otherwise diverse approaches. 

\subsection{Space--Time Conflict} 
\enlargethispage{\baselineskip}
Spatial abstraction is the cornerstone of ROS~2, providing location transparency that decouples applications from their physical deployment. 
However, maintaining this abstraction imposes an unavoidable structural cost on temporal performance. 
This conflict is not confined to a single architectural layer; rather, it forms an escalation structure that originates within the middleware's internal logic and propagates toward the physical network as the scope of spatial independence widens. 
This subsection examines this escalation through four manifestations: serialization overhead at the middleware boundary, latency amplification induced by concealed physical Maximum Transmission Unit~(MTU) limits, resource redundancy caused by location-agnostic delivery, and jitter amplification by physical multicast. \looseness=-1

\vspace{1em}
\subsubsection{Serialization Overhead} 
Communication across heterogeneous hosts requires the middleware to transform application messages into a standardized wire format that is independent of programming language, endianness, and word size, so that data can traverse process and host boundaries~\cite{luo2023modeling}.
Serialization is therefore intrinsic to communication that crosses host boundaries, required whether or not location transparency is sought.
The Space--Time conflict arises because location transparency renders serialization non-elidable even when physical co-location would permit an intra-host non-serialized path. \looseness=-1

Under large-payload conditions, this structural cost amplifies and propagates into the broader communication pipeline. 
Serialization of high-volume messages directly translates into CPU and memory pressure, as each transmission requires allocation, encoding, and subsequent decoding of the full payload. 
Measurements indicate that, under payloads exceeding a few hundred kilobytes, the dominant fraction of end-to-end communication latency originates in the serialization and memory-handling stages of the DDS, resulting in measurable CPU saturation~\cite{kronauer2021latency}. 
Concurrently, the auxiliary data accompanying packet encapsulation consumes a disproportionately large share of total execution time and available bandwidth, independent of whether the application payload itself is small or large~\cite{luo2023modeling}.
The internal queues maintained to relay messages between the logical and physical layers become saturated when transmission volume exceeds their service rate, blocking the publisher write process and introducing queuing latency into the delivery path~\cite{kang2020study}. \looseness=-1

The problem escalates structurally as endpoint counts increase. 
Analytical modeling of DDS internal queue-processing policies demonstrates that buffer saturation produces compounding delays, where thread batch-processing stalls and queue contention interact, causing latency to grow nonlinearly with load~\cite{luo2023modeling}. 
When multiple publishers simultaneously inject large payloads into shared internal queues, the per-message serialization cost accumulates into sustained throughput degradation, propagating backpressure upstream toward the application callback layer. 
Under these conditions, the communication pipeline degenerates from a bounded-latency channel into a congestion-driven bottleneck, undermining the temporal predictability required by closed-loop robotic tasks.

Several approaches have been proposed to reduce this overhead. 
Alternative message formats such as ROS-SF preserve the C++ struct memory layout to eliminate encoding and decoding without requiring source modification~\cite{wang2022ros}. 
Zenoh-based architectures such as Meta-ROS reduce serialization overhead by optimizing the encoding pipeline under high data-rate conditions~\cite{ranjan2026meta}.
Partial serialization schemes such as TZC decompose messages into a control segment transmitted via socket and a bulk-data segment placed in shared memory, avoiding full-payload copying~\cite{wang2019tzc}. 
Zero-copy mechanisms such as ROS~2 Agnocast eliminate serialization entirely by mapping the publisher heap into shared memory and employing dedicated smart pointers, supporting even unsized message types~\cite{ishikawa2025ros}. 
At the framework level, Flexbot bypasses the DDS middleware for intra-host communication entirely, substituting native FIFO~(First In, First Out) structures and pointer passing to remove serialization, copying, and fragmentation costs~\cite{hietasalo2024overview}. 
More fundamentally, the ROS~2 RMW abstraction inherits the strongly typed DataWriter and DataReader model from DDS, which structurally forecloses serialization elision.
Zenoh performs no serialization when transmitting user data placed in Zenoh-managed shared memory, yet this path is unreachable through the ROS~2 API because the API always serializes user data~\cite{rmw_zenoh_github}. \looseness=-1

Each of these mitigations achieves cost reduction by reintroducing the locality awareness that spatial abstraction withholds.
Alternative encodings that deviate from canonical CDR restrict interoperability across heterogeneous nodes, narrowing the scope of cross-host communication~\cite{xiong2010evaluating}.
Zero-copy and pointer-passing approaches explicitly exploit the physical co-location of endpoints within the same process or host, making location transparency conditional on deployment topology. 
Intra-process optimizations eliminate boundary crossings by design but become inapplicable the moment communicating endpoints are distributed across hosts. 
Every proposed mitigation thus converges on the same structural outcome. 
Avoiding serialization cost requires partially relinquishing spatial abstraction.

\vspace{1em}
\subsubsection{Payload Fragmentation} 
The topic-based publish-subscribe model of ROS~2 is designed to completely conceal the transmission constraints of the physical network from the application layer. 
From the application's perspective, each message constitutes a single logical unit: the middleware accepts an arbitrarily sized payload and assumes responsibility for delivering it to all matched subscribers, irrespective of the physical transport characteristics that lie between publisher and subscriber~\cite{macenski2022robot}. 
This design is a direct expression of location transparency; the application is liberated from any knowledge of the MTU constraint that governs the largest packet a given network link can carry~\cite{kronauer2021latency}. 
The abstraction is structurally sound in its intent, but it introduces a hidden vulnerability.
Large payloads such as camera images and LiDAR point clouds, which are common in robotic perception pipelines, are silently decomposed at the physical layer into hundreds of fragments of approximately 1400~bytes each~\cite{lin2024intelligent}, without any visibility of this decomposition being surfaced to the middleware. \looseness=-1

The consequence of this concealment is that the reliability of the entire logical message becomes contingent on the successful delivery of every individual fragment. 
Under nominal network conditions this dependency is largely invisible, but in wireless or resource-constrained deployments, physical-layer packet loss exposes the fragmentation structure that spatial abstraction had concealed. 
Experimental evaluation under simulated wireless conditions confirms that the retransmission process under a reliable QoS profile produces severe latency spikes, and that repackaging lost data into large retransmission units further degrades packet loss rates~\cite{chen2019performance}. 
Quantitative results demonstrate that hierarchical fragmentation amplifies individual packet errors into message-level latency inflation, as a single lost fragment prevents receiver-side reassembly and forces retransmission of the affected fragment window~\cite{lee2025probabilistic}.

Experiments demonstrate that even a moderate packet error rate on a wireless channel is sufficient to invalidate entire multi-fragmented messages, because the loss of a single MAC-layer frame renders the full reassembly attempt unrecoverable~\cite{peeck2021middleware}.
If the reassembly timer maintained by the OS kernel expires before all fragments arrive, the partially assembled datagram is discarded and all successfully delivered fragments are wasted.
Because the middleware operates above the IP fragmentation layer, it remains unaware of a message's existence until all fragments have been reassembled into a complete datagram. 
As a result, the middleware repeatedly requests retransmission of the entire message rather than targeting only the missing fragments, generating redundant traffic~\cite{lee2025poster}. \looseness=-1

\vspace{1em}
Several approaches have been proposed to mitigate fragmentation-induced latency. 
Node composition and zero-copy APIs eliminate serialization and UDP fragmentation overhead entirely for intra-host communication paths~\cite{macenski2023impact}. 
Shared-memory pointer-passing architectures such as CyberRT resolve large-payload fragmentation by bypassing UDP transport altogether~\cite{jung2025open}. 
Configuring the maximum message size to match the network MTU boundary suppresses uncontrolled IP-layer fragmentation and constrains the retransmission burden triggered by individual packet loss~\cite{lee2025optimizing}. 
Application-layer fragmentation protocols employing bitmap-based acknowledgment decompose large samples into independently retransmissible fragments, confining the cost of individual fragment loss to only the affected unit rather than the full message~\cite{peeck2021middleware}. 
Subscriber-selective transfer mechanisms further reduce exposure to fragment loss by allowing subscribers to request only the message portions relevant to their processing context~\cite{sperling2026low}. 
Transport-level alternatives such as TCP and QUIC, evaluated in automotive middleware comparisons, absorb fragment loss through connection-oriented recovery but introduce connection management overhead and head-of-line blocking under congestion~\cite{kluner2025automotive}. \looseness=-1

Each of these mitigation strategies achieves its temporal benefit by partially dismantling the physical-layer independence that spatial abstraction provides. 
Node composition and shared-memory approaches eliminate fragmentation by exploiting physical co-location, making them inapplicable to the distributed, multi-host deployments~\cite{macenski2022robot}. 
Fragment-aware middleware and MTU-aligned configurations must explicitly recognize the boundaries of the underlying network, which means that transport characteristics originally abstracted away must be reintroduced into the middleware’s operational logic.
Application-layer fragmentation schemes require the publisher to reason about physical link properties and message partitioning structure, directly violating the principle that communicating endpoints remain agnostic to deployment topology~\cite{sperling2026low}. 
TCP-based fallback constrains transport behavior to a unicast client--server connectivity model that is structurally incompatible with the decoupled multicast publish-subscribe paradigm~\cite{blancoenhancing}. 
The structural conclusion is therefore consistent across all approaches: reducing the temporal cost of fragmentation requires the middleware to reintroduce physical-layer awareness into its forwarding path.\looseness=-1

\vspace{1em}
\subsubsection{Intra-Host Redundancy} 
The logical endpoint abstraction that underlies location transparency in ROS~2 middleware manages each subscriber as an independent logical entity, irrespective of where that subscriber is physically deployed. 
This design guarantees uniform communication semantics across all endpoints: a subscriber running on a remote host and a subscriber running on the same host as the publisher receive identical treatment at the middleware layer~\cite{maruyama2016exploring}. 
Any differentiation based on physical co-location would require the system to expose deployment topology to the middleware layer, violating the independence that spatial abstraction is designed to provide. 
Absent explicit co-location awareness, the logical endpoint abstraction has no basis for recognizing that multiple subscribers on the same host could share a single physical copy of a message.
It instead treats each subscriber as an isolated destination and issues a separate, independent transmission for each, with the number of message copies produced on a single host scaling linearly with the number of locally deployed subscribers~\cite{wu2021oops}. \looseness=-1

Under inter-host communication, each transmission requires kernel-level copies through the socket buffers, repeated once per subscriber regardless of physical proximity between subscribers. 
Experiments confirm that unicast-based point-to-point distribution consumes CPU and memory resources that scale linearly with endpoint count, a relationship formally established through complexity analysis~\cite{an2014content}. 
In unicast configurations, publisher-side throughput decreases proportionally as subscriber count grows, because each additional subscriber imposes an independent transmission workload on the same fixed send queue~\cite{kang2021comprehensive}.
Even within a single process, the absence of strict smart-pointer optimization causes a distinct copy of the message to be instantiated per subscriber, confirming that the redundancy originates at the logical separation model rather than at the network layer~\cite{ye2023ros2}. 
Across multiple DDS implementations, results show systematically that per-subscriber latency increases by measurable increments with each additional subscriber under one-to-N configurations, a pattern that persists across implementations and payload sizes~\cite{basem2017data, almadani2015performance}.

As subscriber count increases, the cumulative overhead of redundant intra-host delivery escalates from a localized resource pressure into a system-wide timing constraint. 
Benchmarks of standard unicast delivery confirm that publisher-side Ethernet bandwidth utilization doubles under two-subscriber configurations and that latency grows proportionally thereafter, with throughput falling inversely with subscriber count~\cite{bode2023systematic}. 
In many-to-one architectures, once node count exceeds a threshold, receiver-side buffer overload drives message update failure rates toward saturation and causes average data age to increase sharply~\cite{dust2022quantitative}. 
Analytical modeling of the internal HistoryCache buffers and publisher dispatch thread demonstrates that under high-frequency publication, message accumulation in these queues produces worst-case communication delays that grow with publication rate and subscriber count~\cite{luo2023modeling}. 
In scaled multi-robot deployments, the OS scheduling overhead introduced by independently executing processes further compounds this pressure, consuming over 1400\% of available CPU resources under socket-based baseline configurations~\cite{he2025faster} and producing message loss and latency attributable to the redundant dispatch burden~\cite{dust2022dynamic}. \looseness=-1

Several approaches have been proposed to reduce the overhead of intra-host redundancy. 
Intra-process node composition eliminates serialization and copy overhead for co-located endpoints~\cite{macenski2023impact}, a principle extended to multi-process environments by zero-copy mechanisms such as Agnocast~\cite{ishikawa2025ros}. 
Shared memory transport replaces $N$ independent socket transmissions with a single write, with iceoryx-backed deployments demonstrating near-constant latency regardless of message size~\cite{pohnl2022middleware}.
DDS-based deployments address the same problem through shared-memory transport and intra-process communication optimizations natively supported within Fast DDS and Cyclone DDS~\cite{de2024orchestration, pohnl2022middleware}. 
The Zenoh RMW similarly delivers a single copy to local subscribers through shared-memory transport~\cite{rmw_zenoh_github}.

\vspace{1em}
Each of these solutions achieves its efficiency gain by introducing explicit physical co-location awareness into the middleware's forwarding logic, and in doing so, partially relinquishes the location transparency that motivates the logical endpoint abstraction. 
Shared memory transport requires the middleware to detect whether communicating endpoints reside on the same host and to apply a differentiated delivery path based on that detection, violating uniform communication semantics internally even where the application interface appears unchanged~\cite{de2024orchestration}.
Intra-process communication concentrates endpoints within a shared process boundary, so that a failure in one node propagates to all co-located nodes, a consequence that is architecturally unacceptable in safety-critical deployments such as autonomous driving~\cite{ishikawa2025ros}. 
The performance gains are therefore achievable only within deployment configurations explicitly constrained by physical placement, confirming that logical endpoint independence and the elimination of redundant physical delivery have not been simultaneously achieved within the current middleware design space~\cite{kang2020study}.

\vspace{1em}
\subsubsection{Inter-Host Redundancy} 
Inter-host communication faces a structural redundancy problem when a single publisher delivers a message to multiple subscribers across host boundaries. 
Under naive unicast delivery, the publisher issues separate transmissions per subscriber, so network-level transmission cost scales linearly with the number of remote subscribers. 
ROS 2 middleware addresses this through two mechanisms at different layers, namely multicast at the network layer and router-mediated fan-out at the middleware layer~\cite{kang2021comprehensive}.
The OMG DDS specification relies on multicast as the default transport for discovery traffic within a local network domain~\cite{zhang2024comparison}.
Multicast is expected to provide a stable and predictable delivery path regardless of subscriber count, as the network infrastructure absorbs the replication burden without imposing additional cost on the middleware layer.

The structural vulnerability of this assumption is that packet replication under multicast is not performed at the logical layer by the middleware but at the physical layer by the network switch. 
When a multicast packet arrives at a switch, the switch must inspect its group membership state and produce one copy of the packet per outgoing port associated with that multicast group, a process that consumes switch-internal forwarding resources proportional to both the number of destination ports and the rate at which multicast packets arrive. 
IP multicast forces routers to maintain per-group state that varies over time, imposing scalability and reliability penalties that prevent the mechanism from scaling to large numbers of groups~\cite{hakiri2014supporting}. 
The middleware has no visibility into this threshold, as the physical replication constraint is precisely the kind of infrastructure detail that spatial abstraction is designed to conceal. 
In wireless environments, this concealment carries direct operational consequences, as UDP multicast induces uncontrolled flooding that degrades delivery performance beyond what the middleware layer can detect or compensate~\cite{zhang2024comparison}.

Once the subscriber count crosses the switch's processing threshold, queuing delays accumulate from the mismatch between the logical delivery model and the physical infrastructure's replication capacity, escalating into network-wide jitter amplification. 
Because all subscribers in the multicast group receive packets from the same switch output queue, a single switch bottleneck degrades the timing behavior of the entire logical communication graph. 
Benchmarking studies report that delivery latency remains stable below payload-size and subscriber-count thresholds but rises exponentially once exceeded, with maximum latency spikes approaching 5000 microseconds observed at specific configurations\cite{casesbenchmarking, youssef2015dds, youssef2017dds}.
The escalation is therefore not a gradual degradation but a threshold effect, where system-wide timing predictability collapses abruptly rather than deteriorating incrementally, and the middleware possesses no mechanism for anticipating or signaling this transition to the application layer~\cite{peeroo2023survey}.

Several approaches have been proposed to address this inter-host redundancy.
Unicast fallback replaces multicast delivery with individual per-subscriber transmissions once the subscriber count exceeds a configured threshold, eliminating switch-level replication entirely. 
Experiments demonstrate that beyond endpoint-count thresholds of approximately 50 to 75 publishers and subscribers, unicast communication outperforms multicast in both latency and throughput, providing empirical justification for dynamic mode switching~\cite{peeroo2022exploring}.
Container network interface plugins such as Kubernetes WeaveNet intercept multicast packets and transparently convert them to unicast transmissions at the physical network layer, preventing flooding without requiring changes to the middleware configuration~\cite{kang2021comprehensive}. 
Bridge-based domain segmentation approaches such as zenoh-bridge-ros2dds partition large DDS multicast domains into isolated logical segments interconnected by unicast gossip protocols, resolving infrastructure constraints and topology limitations that standard multicast cannot traverse~\cite{blancoenhancing}. 
In swarm deployments of 20 or more robots, standard UDP multicast over Wi-Fi has been found to be entirely unusable in practice, and centralized discovery servers or unicast-based protocol bridges have been proposed as complete replacements for the multicast-dependent default configuration~\cite{jones2022dots}.
Beyond network-layer multicast, middleware-layer router-mediated fan-out provides an alternative absorption point. 
A Zenoh router on the destination host receives a single message over the network from the publisher and delivers it to local subscribers via the host's intra-host mechanisms, decoupling network-level transmission cost from the number of remote subscribers~\cite{corsaro2021reliability, rmw_zenoh_github}. \looseness=-1

Each of these approaches resolves the timing instability by reintroducing physical infrastructure awareness into a system that spatial abstraction was designed to render infrastructure-agnostic. 
Unicast fallback preserves temporal predictability but abandons the spatial efficiency that motivated multicast adoption in the first place. 
Unless a routing intermediary absorbs the fan-out, publisher-side transmission overhead scales linearly with the subscriber count, reproducing at the logical layer the cost that multicast was designed to absorb at the physical layer~\cite{lee2014router}.
Router-mediated fan-out shifts the concentration point from the publisher to the routing intermediary, relocating the redundancy cost rather than eliminating it~\cite{corsaro2021reliability}.
Bridge-based segmentation requires the deployment architect to partition the logical communication graph according to the physical topology of the underlying network, a dependency that directly contradicts the physical-layer independence that location transparency provides~\cite{blancoenhancing}. 
Correct multicast operation therefore requires operators to possess detailed knowledge of the physical infrastructure rather than treating it as an opaque conduit. 
Recovering temporal predictability under high subscriber counts requires either sacrificing multicast's spatial efficiency or accepting infrastructure management costs. \looseness=-1

\begin{table*}[t]
\centering
\caption{Taxonomy of Literature Addressing the Space--Time Conflict}
\label{tab:sp conflict}
\resizebox{\textwidth}{!}{%
\begin{tabular}{|>{\centering\arraybackslash}m{0.7cm}|>{\centering\arraybackslash}m{1cm}|>{\centering\arraybackslash}m{2.5cm}|>{\centering\arraybackslash}m{2.8cm}|>{\centering\arraybackslash}m{0.7cm}|>{\centering\arraybackslash}m{0.7cm}|>{\centering\arraybackslash}m{0.7cm}|>{\centering\arraybackslash}m{0.7cm}|>{\centering\arraybackslash}m{0.7cm}|>{\centering\arraybackslash}m{0.7cm}|>{\centering\arraybackslash}m{2.5cm}|}
\hline
    \multirow{2}{*}{\textbf{[Ref]}} & \multirow{2}{*}{\textbf{Protocol}} & \multirow{2}{*}{\textbf{Application Domain}} & \multirow{2}{*}{\textbf{Evaluation Platform}} & \multicolumn{3}{c|}{\textbf{Network}} & \multicolumn{3}{c|}{\textbf{Research Type}} & \multirow{2}{*}{\textbf{Key Approach}} \\
    \cline{5-10}
    & & & & \textbf{\shortstack{Wired}} & \textbf{\shortstack{Wire-\\less}} & \textbf{\shortstack{Local\\Host}} & \textbf{\shortstack{Evalu-\\ation}} & \textbf{\shortstack{System\\Design}} & \textbf{\shortstack{Model-\\ing}} & \\
    \hline
    \cite{maruyama2016exploring} & \multirow{35}{*}{DDS} & \multirow{1}{*}{Robotics, Auto.} & \multirow{1}{*}{Workstation} & $\checkmark$ &  & $\checkmark$ & $\checkmark$ &  &  & Unicast-baseline \\
    \cline{1-1} \cline{3-11}
    \cite{de2024orchestration} &  & \multirow{14}{*}{Robotics} & \multirow{2}{*}{Embedded, Workstation} &  &  & $\checkmark$ & $\checkmark$ & $\checkmark$ &  & SHM-orchestration \\
    \cline{1-1}
    \cite{kronauer2021latency} &  &  &  &  &  & $\checkmark$ & $\checkmark$ &  &  & Serialization-profiling \\
    \cline{1-1} \cline{4-11}
    \cite{macenski2023impact} &  &  & \multirow{1}{*}{Embedded} &  &  & $\checkmark$ & $\checkmark$ &  &  & Node-composition \\
    \cline{1-1} \cline{4-11}
    \cite{dust2022dynamic} &  &  & \multirow{1}{*}{Simulator} &  &  & $\checkmark$ & $\checkmark$ & $\checkmark$ &  & Process-overhead \\
    \cline{1-1} \cline{4-11}
    \cite{dust2022quantitative} &  &  & \multirow{10}{*}{Workstation} & $\checkmark$ &  & $\checkmark$ & $\checkmark$ &  &  & Subscriber-scaling \\
    \cline{1-1}
    \cite{hietasalo2024overview} &  &  &  & $\checkmark$ &  & $\checkmark$ & $\checkmark$ & $\checkmark$ &  & Middleware-bypass \\
    \cline{1-1}
    \cite{lee2025optimizing} &  &  &  &  & $\checkmark$ &  & $\checkmark$ & $\checkmark$ &  & Burst-analysis \\
    \cline{1-1}
    \cite{lee2025poster} &  &  &  &  & $\checkmark$ &  & $\checkmark$ & $\checkmark$ &  & Burst-analysis \\
    \cline{1-1}
    \cite{lee2025probabilistic} &  &  &  & $\checkmark$ &  &  & $\checkmark$ &  & $\checkmark$ & Heartbeat-modeling \\
    \cline{1-1}
    \cite{luo2023modeling} &  &  &  &  &  & $\checkmark$ & $\checkmark$ &  & $\checkmark$ & Queue-modeling \\
    \cline{1-1}
    \cite{macenski2022robot} &  &  &  &  &  & $\checkmark$ & $\checkmark$ &  &  & P2P-motivation \\
    \cline{1-1}
    \cite{wang2019tzc} &  &  &  &  &  & $\checkmark$ & $\checkmark$ & $\checkmark$ &  & Split-transport \\
    \cline{1-1}
    \cite{wang2022ros} &  &  &  & $\checkmark$ &  & $\checkmark$ & $\checkmark$ & $\checkmark$ &  & Layout-preserving \\
    \cline{1-1}
    \cite{ye2023ros2} &  &  &  &  &  & $\checkmark$ & $\checkmark$ &  &  & Intra-process-copy \\
    \cline{1-1} \cline{3-11}
    \cite{kang2021comprehensive} &  & \multirow{7}{*}{IIoT} & \multirow{1}{*}{Embedded, Workstation} & $\checkmark$ &  & $\checkmark$ & $\checkmark$ &  &  & Container-overhead \\
    \cline{1-1} \cline{4-11}
    \cite{bode2023systematic} &  &  & \multirow{1}{*}{Embedded} & $\checkmark$ &  &  & $\checkmark$ &  &  & Bandwidth-profiling \\
    \cline{1-1} \cline{4-11}
    \cite{almadani2015performance} &  &  & \multirow{5}{*}{Workstation} &  & $\checkmark$ &  & $\checkmark$ &  & $\checkmark$ & Subscriber-latency \\
    \cline{1-1}
    \cite{basem2017data} &  &  &  & $\checkmark$ & $\checkmark$ &  & $\checkmark$ & $\checkmark$ &  & QoS-profiling \\
    \cline{1-1}
    \cite{hakiri2014supporting} &  &  &  & $\checkmark$ &  &  & $\checkmark$ & $\checkmark$ &  & Multicast-scalability \\
    \cline{1-1}
    \cite{kang2020study} &  &  &  & $\checkmark$ &  &  & $\checkmark$ &  &  & AutoThrottle \\
    \cline{1-1}
    \cite{lin2024intelligent} &  &  &  & $\checkmark$ &  & $\checkmark$ & $\checkmark$ & $\checkmark$ &  & App-fragmentation \\
    \cline{1-1} \cline{3-11}
    \cite{wu2021oops} &  & \multirow{6}{*}{Auto.} & \multirow{1}{*}{Embedded, Workstation} &  &  & $\checkmark$ & $\checkmark$ &  &  & Copy-overhead \\
    \cline{1-1} \cline{4-11}
    \cite{pohnl2022middleware} &  &  & \multirow{1}{*}{Embedded} &  &  & $\checkmark$ & $\checkmark$ & $\checkmark$ &  & Zero-copy-SHM \\
    \cline{1-1} \cline{4-11}
    \cite{sperling2026low} &  &  & \multirow{1}{*}{Simulator, Workstation} & $\checkmark$ &  &  & $\checkmark$ & $\checkmark$ &  & Subscriber-selective \\
    \cline{1-1} \cline{4-11}
    \cite{chen2019performance} &  &  & \multirow{2}{*}{Simulator} &  & $\checkmark$ &  & $\checkmark$ & $\checkmark$ &  & Wireless-fragment \\
    \cline{1-1}
    \cite{peeck2021middleware} &  &  &  &  & $\checkmark$ &  & $\checkmark$ & $\checkmark$ &  & Bitmap-fragmentation \\
    \cline{1-1} \cline{4-11}
    \cite{jung2025open} &  &  & \multirow{1}{*}{Workstation} &  &  & $\checkmark$ & $\checkmark$ &  &  & SHM-bypass \\
    \cline{1-1} \cline{3-11}
    \cite{youssef2017dds} &  & \multirow{7}{*}{(None)} & \multirow{1}{*}{Simulator, Workstation} & $\checkmark$ &  &  & $\checkmark$ & $\checkmark$ &  & Threshold-modeling \\
    \cline{1-1} \cline{4-11}
    \cite{lee2014router} &  &  & \multirow{2}{*}{Simulator} & $\checkmark$ &  &  & $\checkmark$ & $\checkmark$ &  & Unicast-fallback \\
    \cline{1-1}
    \cite{youssef2015dds} &  &  &  & $\checkmark$ &  &  & $\checkmark$ & $\checkmark$ &  & Unicast-threshold \\
    \cline{1-1} \cline{4-11}
    \cite{an2014content} &  &  & \multirow{4}{*}{Workstation} & $\checkmark$ &  &  & $\checkmark$ & $\checkmark$ &  & Complexity-modeling \\
    \cline{1-1}
    \cite{casesbenchmarking} &  &  &  & $\checkmark$ &  &  & $\checkmark$ & $\checkmark$ &  & Multicast-spike \\
    \cline{1-1}
    \cite{peeroo2022exploring} &  &  &  &  &  & $\checkmark$ & $\checkmark$ &  &  & Topology-aware \\
    \cline{1-1}
    \cite{xiong2010evaluating} &  &  &  & $\checkmark$ &  &  & $\checkmark$ &  &  & CDR-interoperability \\
    \hline
    \cite{jones2022dots} & \multirow{7}{*}{Both} & \multirow{3}{*}{Robotics, IIoT} & \multirow{1}{*}{Embedded, Simulator} &  & $\checkmark$ &  & $\checkmark$ & $\checkmark$ &  & Swarm-bridging \\
    \cline{1-1} \cline{4-11}
    \cite{zhang2024comparison} &  &  & \multirow{2}{*}{Embedded, Workstation} & $\checkmark$ & $\checkmark$ &  & $\checkmark$ &  &  & Protocol-comparison \\
    \cline{1-1}
    \cite{ranjan2026meta} &  &  &  & $\checkmark$ & $\checkmark$ & $\checkmark$ & $\checkmark$ & $\checkmark$ &  & Encoding-tuning \\
    \cline{1-1} \cline{3-11}
    \cite{blancoenhancing} &  & \multirow{1}{*}{Robotics} & \multirow{1}{*}{Workstation} & $\checkmark$ & $\checkmark$ &  & $\checkmark$ & $\checkmark$ &  & Domain-bridging \\
    \cline{1-1} \cline{3-11}
    \cite{he2025faster} &  & \multirow{3}{*}{Auto.} & \multirow{1}{*}{Embedded, Workstation} &  &  & $\checkmark$ & $\checkmark$ & $\checkmark$ &  & Middleware-bypass \\
    \cline{1-1} \cline{4-11}
    \cite{ishikawa2025ros} &  &  & \multirow{2}{*}{Workstation} &  &  & $\checkmark$ & $\checkmark$ & $\checkmark$ &  & Heap-zerocopy \\
    \cline{1-1}
    \cite{kluner2025automotive} &  &  &  & $\checkmark$ &  & $\checkmark$ & $\checkmark$ &  &  & Transport-comparison \\
    \hline
\end{tabular}
}
\vspace{-0.3em}
\end{table*}
\vspace{0.5em}
Table~\ref{tab:sp conflict} summarizes the 42 studies surveyed under the Space--Time conflict.
The majority target DDS and are evaluated on wired or local-host configurations, a distribution that reflects the dominance of controlled testbed environments in the literature and partially explains why the physical-layer consequences of spatial abstraction have been undercharacterized relative to their operational severity.
Wireless evaluations are concentrated in the fragmentation and multicast subgroups, where physical-layer variability directly exposes the hidden costs that location transparency imposes on delivery timing. 
Modeling studies remain a minority across all four manifestations, indicating that analytical characterization of Space--Time interactions lags behind empirical evaluation. 
The key approach column reveals that proposed mitigations consistently fall into one of two structural categories: those that exploit physical co-location, such as shared-memory transport, zero-copy mechanisms, and node composition, and those that reintroduce network-layer awareness, such as MTU-alignment, bitmap-based fragmentation, and domain bridging. 
This distribution reflects the underlying structure of the conflict itself. 
Regardless of which manifestation a study targets, every viable mitigation either narrows the deployment scope to physically constrained configurations or partially dismantles the network independence that spatial abstraction provides, confirming that the Space--Time conflict has been navigated at the cost of one or both of its constituent guarantees. 

\subsection{Time--State Conflict} 
Mechanisms for contextual continuity, including reliability protocols and liveliness monitoring, ensure logical consistency in the face of dynamic participation and intermittent connectivity. 
Yet, the execution of these mechanisms consumes the same computational and communication resources demanded by temporal performance. 
This conflict initiates at the middleware execution layer and escalates into network-wide resource competition, occasionally triggering uncontrollable positive feedback loops.
The trade-off is structural: the more strictly a system enforces state consistency, the more resources it consumes for state management. 
This subsection examines three manifestations of this escalation.
 We analyze jitter from internal synchronization, bandwidth erosion by control messages, and congestion arising from reliability-driven recovery. \looseness=-1

\vspace{1em}
\subsubsection{Lock Contention} 
Guaranteeing the logical ordering of messages and the consistency of shared state in a distributed environment is a structural requirement for ROS~2 middleware. 
Without enforced ordering and synchronized access to internal data structures, the middleware cannot provide the contextual continuity that applications depend upon across dynamic participation and concurrent message flows.
To satisfy this requirement, the middleware imposes synchronization at its internal queue and callback layers, where multiple execution threads concurrently access shared endpoint state, message histories, and dispatch queues. 
This synchronization is the mechanism by which state consistency is enforced, and its presence is proportional to the strength of the consistency guarantee the middleware is required to provide~\cite{zieba2006preservation}. 
As the continuity requirements intensify, through stricter ordering guarantees, deeper history retention, or higher callback concurrency, the frequency and duration of synchronization operations increase correspondingly.

The primary manifestation of this synchronization cost at the execution layer is lock contention. 
When multiple threads simultaneously attempt to access a shared queue or endpoint state structure, all but one must enter a waiting state until the lock is released by the current holder. 
The duration of each waiting interval is nondeterministically distributed, as it depends on the scheduling behavior of the OS, the execution time of the lock holder, and the number of competing threads. 
A single-threaded ROS~2 executor serializes all subscription callbacks through a fixed polling-point mechanism, causing buffer overflow and message loss when multiple publishers simultaneously saturate the dispatch path~\cite{teper2023timing}. 
In multi-threaded configurations, publishers, flow controllers, and listeners interact through shared pending message queues, where contention introduces context-switching overhead and nondeterministic delays~\cite{dehnavi2021compros}. 
As concurrent callback invocations increase, inter-thread interference at these shared structures compounds, degrading timing predictability across the execution layer~\cite{sciangula2023bounding}. \pagebreak[4]

As system complexity increases, lock contention escalates from isolated execution stalls into a system-wide source of timing nondeterminism that propagates across middleware, OS, and kernel layers. 
At the kernel level, FIFO ring buffer processing and thread scheduling contention have been identified as the root cause of end-to-end response time degradation from five~ms to 122~ms under increasing message load, with jitter scaling directly with publication rate and message size~\cite{kim2025cros}. 
Context switching overhead further amplifies this effect: each lock acquisition failure causes the blocked thread to yield its CPU slot, triggering a context switch whose latency adds to the total waiting time and whose frequency grows with the contention rate. 
Under high publication rates, the combined effect of lock waiting and context switching causes internal dispatch latency to exceed the inter-message interval. 
At this point, the callback queue accumulates a backlog. 
Consequently, the delivery timing of subsequent messages becomes correlated with the processing delays of their predecessors rather than their arrival times~\cite{teper2023timing}. 

Several research directions have addressed contention and timing nondeterminism through structural improvements at multiple layers. 
Distributed state synchronization using conflict-free replicated data types and atomic operations eliminates halting synchronization from concurrent execution paths without requiring modification of the middleware internals, a principle demonstrated using Zenoh~\cite{shih2022scalable}. 
Assigning mutually exclusive callback groups to dedicated executors separates synchronization concerns from data publication without requiring changes to application code~\cite{abaza2024managing}. 
The synchronization burden can further be removed from the execution path by delegating history retention and state consistency to an independent storage layer, structurally decoupling state management from the latency-critical data exchange pipeline~\cite{corsaro2023zenoh}.
To resolve this at the middleware layer, priority-aware executor redesigns introduce native preemptive scheduling models such as producer-consumer and thread-dispatch variants. 
These redesigns allow high-priority callbacks to predictably preempt lower-priority ones without external OS interventions~\cite{randolph2021improving}.
Middleware-integrated QoS resource managers extend this predictability further by interfacing directly with the RTOS to enforce budget scheduling and temporal isolation, maintaining correct execution semantics during real-time system reconfiguration~\cite{valls2012iland}.

Each approach reduces contention overhead by accepting constraints on state consistency or infrastructure.
Storage-based decoupling removes the synchronization burden from the execution path but introduces an independent storage infrastructure, adding resource and configuration overhead.
Priority-aware executors improve latency predictability for prioritized callbacks but do not eliminate contention for lower-priority flows~\cite{randolph2021improving}.
To enable end-to-end timing analysis, communication flows must be decoupled, which means accepting delayed consistency over immediate processing, and strict destination order guarantees must be relaxed to make the system schedulable~\cite{perez2015modeling}.
Under these relaxed consistency conditions, when asynchronous publication rates exceed subscription processing rates, message loss cannot be prevented even under bounded buffer configurations, meaning that temporal predictability can only be preserved by accepting data loss rather than full state continuity~\cite{teper2023timing}.
The structural outcome is consistent: achieving temporal predictability in the presence of concurrent state access requires either weakening the consistency guarantees or redistributing consistency responsibility onto a separate architectural component.

\vspace{1em}
\subsubsection{Control Overhead}
Reliable communication in ROS~2 middleware requires each entity to continuously exchange control messages with its peers, including Heartbeat announcements, AckNack acknowledgments, and session management messages, to verify liveliness and delivery state.
These control traffic constitute a permanent structural tax on network capacity.
The OMG DDS specification mandates that each entity continuously signal its alive status within the configured lease duration, leaving the middleware no architectural basis for suspending this exchange during periods of low data activity~\cite{omgdds1.4}.
Control traffic is a steady-state background load primarily driven by the number of active endpoint relationships, though it remains partially influenced by application data rates.
In multi-robot deployments, control messages for data-readiness notifications and keepalive mechanisms accumulate under scaled configurations. 
This accumulation has been directly measured to cause bandwidth overflow and network saturation~\cite{castillo2024swarm}.

In DDS, control messages and application messages compete for the same network bandwidth and CPU resources without priority separation, so Heartbeat and acknowledgment signals directly consume bandwidth otherwise available for latency-sensitive delivery~\cite{lee2025dependency}.
This contention is further compounded in containerized environments, where network virtualization plugins introduce additional CPU and bandwidth competition that becomes particularly pronounced at high transmission frequencies and small payload sizes~\cite{kang2021comprehensive}.
In wireless lossy networks, this interference manifests as local latency spikes under reliable QoS, where lower-layer loss recovery and middleware-level retransmission triggered by missed acknowledgments compete directly for the same path~\cite{thulasiraman2020evaluation}.
As participant count grows, discovery and Heartbeat or AckNack retransmission traffic scales linearly and saturates the software transmission queue at the data link layer, creating a bottleneck shared with application data~\cite{kim2026mitigating}. \looseness=-1

As system scale increases, the bandwidth erosion caused by control traffic escalates from a marginal overhead into a dominant structural constraint.
This erosion is continuous: each active endpoint relationship contributes a persistent stream of Heartbeat and acknowledgment exchanges that cannot be suspended regardless of application data rate.
Prior work formally proves that the total transfer count in a unicast domain scales as $O(E \times P)$, where $E$ and $P$ denote the number of endpoints and participants in the domain, respectively, exhausting both network resources and memory as the endpoint population grows~\cite{an2014content}.
Empirical scaling experiments confirm that as the number of ROS~2 nodes increases from two to five, latency spikes sharply and message drop rate rises exponentially, demonstrating that this steady-state bandwidth erosion constitutes a systemic bottleneck rather than a localized performance penalty~\cite{thulasiraman2020evaluation}.

Several approaches have been proposed to reduce the bandwidth footprint of control traffic without abandoning the reliability guarantees it supports.
Heartbeat period and lease duration tuning reduces the frequency of liveliness announcements by adjusting the interval between consecutive control message transmissions, and analytical modeling confirms that the Heartbeat period directly governs the frequency of AckNack generation and retransmission attempts~\cite{park2025analytical, lee2025probabilistic}.
DDS parameter optimization, including synchronous publish modes and QoS tuning such as best-effort QoS and keep last history policies for small high-frequency packets, has been shown to reduce communication latency and improve real-time performance in distributed robotic control scenarios~\cite{plasberg2022towards}.
At the protocol level, batching and coalescing mechanisms aggregate multiple messages into single message; both DDS implementations and Zenoh's session protocol employ batching to consolidate control messages and minimize protocol overhead.
More fundamentally, transport-level reliability substitution via connection-oriented protocols such as TCP or QUIC removes the per-endpoint AckNack exchange from the middleware-visible communication path entirely.
This has been demonstrated using Zenoh~\cite{baron2025performance}, and comparable latency benefits are achievable in DDS deployments by switching to TCP-based transport configurations~\cite{chisualițua2025stepping}. 
At the architectural level, Zenoh assigns a configurable priority to control traffic, allowing an architect to shift this competition between control and application flows rather than eliminate it. \looseness=-1

Each mitigation strategy either reduces control traffic at a cost to contextual continuity, displaces that responsibility onto the transport layer, or reorders the contention through priority assignment without removing it.
Heartbeat period tuning and batching reduce message frequency but increase the latency with which liveliness failures are detected, directly weakening the timeliness of contextual continuity maintenance~\cite{lee2025dependency}.
In large-scale multi-robot deployments, sustained state consistency maintenance has been observed to generate excessive data flows that exhaust communication resources and destabilize the network~\cite{ginting2021chord}.
Experiments further demonstrate that, in DDS-based deployments, activating complete security and reliability guarantees reduces throughput from approximately 70,000 to 14,000 packets per second and increases latency by a factor of five~\cite{plasberg2022towards}, and that achieving timing determinism in practice requires adopting best-effort QoS rather than reliable QoS~\cite{diluoffo2018robot}.
The dimensional outcome is consistent: reducing the steady-state cost of control traffic requires either weakening the state consistency that the traffic exists to maintain or accepting degraded temporal predictability~\cite{kang2021comprehensive}. \looseness=-1

\vspace{1em}
\subsubsection{Retransmission Congestion}
\label{subsubsec:3}
Reliable communication in ROS~2 middleware is predicated on the assumption that delivery failures must be detected and corrected as quickly as possible to preserve contextual continuity across distributed nodes.
Under reliable QoS, the DDS specification requires the middleware to retransmit missing samples and replay retained history for reconnecting endpoints, treating every unacknowledged sample as a deficit to be resolved without delay~\cite{omgdds1.4, lee2025dependency}.
Consequently, the DDS architecture prioritizes state completeness over transmission efficiency, as the middleware must reconstruct a correct historical snapshot before exposing new message to the application~\cite{zieba2006preservation}.\looseness=-1

The structural vulnerability of this approach surfaces specifically when the underlying network is already operating under congestion.
Because the reliability mechanism monitors endpoint-level sequence state rather than network-level resource availability, it cannot distinguish between isolated packet loss and symptomatic of link saturation.
In both cases, the response is identical: retransmit immediately.
Analytical modeling shows that periodic Heartbeat messages broadcast by the publisher induce AckNack responses from receivers and trigger immediate retransmission for every detected sequence gap, establishing a structurally unconditional retransmission pathway that operates independently of channel load~\cite{park2025analytical}.
Under congested conditions, this retransmission injects additional traffic into a channel that is already unable to clear its existing load, directly increasing queue occupancy and elevating the probability of further loss.
The contrast between reliable and best-effort QoS makes this dynamic explicit.
Reliable QoS continuously retries every missing sample in the HistoryCache, consuming retransmission resources without bound, whereas best-effort does not attempt recovery and thereby avoids contributing to congestion~\cite{al2012wireless}.

The transition from isolated retransmission events to system-wide degradation follows a positive feedback structure.
Each retransmission attempt adds traffic to an already-saturated channel, which induces further packet loss, which in turn triggers additional retransmission, forming a self-reinforcing loop in which the state recovery mechanism itself amplifies the loss it was designed to correct.
Experiments demonstrate that scaling ROS~2 nodes over DDS in a wireless lossy network under reliable QoS causes average latency and message drop rate to increase exponentially, reaching communication collapse as retransmission attempts overwhelm the available channel capacity~\cite{thulasiraman2022evaluation}.
History recovery at reconnection further compounds this escalation.
The middleware floods the channel with retained samples at the moment of rejoining, competing directly with control messages on the same transport path and prolonging the period during which the feedback loop remains active.
As system scale grows, the communication cost and computation overhead associated with state synchronization increase concurrently~\cite{hu2024wireless}, making the system increasingly susceptible to retransmission feedback loops.

Several approaches have been proposed to interrupt the positive feedback loop.
Time-based filtering mechanisms discard intermediate updates under high-volume delivery conditions, preventing channel flooding without disabling the reliability mechanism entirely~\cite{almadani2016qos}.
Custom QoS policies incorporating inter-robot velocity differentials can dynamically adjust transmission window sizes to minimize retransmission events~\cite{dey2023novel}.
Congestion-aware retransmission control feeds network congestion signals from the 5G layer back into the middleware, enabling proactive throttling before saturation is reached~\cite{szabo2023toward}.
Disabling MAC-layer retransmission and replacing the AckNack loop with a purpose-built protocol prevents the collision between lower-layer and middleware-layer retransmission cycles~\cite{peeck2021middleware}.
Zenoh supports storage-based separation of history recovery from active data exchange.
Zenoh provides dedicated storage managers that retain samples for specific key expressions and serve them to late joiners via pull-based queries, ensuring that history delivery no longer competes with ongoing transmissions on the primary data channel~\cite{zettascale_zenoh_storage}.
DDS can achieve analogous separation through persistent durability services.
Fundamentally, leveraging the frame replication within Time-Sensitive Networking~(TSN) bypasses middleware-level retransmissions entirely by providing lossless redundancy at the network layer~\cite{li2026fault}, but doing so couples the middleware to a network configuration, introducing a Space--State conflict. \looseness=-1

These approaches achieve their benefits by accepting a constraint on either the timeliness of state recovery or the completeness of delivery guarantees.
Congestion-aware throttling introduces delay between loss detection and retransmission delivery.
Multi-robot results demonstrate that network bottlenecks can only be prevented by differentially applying reliable and transient local QoS to high-priority information while relegating other data to best-effort and volatile QoS, confirming that complete reliability guarantees and real-time network performance have not been simultaneously achieved in current deployments~\cite{jo2025generation}.
Storage-based decoupling removes the reconnection burst but introduces a consistency boundary between the storage manager and the producing endpoint whose divergence must be managed as a separate operational concern, trading one form of state overhead for another.
No mechanism simultaneously preserves the immediacy of state recovery and protects temporal performance under congestion; every viable approach either delays state restoration or redistributes its cost onto a separate architectural component.

\begin{table*}[t]
\centering
\caption{Taxonomy of Literature Addressing the Time--State Conflict}
\label{tab:ts conflict}
\resizebox{\textwidth}{!}{%
\begin{tabular}{|>{\centering\arraybackslash}m{0.7cm}|>{\centering\arraybackslash}m{1cm}|>{\centering\arraybackslash}m{2.5cm}|>{\centering\arraybackslash}m{2.8cm}|>{\centering\arraybackslash}m{0.7cm}|>{\centering\arraybackslash}m{0.7cm}|>{\centering\arraybackslash}m{0.7cm}|>{\centering\arraybackslash}m{0.7cm}|>{\centering\arraybackslash}m{0.7cm}|>{\centering\arraybackslash}m{0.7cm}|>{\centering\arraybackslash}m{2.5cm}|}
    \hline
    \multirow{2}{*}{\textbf{[Ref]}} & \multirow{2}{*}{\textbf{Protocol}} & \multirow{2}{*}{\textbf{Application Domain}} & \multirow{2}{*}{\textbf{Evaluation Platform}} & \multicolumn{3}{c|}{\textbf{Network}} & \multicolumn{3}{c|}{\textbf{Research Type}} & \multirow{2}{*}{\textbf{Key Approach}} \\
    \cline{5-10}
    & & & & \textbf{\shortstack{Wired}} & \textbf{\shortstack{Wire-\\less}} & \textbf{\shortstack{Local\\Host}} & \textbf{\shortstack{Evalu-\\ation}} & \textbf{\shortstack{System\\Design}} & \textbf{\shortstack{Model-\\ing}} & \\
    \hline
    \cite{szabo2023toward} & \multirow{31}{*}{DDS} & Robotics, IIoT & Workstation, Simulator & & $\checkmark$ & & $\checkmark$ & $\checkmark$ & & Congestion-throttle \\
    \cline{1-1} \cline{3-11}
    \cite{teper2023timing} & & \multirow{5}{*}{Robotics, Auto.} & Workstation, Simulator & & & $\checkmark$ & $\checkmark$ & $\checkmark$ & $\checkmark$ & Executor-modeling \\
    \cline{1-1} \cline{4-11}
    \cite{thulasiraman2022evaluation} & & & Simulator & & $\checkmark$ & & $\checkmark$ & & & Wireless-scaling \\
    \cline{1-1} \cline{4-11}
    \cite{abaza2024managing} & & & \multirow{3}{*}{Workstation} & & & $\checkmark$ & $\checkmark$ & $\checkmark$ & & Callback-isolation \\
    \cline{1-1}
    \cite{randolph2021improving} & & & & & & $\checkmark$ & $\checkmark$ & $\checkmark$ & $\checkmark$ & Preemptive-executor \\
    \cline{1-1}
    \cite{sciangula2023bounding} & & & & & & $\checkmark$ & $\checkmark$ & & $\checkmark$ & Contention-modeling \\
    \cline{1-1} \cline{3-11}
    \cite{hu2024wireless} & & \multirow{15}{*}{Robotics} & \shortstack{Embedded, Workstation, \\ Simulator} & & $\checkmark$ & & $\checkmark$ & $\checkmark$ & & Sync-scaling \\
    \cline{1-1} \cline{4-11}
    \cite{dey2023novel} & & & \multirow{2}{*}{Embedded, Simulator} & & $\checkmark$ & & $\checkmark$ & $\checkmark$ & & Adaptive-QoS \\
    \cline{1-1}
    \cite{jo2025generation} & & & & & $\checkmark$ & & $\checkmark$ & $\checkmark$ & & Policy-selection \\
    \cline{1-1} \cline{4-11}
    \cite{ginting2021chord} & & & \multirow{4}{*}{Embedded} & & $\checkmark$ & & $\checkmark$ & $\checkmark$ & & Swarm-overhead \\
    \cline{1-1}
    \cite{castillo2024swarm} & & & & & $\checkmark$ & & $\checkmark$ & & & Wireless-scaling \\
    \cline{1-1}
    \cite{kim2025cros} & & & & $\checkmark$ & & & $\checkmark$ & $\checkmark$ & $\checkmark$ & Kernel-contention \\
    \cline{1-1}
    \cite{dehnavi2021compros} & & & & & & $\checkmark$ & $\checkmark$ & $\checkmark$ & & Queue-isolation \\
    \cline{1-1} \cline{4-11}
    \cite{thulasiraman2020evaluation} & & & Simulator & & $\checkmark$ & & $\checkmark$ & & & Control-scaling \\
    \cline{1-1} \cline{4-11}
    \cite{plasberg2022towards} & & & \multirow{6}{*}{Workstation} & $\checkmark$ & & & $\checkmark$ & & & BestEffort-tuning \\
    \cline{1-1}
    \cite{diluoffo2018robot} & & & & & $\checkmark$ & $\checkmark$ & $\checkmark$ & $\checkmark$ & & Reliability-cost \\
    \cline{1-1}
    \cite{lee2025probabilistic} & & & & $\checkmark$ & & & $\checkmark$ & & $\checkmark$ & Heartbeat-modeling \\
    \cline{1-1}
    \cite{kim2026mitigating} & & & & & $\checkmark$ & & $\checkmark$ & $\checkmark$ & $\checkmark$ & Discovery-saturation \\
    \cline{1-1}
    \cite{lee2025dependency} & & & & & $\checkmark$ & & & $\checkmark$ & $\checkmark$ & Control-dependency \\
    \cline{1-1}
    \cite{park2025analytical} & & & & $\checkmark$ & & & $\checkmark$ & & $\checkmark$ & Retransmit-modeling \\
    \cline{1-1} \cline{3-11}
    \cite{kang2021comprehensive} & & \multirow{3}{*}{IIoT} & Embedded, Workstation & $\checkmark$ & & $\checkmark$ & $\checkmark$ & & & Container-overhead \\
    \cline{1-1} \cline{4-11}
    \cite{almadani2016qos} & & & \multirow{2}{*}{Workstation} & & $\checkmark$ & & $\checkmark$ & $\checkmark$ & & Time-filtering \\
    \cline{1-1}
    \cite{valls2012iland} & & & & $\checkmark$ & & & $\checkmark$ & $\checkmark$ & & Budget-scheduling \\
    \cline{1-1} \cline{3-11}
    \cite{li2026fault} & & \multirow{3}{*}{Auto.} & Embedded, Workstation & $\checkmark$ & & & $\checkmark$ & $\checkmark$ & $\checkmark$ & TSN-replication \\
    \cline{1-1} \cline{4-11}
    \cite{perez2015modeling} & & & Embedded & $\checkmark$ & & & $\checkmark$ & & $\checkmark$ & Order-relaxation \\
    \cline{1-1} \cline{4-11}
    \cite{peeck2021middleware} & & & Simulator & & $\checkmark$ & & $\checkmark$ & $\checkmark$ & & Bitmap-fragmentation \\
    \cline{1-1} \cline{3-11}
    \cite{al2012wireless} & & \multirow{4}{*}{(None)} & \multirow{2}{*}{Workstation} & & $\checkmark$ & & $\checkmark$ & & & QoS-contrast \\
    \cline{1-1}
    \cite{an2014content} & & & & $\checkmark$ & & & $\checkmark$ & $\checkmark$ & & Complexity-modeling \\
    \cline{1-1} \cline{4-11}
    \cite{zieba2006preservation} & & & \multirow{2}{*}{-} & & & & & $\checkmark$ & & Consistency-model \\
    \cline{1-1}
    \cite{omgdds1.4} & & & & & & & & $\checkmark$ & & Spec-analysis \\
    \hline
    \cite{baron2025performance} & Zenoh & IIoT & Embedded & $\checkmark$ & $\checkmark$ & & $\checkmark$ & & & Transport-replace \\
    \hline
    \cite{corsaro2023zenoh} & \multirow{3}{*}{Both} & \multirow{2}{*}{Robotics, IIoT, Auto.} & \multirow{2}{*}{Workstation} & $\checkmark$ & & $\checkmark$ & $\checkmark$ & $\checkmark$ & & State-decoupling \\
    \cline{1-1}
    \cite{shih2022scalable} & & & & & & $\checkmark$ & $\checkmark$ & $\checkmark$ & & Lock-free \\
    \cline{1-1} \cline{3-11}
    \cite{chisualițua2025stepping} & & Auto. & Embedded, Workstation & $\checkmark$ & $\checkmark$ & & $\checkmark$ & & & Protocol-comparison \\
    \hline
\end{tabular}
}
\end{table*}
\vspace{0.5em}
Table~\ref{tab:ts conflict} summarizes the 34 studies surveyed under the Time--State conflict. 
Wireless evaluations account for a notably higher proportion than in the Space--Time group, reflecting that retransmission congestion and control overhead manifest most severely under lossy network conditions where the gap between reliability demands and available capacity is widest. 
Modeling studies are more evenly distributed across the three subgroups, indicating that the Time--State interaction has received comparatively more formal analytical attention, particularly in characterizing Heartbeat periodicity, retransmission dynamics, and executor contention. 
The key approach column reveals that mitigation strategies cluster around two structural directions.
Those that reduce state enforcement cost by relaxing consistency guarantees, including best-effort-tuning, time-filtering, order-relaxation, and policy-selection, and those that relocate state management responsibility away from the latency-critical path, including state-decoupling, lock-free, and TSN-replication. 
This distribution mirrors the structural conclusion common to all three subgroups: no proposed approach simultaneously preserves the immediacy of state recovery and the predictability of data delivery.

\subsection{Space--State Conflict}
While spatial abstraction aims to provide location transparency by shielding applications from physical placement, maintaining this independence either increases the structural cost of state management or makes state tracking structurally constrained under shared resource bounds.
Conversely, attempts to reduce state management overhead consistently converge toward restricting the scope of spatial abstraction.
This conflict forms a paradoxical escalation structure: it originates at the endpoint level, extends to transmission paths, and intensifies at the network level.
This subsection examines four manifestations of this tension: node anonymity that obscures lifecycle state, path-blind forwarding that conceals buffer conditions, the discovery storms that emerge from P2P state growth, and the structural vulnerabilities introduced by intermediary-dependent state management. \looseness=-1

\vspace{1em}
\subsubsection{Node Anonymity}
Location transparency in ROS~2 middleware is realized through data-centric anonymity: communication is organized around named topics and key expressions rather than physical node identities, so that receivers obtain data by declaring interest in a logical channel without any knowledge of who is producing it or from where.
This design is a deliberate structural principle that allows applications to remain indifferent to deployment topology and to interoperate across heterogeneous nodes without prior coordination~\cite{an2014content}.
The publish-subscribe model explicitly decouples data sources and data sinks with respect to node location and temporal requirements, ensuring that applications require no knowledge of the existence or location of other participating entities~\cite{calisi2010software}.
The same anonymity, however, imposes a structural constraint on state management. 
Because the middleware does not track the physical identity of communicating entities, it has no architectural basis for associating the persistent state accumulated during middleware operations with the specific physical entity that produced it, making contextual continuity dependent on an identity resolution capability that spatial abstraction structurally withholds. \looseness=-1

Under the anonymous endpoint model, the middleware has no architectural basis for distinguishing a rejoining node from a newly appeared one across deployment events.
Runtime endpoint identifiers such as the DDS GUID and the Zenoh session ID can distinguish a reconnecting endpoint from a first-time connection during the identifier's lifetime, but they do not survive redeployment or identifier change.
The DDS liveliness QoS policy governs whether an entity is still active by treating any entity that fails to assert its alive status within the configured lease duration as no longer active~\cite{omgdds1.4}.
Each participant therefore stores the information of all remote endpoints it has discovered in an internal database, and per-participant memory consumption grows linearly with the total number of endpoints in the domain~\cite{an2014content}.
Because DDS operates without a message broker, the burden of tracking which stored state belongs to which physical entity falls entirely on the individual participant~\cite{wytrkebowicz2021messaging}.

The system-level consequences of this identity ambiguity escalate along two distinct paths.
In the first path, ghost nodes that persist beyond their operational lifetime occupy endpoint slots in the discovery state and consume liveliness monitoring resources as the middleware continues issuing Heartbeat checks to unreachable addresses.
In some configurations, newly joining nodes with matching topic profiles receive stale cached state intended for the original terminated entity.
Prior work demonstrates that when a publisher is deleted without calling a disconnect service, the tracker continues to register the publisher as active, accumulating stale state that cannot be cleared without explicit lifecycle signaling~\cite{dust2022dynamic}.
In the second path, because the middleware associates endpoints with logical topic identifiers rather than persistent physical identities, a reconnecting node is indistinguishable from a newly arrived entity sharing the same configuration.
The middleware has no criterion for determining whether a reconnecting endpoint should receive previously accumulated lifecycle context or be treated as a new lifecycle context.
When both paths are active simultaneously, the system must manage an expanding ghost state while attempting to restore contextual continuity for reconnecting nodes.
It produces a state management burden whose cost grows with both system scale and the frequency of node lifecycle events. \enlargethispage{\baselineskip}

Several structural approaches address the node anonymity problem from different layers.
Assigning unique physical identifiers to each robot and creating dedicated per-robot communication channels through topic name suffixing, combined with explicit connection and disconnection service calls, provides a functional approach to dynamic participation management~\cite{dust2022dynamic}.
Purpose-built DDS-based middleware circumvent the anonymity-induced tracking gap through application-level identity enforcement, such as explicit join codes and periodic Heartbeats~\cite{asjad2024neuroreality}.
A Named Data Networking~(NDN) approach combined with a distributed query interface identifies data for specific key expressions regardless of which physical node originally produced it, allowing a reconnecting node to retrieve its prior state via a pull query without requiring identity resolution at the middleware layer. 
This addresses state recovery continuity rather than the underlying anonymity constraint, and Zenoh demonstrates the principle in practice~\cite{shih2022scalable}.
Combining ROS~2 managed nodes with external orchestrator health-check control loops such as Kubernetes binds logical component identities to physical process lifecycles and enables automatic restart management at the deployment level~\cite{toffetti2023ros}. \enlargethispage{\baselineskip}

Each approach resolves the identity-state coupling by partially dismantling the anonymity on which location transparency depends.
Storage-based decoupling requires the storage manager to explicitly track which endpoint produces which data and to keep its retained state aligned with that endpoint, a coupling that the fully anonymous publish-subscribe model was designed to render irrelevant~\cite{corsaro2023zenoh}.
Process-level lifecycle management requires external orchestrators to explicitly track deployment-level information that location transparency was designed to render irrelevant~\cite{toffetti2023ros}.
Prior work further establishes that in DDS architectures, state tracking cannot be fully decoupled from physical identity to maintain mutual consistency~\cite{zieba2006preservation}.
The structural outcome is consistent across all approaches: achieving lifecycle state continuity under dynamic node participation requires the system to acquire and use physical identity information that spatial abstraction withholds by design.

\vspace{1em}
\subsubsection{Path-Blind Forwarding}
Spatial abstraction in ROS~2 middleware structurally constrains the state that the middleware is permitted to maintain.
Because location transparency requires the routing path to be treated as an opaque conduit, the middleware holds no representation of downstream buffer occupancy, queue depth, or link capacity as part of its state.
This is not an implementation omission but a structural consequence of spatial abstraction.
The middleware accepts a message, serializes it, and injects it into the socket send buffer without consulting any lower-layer information about the physical path connecting sender to receiver~\cite{lee2025optimizing}.
The middleware's state is therefore spatially bounded, complete with respect to logical endpoint relationships, but blind to the physical infrastructure over which those relationships are realized.

When network load is low, this state incompleteness proceeds without visible consequence because downstream buffers drain faster than the middleware fills them.
Under saturation, however, the kernel socket buffer and switch queues accumulate backlogged packets at a rate that exceeds their service rate, causing subsequently injected packets to experience queuing delays independent of the physical propagation latency between sender and receiver.
Results show that when a wireless link outage occurs, ROS~2 applications and the DDS middleware continue accumulating unacknowledged samples in the HistoryCache until its capacity limit is reached~\cite{lee2025poster}.
Upon link recovery, the middleware releases the accumulated backlog in an instantaneous burst without regulating the transmission rate, while the application continues publishing new messages into the cache at its original rate, compounding the saturation~\cite{lee2025optimizing}.
Because the middleware maintains no state representing the physical path condition, it possesses no mechanism to detect this accumulation or moderate its injection behavior in response.

The escalation from localized buffer saturation to system-wide delivery degradation follows the same positive feedback structure identified in Section~\ref{subsubsec:3}.
CSMA/CA-based wireless channel contention at the data link layer causes backlog to accumulate in the software transmission queue, resulting in packet drops and a bottleneck shared across all flows on the affected interface~\cite{kim2026mitigating}.
This backlog propagates upward.
Because ROS~2 employs FIFO buffering, end-to-end latency increases linearly with backlog depth when the publisher injection rate exceeds the subscriber processing rate~\cite{teper2023timing}, so that a single saturation event corrupts the timing of an extended sequence of subsequent messages~\cite{dust2022dynamic}.
When multiple high-frequency topics share the same physical interface, the saturation of one topic's buffer path induces secondary saturation on co-located flows, spreading the impact of a single overload event across the entire communication graph~\cite{teper2022end}.

Several approaches have been proposed to mitigate the performance consequences of path-blind forwarding.
Socket buffer resizing offers a structurally simpler alternative by raising the saturation threshold, reducing the frequency of bloat events under transient load spikes without modifying the injection model~\cite{teper2022end}.
Local cache mechanisms discard redundant data before it enters the middleware buffer, and mathematical modeling of the relationship between deadline and depth QoS parameters enables balancing optimization that eliminates transmission rate and buffer size mismatches~\cite{jalil2022efficacy, jalil2023performance}.
Building on this, adaptive task execution rate regulation addresses the injection-consumption mismatch by analyzing message consumption capacity at runtime and dynamically adjusting the sensor message sampling and injection rate to match the actual processing bottleneck~\cite{li2025ater}.
Within the middleware, runtime queue occupancy monitoring with dynamic QoS reconfiguration, including queue depth and time-based filters, provides a self-adaptive framework for preventing buffer overflow~\cite{romero2013self}.
Configuring DDS history and resource limits QoS policies in proportion to the available link bandwidth prevents both buffer overflow and link saturation~\cite{lee2025dependency, lee2025optimizing}.
Congestion-aware transmission control introduces a feedback path from the network layer to the middleware's injection logic, where DDS tracks send window occupancy and Nack volume to actively modulate the publication rate under congestion~\cite{kang2020study}.

Each mitigation strategy achieves its benefit through a different degree of path state awareness, but all share the structural consequence of partially undermining the path independence that spatial abstraction provides.
Socket buffer resizing defers rather than eliminates saturation, addressing the consequence of blind forwarding without resolving its structural cause~\cite{blass2021automatic}.
Congestion-aware control and queue monitoring require the middleware to actively track physical forwarding state~\cite{kang2020study}.
Adaptive injection rate control binds transmission behavior to deployment-specific execution characteristics that location transparency was designed to render irrelevant~\cite{li2025ater}.
QoS-based resource bounding requires the operator to configure limits explicitly calibrated to the physical link's capacity~\cite{lee2025optimizing}.
The dimensional outcome is consistent: approaches that preserve path independence defer saturation rather than eliminate it, while those that substantively reduce forwarding overhead do so by reintroducing physical path awareness, thereby narrowing the scope of spatial abstraction.

\vspace{1em}
\subsubsection{Discovery Storm}
Maintaining location transparency requires the middleware to continuously construct and update a logical visibility graph in which every participating entity possesses awareness of every other entity's existence and metadata.
This full-mesh visibility is a structural prerequisite of location transparency, since without it the middleware cannot guarantee that newly joining entities will be incorporated into the communication graph without manual configuration.
Prior work establishes that DDS creates and maintains a fully connected graph by design, causing all participants, topics, and services to discover one another mutually and producing $O(N^{2})$ network traffic as a structural consequence~\cite{ros2_rmw_report_2023}.
The coupling between spatial scale and state management cost is therefore not a contingent implementation choice but a necessary consequence of the full mutual visibility that location transparency demands.

The total number of DDS discovery messages scales as $O(P^2)$ with participant count, with per-node transmissions on the order of $P \times E$, where $E$ is the endpoint count.
Memory consumption at each node is closely correlated with the number of endpoints it must store, forcing nodes in large networks to retain endpoint information entirely unrelated to their own communication context~\cite{sanchez2011bloom}.
In UAV swarm environments, the default discovery process generates substantial communication overhead that prevents reliable peer detection between vehicles, with network traffic growing nonlinearly with node count to the point where the discovery process itself can stall~\cite{lee2025optimizing_drones}.

The transition from elevated discovery traffic to system-wide determinism collapse follows a congestion-driven escalation path.
The initial burst of DDS discovery traffic during simultaneous node startup interacts with the software transmission queue's waiting-time threshold to cause severe network saturation and packet drops. This interaction has been formally modeled as the root cause of discovery-driven packet loss in multi-robot ROS~2 systems~\cite{kim2026mitigating}.
Experiments demonstrate that latency increases from approximately 5.5~ms at five nodes to 1309~ms at twenty nodes, and that at twenty-five nodes the packet loss rate reaches 94.98\%, rendering the network functionally unusable~\cite{castillo2024swarm}.
This escalation is particularly damaging for latency-sensitive robotic applications, because the periods of highest discovery load, such as system initialization and dynamic reconfiguration, coincide precisely with the moments when timely message delivery is most critical for system safety and coordination.

Several approaches have been proposed to reduce the discovery state management overhead and prevent discovery storm events.
Memory-efficient discovery optimization using extended threshold Bloom filter vectors transmits and stores only compact filter representations of participant endpoint descriptors rather than full metadata records, substantially reducing both memory consumption and network transmission volume~\cite{nwadiugwu2023mad}.
Domain segmentation partitions the participant space into chip-local and vehicle-area network segments, confining discovery traffic within each segment and using message bridges to forward data across boundaries, preventing cross-segment $O(N^{2})$ growth while preserving inter-segment communication~\cite{sang2025service}.
Hierarchical discovery structures replace the P2P model with a topology in which sessions register with a designated router; measurements show reductions in discovery overhead of 97\% to 99.9\% relative to P2P DDS implementations, with corresponding reductions in CPU utilization and latency~\cite{chovet2025performance}, and DDS deployments address the same scalability problem through optional discovery servers that similarly replace flat multicast discovery with a star topology~\cite{eprosima_discovery_server}.
Zenoh-DDS bridges in swarm testbeds achieve transparent ROS~2 connectivity while substantially lowering discovery traffic levels and eliminating the multicast dependency of flat DDS discovery~\cite{jones2022dots}.

Each mitigation strategy reduces discovery state cost by constraining the scope of mutual visibility, and in doing so partially relinquishes the full-mesh awareness on which location transparency depends.
Domain segmentation reduces per-domain discovery cost but introduces explicit topological boundaries that applications must account for, requiring deployment-level knowledge that location transparency was designed to render unnecessary.
Centralized discovery architectures achieve the most substantial scalability improvement but concentrate the entire system's visibility state in a single intermediary; prior work explicitly notes that if the intermediary fails, the entire system may fail to function~\cite{sanchez2011bloom, koksal2017obstacles}.
This structural consequence introduces a different conflict examined in the following subsection.
The structural outcome is consistent: no approach simultaneously preserves complete mutual visibility and eliminates $O(N^{2})$ discovery cost, and every viable reduction in state management overhead is obtained by accepting a corresponding reduction in the spatial scope of location transparency.

\vspace{1em}
\subsubsection{Intermediary Dependency}
The centralized discovery architectures introduced in the preceding subsection replace the flat P2P visibility graph with a hierarchical structure in which contexts no longer maintain direct mutual awareness.
Instead, each context registers its endpoint state with a central intermediary, such as a DDS Discovery Server or a Zenoh router, which assumes responsibility for matching and forwarding decisions~\cite{ros2_rmw_report_2023}.
This restructuring shifts visibility responsibility from each context to one or more intermediaries, reducing per-participant overhead while concentrating the discovery process at the intermediary tier. 
When the intermediary tier lacks redundancy, the loss of a router suspends new endpoint discovery and cross-host communication, even though already-established peer sessions on the same host continue to exchange data directly~\cite{desbiens2023building}.
Although this resolves the $O(N^{2})$ scalability problem, it introduces a structural dependency whose failure semantics are categorically different from those of the P2P model it replaces.
Notably, ROS~2 adopted P2P DDS discovery precisely to eliminate the single point of failure inherent in the ROS~1 centralized name server~\cite{macenski2022robot}, meaning that the centralization adopted for scalability reintroduces a single point of failure of the kind that P2P discovery was designed to avoid. \looseness=-1

The failure consequences of these two topologies differ not merely in degree but in kind.
In a flat P2P topology, the failure of any single node degrades the visibility graph only locally, where the affected node's endpoints become unreachable but all remaining participants retain mutual awareness and continue to communicate.
In a centralized topology, such as a single router with all applications in client mode, the broker mediates all communication between applications.
Its unavailability therefore halts both endpoint discovery and ongoing data exchange, since client-mode applications do not establish direct sessions with one another~\cite{portugal2025inquiring}.
Prior work argues that centralized approaches are fundamentally unsuitable for real-time and fault-tolerant applications, as concentrating the true values of all data objects on a single computer is architecturally unsound~\cite{pardo2003omg}.

The severity of this vulnerability is compounded by the operational position of the intermediary within the system.
The discovery intermediary handles the aggregated discovery traffic of all contexts and must process endpoint registration and matching requests at a rate that scales with system size, making it the node most exposed to sustained load.
As systems grow, maintaining availability and throughput requires deploying additional intermediaries, with cost, complexity, and latency all worsening proportionally~\cite{rti2014datacentric_misc}.
Achieving high availability further demands non-default redundancy mechanisms that lie outside the standard specification scope~\cite{wytrkebowicz2021messaging}.
In cloud robotics deployments, this structural constraint has been formally characterized through the LSC theorem, which proves that latency reliability, singleton deployment, and commodity infrastructure cannot be simultaneously achieved~\cite{chen2025fogros2}.

Several mechanisms have been proposed to mitigate this vulnerability.
Multi-level service discovery eliminates central dependency by partitioning communication into local and wide-area segments using message bridges, which decentralizes discovery and prevents any localized failures from compromising the entire system~\cite{sang2025service}.
Repository federation allows multiple centralized discovery servers to collaborate, effectively eliminating the single point of failure by ensuring system robustness even if the primary repository becomes unavailable~\cite{sanchez2011bloom}.
Redundant router deployment over independent network paths distributes failure risk across replicated intermediaries and allows participants to fall back to an available replica; this approach is supported by both Zenoh's routed topology~\cite{desbiens2023building} and DDS deployment configurations that combine multiple discovery servers~\cite{eprosima_discovery_server}.
Probabilistic latency-reliability models that replicate requests across independent network interfaces and cloud servers reduce communication failure probability exponentially, providing a quantitative framework for trading infrastructure cost against availability guarantees~\cite{chen2025fogros2}.

Each of these responses reveals the same structural cycle that connects this subsection to the preceding one.
Intermediary replication reduces failure risk but increases the complexity and resource cost of maintaining consistent state across replicas, partially negating the operational simplicity that centralization provided~\cite{wytrkebowicz2021messaging}.
Hybrid topologies that maintain limited cross-group P2P links bound the failure impact but reintroduce partial peer-to-peer state management within each group, reproducing a bounded version of the $O(N^2)$ cost that centralization was designed to eliminate~\cite{desbiens2023building}.
Automatic P2P fallback preserves communication continuity after intermediary failure but does so by reverting to precisely the scalability problem that motivated centralization~\cite{ros2_rmw_report_2023}.
This cycle constitutes a structural boundary of the present design space. 
Eliminating the single point of failure reintroduces distributed state cost, while eliminating distributed state cost reintroduces the single point of failure.
The structural outcome is consistent: no architecture within the current middleware design space resolves both constraints simultaneously~\cite{rti2014datacentric_misc}. \looseness=-1

\begin{table*}[t]
\centering
\caption{Taxonomy of Literature Addressing the Space--State Conflict}
\label{tab:ss conflict}
\resizebox{\textwidth}{!}{%
\begin{tabular}{|>{\centering\arraybackslash}m{0.7cm}|>{\centering\arraybackslash}m{1cm}|>{\centering\arraybackslash}m{2.5cm}|>{\centering\arraybackslash}m{2.8cm}|>{\centering\arraybackslash}m{0.7cm}|>{\centering\arraybackslash}m{0.7cm}|>{\centering\arraybackslash}m{0.7cm}|>{\centering\arraybackslash}m{0.7cm}|>{\centering\arraybackslash}m{0.7cm}|>{\centering\arraybackslash}m{0.7cm}|>{\centering\arraybackslash}m{2.5cm}|}
    \hline
    \multirow{2}{*}{\textbf{[Ref]}} & \multirow{2}{*}{\textbf{Protocol}} & \multirow{2}{*}{\textbf{Application Domain}} & \multirow{2}{*}{\textbf{Evaluation Platform}} & \multicolumn{3}{c|}{\textbf{Network}} & \multicolumn{3}{c|}{\textbf{Research Type}} & \multirow{2}{*}{\textbf{Key Approach}} \\
    \cline{5-10}
    & & & & \textbf{\shortstack{Wired}} & \textbf{\shortstack{Wire-\\less}} & \textbf{\shortstack{Local\\Host}} & \textbf{\shortstack{Evalu-\\ation}} & \textbf{\shortstack{System\\Design}} & \textbf{\shortstack{Model-\\ing}} & \\
    \hline
    \cite{chen2025fogros2} & \multirow{30}{*}{DDS} & \multirow{3}{*}{Robotics, Auto.} & Embedded, Workstation &  & $\checkmark$ &  & $\checkmark$ & $\checkmark$ &  & LSC-theorem \\
    \cline{1-1} \cline{4-11}
    \cite{teper2023timing} &  &  & Workstation, Simulator &  &  & $\checkmark$ & $\checkmark$ & $\checkmark$ & $\checkmark$ & Executor-modeling \\
    \cline{1-1} \cline{4-11}
    \cite{li2025ater} &  &  & Workstation &  &  & $\checkmark$ & $\checkmark$ & $\checkmark$ &  & Rate-adaptation \\
    \cline{1-1} \cline{3-11}
    \cite{koksal2017obstacles} &  & \multirow{2}{*}{Robotics, IIoT} & \multirow{2}{*}{-} &  &  &  &  & $\checkmark$ &  & Intermediary-risk \\
    \cline{1-1}
    \cite{rti2014datacentric_misc} &  &  &  &  &  &  &  & $\checkmark$ &  & Centrality-cost \\
    \cline{1-1} \cline{3-11}
    \cite{hu2024wireless} &  & \multirow{14}{*}{Robotics} & Embedded, Workstation, Simulator &  & $\checkmark$ &  & $\checkmark$ & $\checkmark$ &  & Sync-scaling \\
    \cline{1-1} \cline{4-11}
    \cite{jalil2022efficacy} &  &  & \multirow{2}{*}{Embedded, Workstation} &  & $\checkmark$ &  & $\checkmark$ & $\checkmark$ &  & Buffer-modeling \\
    \cline{1-1}
    \cite{jalil2023performance} &  &  &  &  & $\checkmark$ &  & $\checkmark$ & $\checkmark$ &  & Deadline-balancing \\
    \cline{1-1} \cline{4-11}
    \cite{blass2021automatic} &  &  & \multirow{2}{*}{Embedded} &  &  & $\checkmark$ & $\checkmark$ & $\checkmark$ &  & Path-blind \\
    \cline{1-1}
    \cite{castillo2024swarm} &  &  &  &  & $\checkmark$ &  & $\checkmark$ &  &  & Wireless-scaling \\
    \cline{1-1} \cline{4-11}
    \cite{dust2022dynamic} &  &  & \multirow{1}{*}{Simulator} &  &  & $\checkmark$ & $\checkmark$ & $\checkmark$ &  & Process-overhead \\
    \cline{1-1} \cline{4-11}    
    \cite{asjad2024neuroreality} &  &  & \multirow{7}{*}{Workstation} &  & $\checkmark$ &  & $\checkmark$ & $\checkmark$ &  & Identity-enforce \\
    \cline{1-1}
    \cite{kim2026mitigating} &  &  &  &  & $\checkmark$ &  & $\checkmark$ & $\checkmark$ & $\checkmark$ & Discovery-saturation \\
    \cline{1-1}
    \cite{lee2025dependency} &  &  &  &  & $\checkmark$ &  &  & $\checkmark$ & $\checkmark$ & Control-dependency \\
    \cline{1-1}
    \cite{lee2025optimizing} &  &  &  &  & $\checkmark$ &  & $\checkmark$ & $\checkmark$ &  & Burst-analysis \\
    \cline{1-1}
    \cite{lee2025poster} &  &  &  &  & $\checkmark$ &  & $\checkmark$ & $\checkmark$ &  & Burst-analysis \\
    \cline{1-1}
    \cite{macenski2022robot} &  &  &  &  &  & $\checkmark$ & $\checkmark$ &  &  & P2P-motivation \\
    \cline{1-1}
    \cite{romero2013self} &  &  &  &  &  & $\checkmark$ & $\checkmark$ & $\checkmark$ &  & Self-adaptive \\
    \cline{1-1} \cline{4-11}
    \cite{calisi2010software} &  &  & - &  &  &  &  & $\checkmark$ &  & Anonymity-model \\
    \cline{1-1} \cline{3-11}
    \cite{almadani2016qos} &  & \multirow{6}{*}{IIoT} & \multirow{4}{*}{Workstation} &  & $\checkmark$ &  & $\checkmark$ & $\checkmark$ &  & Time-filtering \\
    \cline{1-1}
    \cite{kang2020study} &  &  &  & $\checkmark$ &  &  & $\checkmark$ &  &  & AutoThrottle \\
    \cline{1-1}
    \cite{nwadiugwu2023mad} &  &  &  & $\checkmark$ &  &  & $\checkmark$ & $\checkmark$ & $\checkmark$ & Bloom-filter \\
    \cline{1-1}
    \cite{sanchez2011bloom} &  &  &  & $\checkmark$ &  &  & $\checkmark$ & $\checkmark$ &  & Bloom-filter \\
    \cline{1-1} \cline{4-11}
    \cite{pardo2003omg} &  &  & \multirow{2}{*}{-} &  &  &  &  & $\checkmark$ &  & Centrality-risk \\
    \cline{1-1}
    \cite{wytrkebowicz2021messaging} &  &  &  &  &  &  & $\checkmark$ &  &  & Broker-analysis \\
    \cline{1-1} \cline{3-11}
    \cite{sang2025service} &  & \multirow{2}{*}{Auto.} & Embedded &  &  & $\checkmark$ & $\checkmark$ & $\checkmark$ &  & Domain-segment \\
    \cline{1-1} \cline{4-11}
    \cite{teper2022end} &  &  & Workstation, Simulator &  &  & $\checkmark$ & $\checkmark$ &  & $\checkmark$ & Buffer-sizing \\
    \cline{1-1} \cline{3-11}
    \cite{an2014content} &  & \multirow{3}{*}{(None)} & Workstation & $\checkmark$ &  &  & $\checkmark$ & $\checkmark$ &  & Complexity-modeling \\
    \cline{1-1} \cline{4-11}
    \cite{omgdds1.4} &  &  & \multirow{2}{*}{-} &  &  &  &  & $\checkmark$ &  & Spec-analysis \\
    \cline{1-1}
    \cite{zieba2006preservation} &  &  &  &  &  &  &  & $\checkmark$ &  & Consistency-model \\
    \hline
    \cite{desbiens2023building} & Zenoh & Robotics, IIoT & - &  &  &  &  & $\checkmark$ &  & Router-federation \\
    \hline
    \cite{shih2022scalable} & \multirow{8}{*}{Both} & Robotics, IIoT, Auto. & Workstation &  &  & $\checkmark$ & $\checkmark$ & $\checkmark$ &  & Lock-free \\
    \cline{1-1} \cline{3-11}
    \cite{jones2022dots} &  & Robotics, IIoT & Embedded, Simulator &  & $\checkmark$ &  & $\checkmark$ & $\checkmark$ &  & Swarm-bridging \\
    \cline{1-1} \cline{3-11}
    \cite{lee2025optimizing_drones} &  & \multirow{4}{*}{Robotics} & Embedded, Simulator & $\checkmark$ & $\checkmark$ & $\checkmark$ & $\checkmark$ & $\checkmark$ &  & UAV-discovery \\
    \cline{1-1} \cline{4-11}
    \cite{chovet2025performance} &  &  & Embedded &  & $\checkmark$ &  & $\checkmark$ &  &  & Router-overhead \\
    \cline{1-1} \cline{4-11}
    \cite{portugal2025inquiring} &  &  & \multirow{2}{*}{-} &  &  &  & $\checkmark$ &  &  & Failure-semantics \\
    \cline{1-1}
    \cite{ros2_rmw_report_2023} &  &  &  &  &  &  &  & $\checkmark$ &  & P2P-cost \\
    \cline{1-1} \cline{3-11}
    \cite{toffetti2023ros} &  & IIoT & - &  &  &  &  & $\checkmark$ &  & Lifecycle-mgmt \\
    \cline{1-1} \cline{3-11}
    \cite{chisualițua2025stepping} &  & Auto. & Embedded, Workstation & $\checkmark$ & $\checkmark$ &  & $\checkmark$ &  &  & Protocol-comparison \\
    \hline
\end{tabular}
}
\end{table*}
\vspace{0.5em}
Table~\ref{tab:ss conflict} summarizes the 39 studies surveyed under the Space--State conflict, the largest of the three groups. 
Wireless evaluations dominate, consistent with the observation that node anonymity and path-blind forwarding manifest most severely under intermittent connectivity, where the gap between logical endpoint state and physical network reality is widest. 
A notable portion of entries lack a specified evaluation platform, concentrated among architectural analyses and specification-level studies that provide the structural foundation for the discovery storm and intermediary dependency subgroups. 
The key approach column reveals a mitigation landscape that is markedly more diverse than in the other two conflict pairs, spanning identity enforcement, path-aware congestion control, Bloom-filter-based discovery, domain segmentation, NDN-based querying, and router federation. 
This diversity is itself a structural observation: unlike the Space--Time conflict, where mitigations cluster around two recognizable categories, the Space--State conflict has not converged on a dominant architectural response. 
Every proposed approach addresses one aspect of the anonymity-state or scale-fragility cycle, confirming that the Space--State conflict has not been resolved within the current middleware design space.

\begin{figure}
    \centering
    \includegraphics[width=1\linewidth]{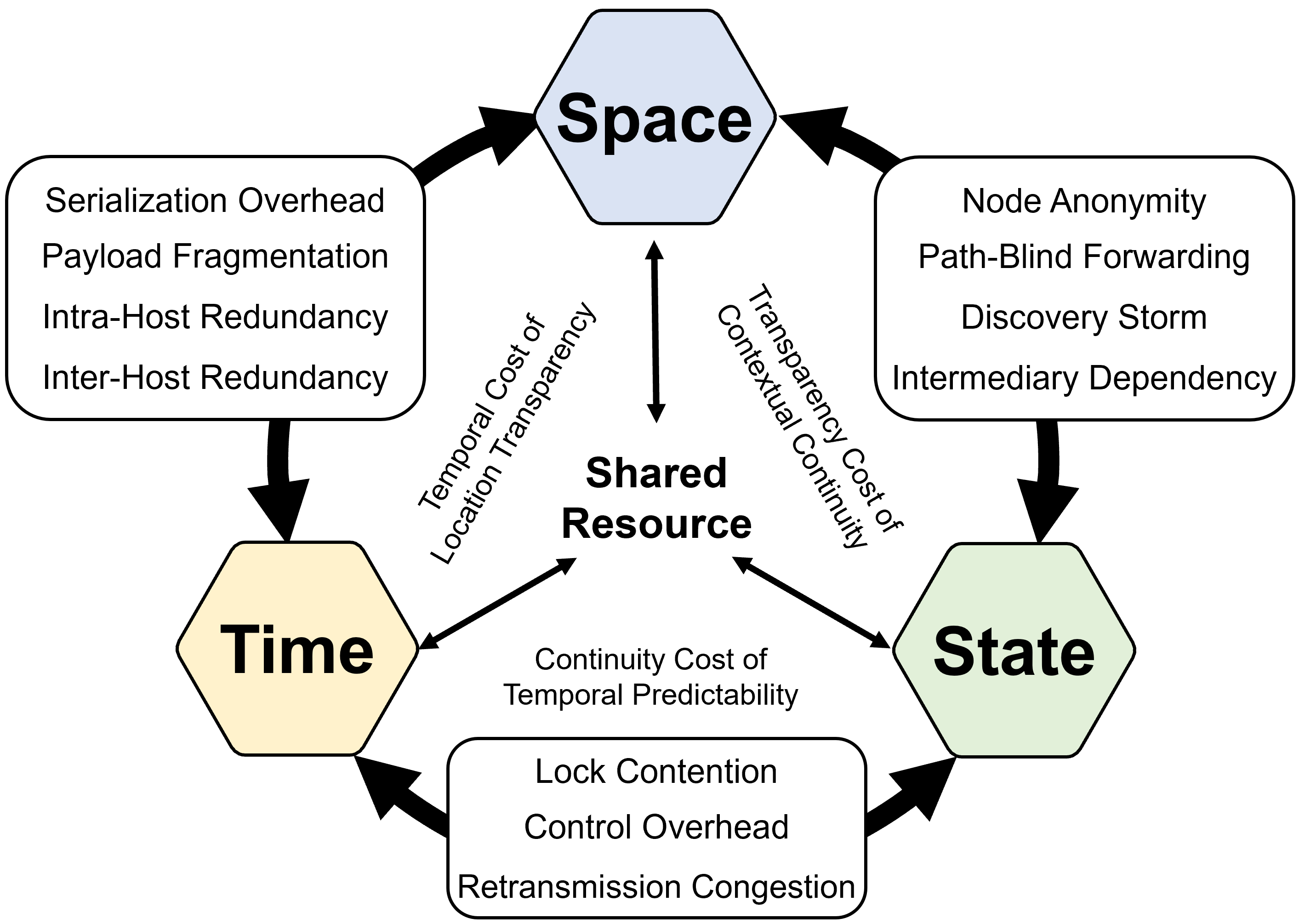}
    \caption{Structural Trade-off Cycle of ROS~2 Middleware}
    \label{fig:tradeoffs}
    \vspace{-1em}
\end{figure}
Fig.~\ref{fig:tradeoffs} synthesizes the structural trade-off cycle identified across the eleven manifestations analyzed in this section.
Each directed arc represents an unavoidable cost: location transparency imposes a temporal cost by concealing physical network state from the middleware, temporal predictability imposes a continuity cost by forcing reliability guarantees to be relaxed under real-time constraints, and contextual continuity imposes a transparency cost by requiring physical identity information that spatial abstraction structurally withholds.
Because all three mechanisms draw upon the same bounded shared resources, namely execution capacity, network bandwidth, and buffer memory, fully satisfying one dimension imposes pressure on at least one of the others.
No architecture within the current middleware design space resolves all three simultaneously; any proposed mitigation navigates the trade-off surface rather than escaping it.
This unresolved tension is not an artifact of implementation choices but a structural property of distributed robotic middleware under constrained deployment conditions.
Section~\ref{sec:roadmap} addresses this structural boundary at two levels: immediate gaps in current middleware practice that can be closed without architectural redesign, and long-term research directions that require rethinking the foundations of how \emph{Space}, \emph{Time}, and \emph{State} are jointly realized in distributed robotic systems. \looseness=-1

The central finding that emerges from this review is not that individual solutions are inadequate, but that every proposed mitigation trades one dimensional guarantee for another. Studies addressing the Space--Time conflict consistently demonstrate that reducing serialization cost, fragmentation loss, or multicast instability requires the middleware to reintroduce physical-layer awareness that location transparency was designed to conceal. 
Studies addressing the Time--State conflict reveal that the control traffic and retransmission mechanisms required to maintain contextual continuity consume the same execution and network resources that temporal predictability depends upon, and that relaxing reliability guarantees is the only operationally viable path to recovering bounded latency under congestion. 
Studies addressing the Space--State conflict show that the anonymity and full-mesh visibility on which location transparency relies simultaneously prevent lifecycle state from being tracked and drive discovery overhead that grows with system scale, while the centralized architectures introduced to contain that growth reintroduce a single point of failure of the kind that peer-to-peer discovery was designed to avoid.
Across all three pairs, the structural conclusion is the same: proposed solutions navigate the trade-off surface rather than escape it, and the boundaries of that surface are determined by the shared resource constraints under which all three dimensions must be simultaneously realized. 

\section{Structural Gaps and Research Roadmap}
\label{sec:roadmap}
The dimensional conflict analysis in Section~\ref{sec:tradeoffs} characterizes the trade-off surface that current middleware architectures must navigate.
Navigating this surface effectively, however, depends on resolving practical deficiencies that exist independently of the dimensional conflicts themselves.
This section addresses both concerns at two distinct levels.
The first level identifies structural gaps in current middleware design and operational practice whose resolution is a prerequisite for reliable deployment under realistic conditions.
The second level proposes long-term architectural directions that target the root causes of the dimensional conflicts identified in Section~\ref{sec:tradeoffs}. \looseness=-1

\subsection{Structural Gaps}
\label{sec:gaps}
This subsection identifies three structural gaps in current middleware practice, namely QoS semantic loss, evaluation inadequacy, and security overhead.
These gaps are not derived from the dimensional conflict analysis but represent independent deficiencies in how middleware is configured, evaluated, and secured in practice.
Each admits targeted improvements within the current design space, yet none has been comprehensively resolved in the existing literature.

\vspace{0.5em}
\subsubsection{QoS Semantic Loss}
QoS semantic loss refers to the degradation of QoS enforcement at the boundary between the middleware layer and the underlying OS and physical layer.
Temporal predictability has repeatedly been undermined because logical QoS configurations are not propagated into OS kernel scheduling or network-level priority mechanisms.
This enforcement failure produces priority inversions and unpredictable delivery latency under congested wireless conditions~\cite{kim2025cros, jaiswal2023wiros}.
In wide-area deployments, the absence of a standard mechanism for end-to-end QoS propagation further causes bandwidth and latency violations that the middleware cannot detect or compensate~\cite{hakiri2014supporting}.
This breakdown has been systematically identified as a major deployment obstacle in DDS-based systems~\cite{koksal2017obstacles}. 
Its recurrence across wireless deployments, wide-area networks, and real-time control scenarios confirms that it is not an isolated tuning problem but a structural gap in how QoS intent is realized across the communication stack. \looseness=-1

The structural origin of this gap lies in how the middleware maps application-level QoS parameters to physical transport mechanisms.
QoS configurations are treated as isolated per-endpoint attributes rather than cross-layer properties with dependency chains that span the execution, transport, and network layers~\cite{lee2025dependency}.
The strict architectural separation between the RMW interface and the OS prevents the middleware from directly influencing hardware acceleration pipelines or network scheduling queues, creating a systematic mismatch between logical intent and physical packet processing~\cite{scordino2022hardware}. 
This mismatch does not require a fundamental redesign of the core communication architecture to address, as the enforcement logic and the translation infrastructure surrounding it can be improved independently within the current design space~\cite{szabo2023toward, perez2015modeling}. \looseness=-1

Prior work has proposed modeling and validation approaches that reduce configuration errors and improve semantic consistency within the middleware layer.
Automated dependency validation and QoS capability profiling at design time can prevent semantic inconsistencies between component ports before execution begins~\cite{parra2021specifying, lee2025dependency}.
Domain-specific modeling languages and generative frameworks can synthesize semantically compatible policy configurations and eliminate trial-and-error tuning by producing correct-by-construction parameter sets prior to deployment~\cite{hoffert2007qos, an2013model}.
Mathematical optimization over conflicting QoS parameters such as publication rate and buffer depth provides a further tractable path to reducing packet loss and memory pressure without modifying the internal data exchange logic of the middleware~\cite{jalil2023qos}.
These contributions improve configuration correctness and reduce semantic inconsistency within the middleware boundary, but they do not resolve the underlying enforcement gap: QoS intent formulated at the application layer still fails to propagate into OS kernel scheduling or physical network priority mechanisms.

Narrowing this gap requires a targeted bridging component that connects middleware-level QoS semantics to lower-layer execution and scheduling mechanisms.
Rather than redesigning the middleware architecture, this component would either translate existing ROS~2 QoS parameters into a form that OS schedulers and network interface queues can natively interpret, or intercept middleware-level policy expressions and forward them to the appropriate lower-layer primitives such as Linux traffic control, CPU scheduling classes, or hardware offload queues.
Such a connector would allow temporal constraints formulated at the application layer to be enforced continuously down the communication stack without requiring changes to the RMW interface or the application programming model.
Developing and standardizing this bridging layer within the ROS~2 ecosystem represents a concrete and tractable near-term improvement that does not presuppose a fundamental redesign of the core middleware architecture.

\vspace{0.5em}
\subsubsection{Evaluation Inadequacy}
Evaluation inadequacy refers to the failure of current evaluation methodologies to capture the emergent degradation that arises under realistic distributed deployment conditions.
Isolated component benchmarks consistently fail to surface the latency fluctuations and network saturation that characterize operational robotic systems, because they rely on static payloads or uncongested traffic conditions that conceal the architectural costs of spatial abstraction~\cite{bode2023systematic, maruyama2016exploring, chen2019performance}.
Real-time scheduling and intra-process optimization studies conducted without realistic network contention produce characterizations that do not transfer to multi-host wireless deployments~\cite{fasano2025optimizing, gutierrez2018towards}.
This pattern is not attributable to individual experimental choices but reflects a structural deficiency in how middleware evaluation is currently scoped~\cite{park2020real, kluner2026modern}. \looseness=-1

The structural origin of this gap lies in the disjointed operational domains of physics-based robot simulators and discrete-event network simulators, which prevents the Perception--Action--Communication loop from being evaluated as an integrated whole.
Because the two simulation domains advance time independently and apply different abstraction models to shared events, the emergent interactions among discovery traffic, reliability retransmission, and application data exchange cannot be faithfully reproduced within either domain alone.
This mismatch does not require a redesign of the ROS~2 middleware architecture to address, as the evaluation infrastructure surrounding the middleware can be improved independently within the current design space~\cite{dey2023novel, richart2022cocosim, acharya2020cornet, moon2020gazebo}. \enlargethispage{\baselineskip}

Prior work has proposed co-simulation and domain adaptation approaches that improve evaluation fidelity without modifying the middleware itself.
Synchronized co-simulation frameworks enforce consistent temporal boundaries across heterogeneous simulation components, and modular wrappers around existing robotic and network simulators substantially improve the fidelity of emulated traffic behavior.
Deep learning-based domain adaptation provides a tractable path to sim-to-real transfer by aligning estimated network parameters with observed physical behavior, reducing the divergence between simulated evaluation results and operational outcomes~\cite{shi2023adapting, agarwal2025sim}.
These contributions improve evaluation fidelity within individual simulation domains, but they do not resolve the underlying integration gap: the emergent interactions among discovery traffic, reliability retransmission, and application data exchange still cannot be characterized as a unified whole under realistic scale and contention. \enlargethispage{\baselineskip}

Narrowing this gap requires a targeted integration of robotic and network simulation environments into unified testbeds that reflect the scale and contention characteristics of physical deployments.
Open multi-robot testbeds capable of exercising publish-subscribe middleware under varying traffic densities and node counts would allow the scalability boundaries imposed by discovery storms and control traffic overhead to be quantified before physical deployment~\cite{jones2022dots, kang2020study, calvo2021ros}.
Memory-efficient discovery protocols and filtered state propagation mechanisms should be systematically included in these environments to ensure that state management costs are characterized alongside data exchange performance~\cite{sanchez2011bloom, nwadiugwu2023mad}.
Developing and standardizing such integrated evaluation infrastructure within the ROS~2 ecosystem represents a concrete and tractable near-term improvement that does not presuppose a fundamental redesign of the core middleware architecture.

\vspace{0.5em}
\subsubsection{Security Overhead}
Security overhead refers to the structural incompatibility between security enforcement and temporal predictability under realistic deployment conditions.
Enabling mandatory security plugins, including encryption and authentication, introduces latency and throughput degradation that directly competes with time-critical data exchange~\cite{kim2018security, sandoval2019cyber}.
Heterogeneous implementations further exhibit interoperability failures and disproportionate computational overhead when strict security policies are enforced at the transport boundary~\cite{aartsen2022analyzing, takemoto2019performance}.
Cryptographic operations do not impose a fixed additive cost but instead interact with the same execution and network resources that temporal predictability depends upon, compounding timing nondeterminism under load~\cite{xia2025investigating}.

The structural origin of this gap lies in the decoupling of security verification from the middleware's core discovery and state management dynamics.
Unprotected discovery services and software-level policy evaluation expose the system to replay attacks and identity spoofing, and the resulting need for redundant cryptographic checks amplifies resource consumption at precisely the moments when discovery and liveliness traffic are already competing for shared capacity~\cite{wang2024formal, du2022formal}.
The disjointed management of security artifacts further allows adversarial entities to exploit the gap between logical permissions and physical execution, bypassing access controls through the same anonymity mechanisms that location transparency relies upon~\cite{deng2022security, patel2022analyzing}.
This mismatch does not require a redesign of the spatial abstraction layer to address, as the overhead originates in how cryptographic operations are integrated into the execution path rather than in the communication model itself, and the surrounding security infrastructure can be improved independently within the current design space~\cite{david2013dds}.

Prior work has proposed policy provisioning and cryptographic execution approaches that reduce overhead without modifying the core middleware architecture.
Procedurally generated access control policies can eliminate manual misconfigurations and reduce the computational footprint of authorization checks during endpoint matching, without altering the QoS or discovery mechanisms that govern communication behavior~\cite{white2018procedurally}.
Delegating cryptographic operations to trusted execution environments isolates verification workloads from the primary communication pipeline, preventing security enforcement from consuming the execution resources required by latency-sensitive callbacks~\cite{wang2024support}.
Trusted Platform Module~(TPM)-based remote attestation integrated into the existing handshake protocol further provides continuous hardware-backed validation of participant identity, structurally preventing spoofing while preserving the location transparency on which spatial abstraction depends~\cite{wagner2024dds}.
At the architectural level, Zenoh adopts a boundary-security model where security is enabled selectively at chosen boundaries rather than uniformly across all communication, so that the throughput and latency cost of cryptographic operations applies only to traffic that requires it~\cite{corsaro2023zenoh}.
These contributions reduce the per-operation cost of security enforcement, but they do not resolve the underlying coupling gap: cryptographic verification remains structurally entangled with the same execution and network resources that discovery, liveliness maintenance, and data exchange depend upon.

Narrowing this gap requires a targeted mechanism that structurally separates security verification from the latency-critical communication path.
Rather than redesigning the middleware architecture, this mechanism would intercept authentication and attestation operations and delegate them to a dedicated hardware-backed execution context, so that cryptographic overhead no longer competes with time-sensitive callbacks on shared execution resources.
Such a separation would allow security guarantees to be maintained continuously without consuming the bounded execution capacity that real-time data exchange depends upon.
Developing and standardizing this separation mechanism within the ROS~2 security architecture represents a concrete and tractable near-term improvement that does not presuppose a fundamental redesign of the core middleware architecture. \looseness=-1

\vspace{0.5em}
Although each gap arises in a distinct operational domain, they are not independent.
QoS semantic loss prevents middleware-level policies from being faithfully enforced at the transport and OS layers, and evaluation inadequacy conceals the emergent consequences of this enforcement failure under realistic deployment conditions.
Evaluation inadequacy further prevents security overhead from being accurately characterized, because isolated benchmarks conducted without realistic traffic contention cannot reproduce the timing nondeterminism that cryptographic operations introduce under load.
Security overhead compounds both by introducing additional resource contention that interacts with the same execution and network resources that temporal predictability depends upon.
Addressing any one of these gaps in isolation therefore provides incomplete relief; progress toward robust middleware deployment requires that all three be pursued in parallel.

\subsection{Research Roadmap}
The research directions presented in this section target the root cause of the dimensional conflicts, namely the shared execution boundary and structural coupling among the three dimensions that render their simultaneous satisfaction structurally constrained under shared resource bounds.
Each direction directly addresses a specific structural conflict identified in Section~\ref{sec:tradeoffs}, and together they outline a long-term trajectory toward middleware architectures that renegotiate, rather than merely navigate, the trade-off surface defined by \emph{Space}, \emph{Time}, and \emph{State}. \looseness=-1

\vspace{0.5em}
\subsubsection{Cross-Layer Co-Design}
The ROS~2 communication stack is hierarchically structured across user space, RCL, RMW, middleware, and the kernel OS, with information flowing strictly in one direction~\cite{ros2_internal_interfaces}.
Upper layers issue configurations and commands to lower layers, but the runtime state of these underlying components is never fed back to the application level.
The middleware therefore remains unaware of OS-level scheduling dynamics and physical network conditions, while applications cannot observe internal middleware states~\cite{sciangula2025end, yu2026ros2probe}.
This structural unidirectionality is the fundamental source of the Space--Time conflict, as QoS parameters lose their semantic enforcement across these decoupled boundaries~\cite{hakiri2013supportinginternet}.

Addressing this limitation requires a cross-layer co-design approach that enables bidirectional interaction across the communication stack~\cite{almadani2017qos}.
The core of this approach is the bidirectional integration between the RCL and the middleware via the RMW, so that middleware states are dynamically propagated upward while the execution context of the RCL actively shapes middleware behavior.
The interaction between the kernel OS and the middleware must rely on indirect state estimation rather than direct coupling~\cite{zhao2026distributed}.
This strategy allows the middleware to recognize physical infrastructure constraints without compromising the location transparency that the Space dimension requires~\cite{bosk2024towards}.
This direction focuses specifically on the upward surfacing of middleware states and the indirect estimation of lower-layer conditions.

Several open problems must be addressed before this bidirectional paradigm can be realized in operational robotic middleware.
Expanding the RMW interface for bidirectional state exchange must preserve backward compatibility with established application programming models~\cite{hoffert2008dqml}.
Estimation mechanisms that infer kernel OS conditions without directly exposing low-level system states require principled design and standardization before they can be deployed reliably across heterogeneous platforms.
The continuous monitoring and state-sharing processes inherent to bidirectional interaction impose additional execution and network load, and robust methodologies are needed to ensure that this overhead does not itself degrade the temporal predictability of time-critical control loops~\cite{hoffert2011timely}.
Resolving these problems is a prerequisite for middleware infrastructure that structurally closes the unidirectional information gap across the ROS~2 communication stack. \looseness=-1

The fundamental difficulty of cross-layer co-design lies in the fact that the unidirectionality it seeks to overcome is not an accidental design choice but a deliberate architectural boundary that preserves modularity, portability, and vendor independence across the ROS~2 ecosystem.
Introducing upward state propagation through the RMW interface risks exposing middleware-internal representations to the RCL, creating implicit coupling between layers that the RMW abstraction was specifically designed to prevent.
Similarly, indirect OS state estimation requires the middleware to maintain persistent runtime models of lower-layer behavior, which must remain accurate across heterogeneous hardware platforms, OS configurations, and network interface drivers without direct access to the state being estimated.
The tension between the architectural openness required for cross-layer feedback and the strict boundary enforcement required for middleware portability means that any realization of this direction must resolve a structural conflict that is internal to the ROS~2 design philosophy itself.

\vspace{0.5em}
\subsubsection{Adaptive Middleware}
The structural conflict between the Space and Time dimensions isolates logical endpoint abstractions from physical network states, preventing the middleware execution layer from perceiving underlying link fluctuations and congestion.
Complete decoupling of the application from the deployment topology removes the opportunity for the middleware to adapt its transmission behavior to deteriorating network conditions~\cite{garcia2009topology}.
Attempts to resolve these inefficiencies exclusively at the application layer are structurally insufficient, as the application lacks visibility into lower-level network conditions~\cite{szabo2025enhancing}.
This rigid separation leaves distributed components unable to reliably cope with the intermittent connectivity and fluctuating resource availability characteristic of dynamic edge environments~\cite{orsini2016cloudaware}.

Addressing this limitation without dismantling the benefits of spatial abstraction requires an adaptive strategy that captures physical constraints with minimal latency and computational overhead.
Rather than directly exposing raw physical network states to the execution layer, adaptive middleware must continuously estimate underlying link capacity using observable, lightweight indicators~\cite{bellavista2008context}.
Monitoring accessible endpoint metrics such as acknowledgment timing, message delivery age, and internal queue delays allows the system to indirectly infer physical constraints and congestion boundaries without violating location transparency.
Based on these context-aware estimations, the middleware can adapt its transmission behavior and failover mechanisms to sustain temporal predictability without introducing rigid cross-layer dependencies~\cite{balasubramanian2009adaptive}.

Several open problems must be resolved before adaptive middleware can be operationalized in distributed robotic deployments.
Achieving a deterministic integration of fault-tolerance mechanisms with real-time scheduling remains challenging, as dynamic recovery procedures frequently disrupt strict temporal deadlines~\cite{balasubramanian2008towards}.
Formalizing consistency models that tolerate bounded state divergence without destabilizing closed-loop control remains a further necessity, particularly for deployments where intermittent connectivity prevents liveliness and history restoration mechanisms from operating under their design assumptions~\cite{pitoura2002data}.
Resolving these problems requires novel cross-dimensional algorithms that quantitatively model the trade-offs among latency, replication overhead, and network topology, establishing a principled foundation for middleware that adapts across the \emph{Space}, \emph{Time}, and \emph{State} dimensions simultaneously rather than optimizing each in isolation.

The core difficulty of adaptive middleware lies in the constraint that physical inference must be performed using only information that is already visible at the middleware boundary, without penetrating the spatial abstraction that the middleware is obligated to preserve.
Observable endpoint metrics such as acknowledgment timing and queue delays are indirect proxies that reflect physical conditions only under specific traffic patterns, and their predictive accuracy degrades precisely under the congestion and link variability conditions where adaptation is most urgently needed.
Furthermore, the adaptation logic itself consumes execution resources on the same threads and queues that carry time-critical callbacks, meaning that the overhead of estimating and responding to physical constraints competes directly with the temporal predictability it is intended to protect.
Any adaptive strategy must therefore operate within a narrow execution budget that shrinks as network conditions deteriorate, imposing a self-limiting constraint on the sophistication of the inference and adaptation mechanisms that can be practically deployed. \looseness=-1
 
\vspace{0.5em}
\subsubsection{Decoupled State Architecture}
The Space--State conflict arises because discovery and state management mechanisms are structurally coupled to the same resources required for temporal predictability, preventing spatial scalability and state continuity from being simultaneously sustained within the current architecture~\cite{paul2024performance, yoshino2025high}.
Fully meshed peer-to-peer topologies carry an $O(N^2)$ metadata exchange cost, while centralized architectures concentrate the state management burden at a single point of failure~\cite{kurte2024decentralised, ebadi2011new}.
The continuous metadata exchange in distributed topologies directly competes with latency-sensitive data transmission, whereas centralized intermediaries introduce potential network bottlenecks~\cite{detzner2023sola}.
This architectural dichotomy establishes a structural tension that remains unresolved within the current middleware design space~\cite{jung2008efficient}. \looseness=-1

Addressing this tension without permanently committing to either extreme requires an adaptive hybrid state architecture that maintains a centralized structure for baseline efficiency while falling back to decentralized mechanisms under varying failure and scaling conditions~\cite{harbird2004adaptive, mokadem2010resource}.
During stable operation, this approach exploits the low computational and network overhead of centralized brokers, mitigating the discovery traffic typically associated with dynamic node participation~\cite{tang2019design, gomes2015federated}.
When the system encounters intermediary broker failures, severe network partitions, or abrupt surges in node density, the architecture transitions to a peer-to-peer mechanism to preserve contextual continuity~\cite{david2013dds}.
This structural flexibility allows the middleware to combine the spatial efficiency of hierarchical control with the fault tolerance of distributed state retention~\cite{kim2025dynamic, ros2_domain_bridge_design}. \looseness=-1

Several open problems must be addressed before a decoupled state architecture can be operationalized in distributed robotic deployments.
Minimizing transition costs and establishing precise criteria for triggering the fallback from centralized to distributed modes remain open design questions~\cite{song2024ros, sim2021data}.
Reintegrating the distributed state accumulated during the fallback phase into a unified centralized structure upon recovery requires mechanisms that prevent data conflicts and logical inconsistencies~\cite{boari2008middleware, aragao2016middleware}.
Rapid structural shifts between the two operational modes must not disrupt or delay time-critical communication sessions already in progress across partitioned domains~\cite{zenoh_bridge_dds_doc}.
Resolving these problems requires deeper integration of dynamic reconfiguration logic with underlying transport interfaces, so that location transparency is preserved across volatile network boundaries~\cite{song2020integration}.

Recent developments illustrate partial progress in this direction. 
Zenoh introduces a Regions architecture that replaces the fixed router, peer, and client tiers with arbitrary-depth nested topologies and operator-defined gateways, structurally addressing the static-topology dimension of the problem~\cite{zenoh_longwang_2026}. 
The dynamic case that this direction targets remains open, since runtime transition between centralized and peer-to-peer operation under broker failure or severe partition still triggers a discovery burst from a state of mutual ignorance, and the divergent state accumulated during the fallback interval must still be reconciled on recovery. 
These are open challenges across distributed-systems research, not unique to any single middleware.

\vspace{1em}
The core difficulty of the hybrid fallback architecture lies in the asymmetry between the two operational modes it must support.
Centralized operation accumulates a globally consistent state view at the broker, but this view is not replicated across participants, meaning that a fallback to peer-to-peer mode begins from a state of mutual ignorance among nodes that have never maintained direct visibility of one another.
Reconstructing sufficient mutual awareness to sustain communication during the fallback phase requires a burst of discovery traffic at precisely the moment when network conditions are most stressed, reproducing a bounded version of the discovery storm dynamics identified in Section~\ref{sec:tradeoffs}.
The transition back to centralized operation introduces a symmetric problem: the broker must reconcile the divergent state accumulated across distributed participants during the fallback interval, and the cost of this reconciliation scales with both the duration of the fallback and the rate of topology change during that period.

Taken together, these three directions represent a qualitative departure from the mitigation strategies surveyed in Section~\ref{sec:tradeoffs}.
Where existing approaches improved one dimension by partially relinquishing another, each roadmap direction attempts to restructure the underlying mechanism so that both dimensions of the targeted conflict can be addressed without forcing an explicit sacrifice.
Cross-layer co-design pursues bidirectional state visibility without dismantling the modularity boundaries that spatial abstraction depends upon.
Adaptive middleware pursues physical constraint awareness without exposing deployment topology to the application layer.
Decoupled state architecture pursues the efficiency of centralization without permanently surrendering the fault tolerance of distributed operation.
None of these directions eliminates the structural tension identified in Section~\ref{sec:tradeoffs}; the shared resource bounds that produce dimensional conflict remain a physical reality.
What they collectively offer is a principled basis for operating closer to the boundary of what those resource bounds permit, rather than accepting the conservative compromises that characterize current deployments.
Realizing this potential requires sustained research investment across middleware design, OS integration, and distributed systems theory, guided by the evaluation infrastructure and cross-layer modeling tools.

\section{Conclusion}
\label{sec:conclusion}
This paper has presented a systematic survey of ROS~2 middleware through the lens of three structural dimensions, namely \emph{Space}, \emph{Time}, and \emph{State}.
We first analyzed the architectural constructs and operational dynamics of ROS~2 middleware, encompassing discovery, data exchange, and state management mechanisms across DDS and Zenoh.
Building on this foundation, we formalized \emph{Space} as the requirement for location transparency, \emph{Time} as the requirement for temporal predictability, and \emph{State} as the requirement for contextual continuity.
Through a comprehensive review of 94 middleware-focused studies, we mapped the structural trade-offs among these dimensions into three conflict pairs, Space--Time, Time--State, and Space--State, and identified eleven distinct manifestations through which these conflicts propagate across middleware, OS, and physical layers.

A central finding of this survey is that there is no silver bullet for robotic middleware design.
Every mitigation strategy examined across the eleven manifestations improves one dimension by partially relinquishing the guarantees of another.
Location transparency imposes a cost on temporal performance by concealing physical network state from the middleware.
Temporal predictability can only be preserved by relaxing the reliability guarantees that contextual continuity depends upon.
Contextual continuity, in turn, demands physical identity information that spatial abstraction is designed to withhold.
Because all three dimensions compete for the same bounded shared resources, namely execution capacity, network bandwidth, and buffer memory, satisfying any one dimension fully exerts pressure on at least one of the others, and any proposed mitigation navigates the trade-off surface rather than escaping it.
Middleware designers and practitioners must therefore accept that deploying a robotic system always entails an explicit or implicit choice of which dimensional guarantee to prioritize and which to partially sacrifice, and this choice should be made deliberately rather than by default.

Looking ahead, the primary research challenge is not to escape the trade-off surface but to navigate it more efficiently, by designing middleware architectures that maximize the achievable guarantees across all three dimensions simultaneously within the constraints imposed by shared resources.
The roadmap directions proposed in this paper, namely cross-layer co-design, adaptive middleware, and state architecture, each represent a principled step toward shrinking the region of the trade-off space in which current systems operate, rather than merely shifting cost from one dimension to another.
This challenge is acquiring renewed urgency in the era of Physical AI, where vision-language-action models, embodied agents, and large-scale multi-robot coordination are driving an unprecedented increase in the volume, heterogeneity, and latency-sensitivity of data exchanged over robotic networks.
As the communication demands of intelligent robotic systems continue to expand, the architectural choices embedded in middleware design will increasingly determine whether distributed robotic platforms can operate reliably and responsively in the open-world environments they are expected to inhabit.

\bibliographystyle{IEEEtran}
\bibliography{main}

@inproceedings{kronauer2021latency,
  title={Latency analysis of {ROS} 2 multi-node systems},
  author={Kronauer, Tobias and Pohlmann, Joshwa and Matth{\'e}, Maximilian and Smejkal, Till and Fettweis, Gerhard},
  booktitle={Proc. 2021 {IEEE} Int. Conf. on Multisensor Fusion and Integration for Intelligent Systems ({MFI})},
  pages={1--7},
  year={2021},
  publisher={IEEE}
}

@article{jung2025open,
  title={Open-source autonomous driving software platforms: Comparison of {Autoware} and {Apollo}},
  author={Jung, Hee-Yang and Paek, Dong-Hee and Kong, Seung-Hyun},
  journal={arXiv preprint arXiv:2501.18942},
  year={2025}
}

@inproceedings{peeck2021middleware,
  title={A middleware protocol for time-critical wireless communication of large data samples},
  author={Peeck, Jonas and M{\"o}stl, Mischa and Ishigooka, Tasuku and Ernst, Rolf},
  booktitle={Proc. 2021 {IEEE} Real-Time Systems Symp. ({RTSS})},
  pages={1--13},
  year={2021},
  publisher={IEEE}
}

@inproceedings{kluner2025automotive,
  title={Automotive middleware performance: Comparison of {FastDDS}, {Zenoh} and {vSomeIP}},
  author={Kl{\"u}ner, David Philipp and Hegerath, Lucas and Hatib, Amin Dieter and Kowalewski, Stefan and Alrifaee, Bassam and Kampmann, Alexandru},
  booktitle={Proc. 2025 {IEEE} Int. Conf. on Vehicular Electronics and Safety ({ICVES})},
  year={2025},
  publisher={IEEE}
}

@inproceedings{zhang2022distributed,
  title={Distributed robotic systems in the edge-cloud continuum with {ROS} 2: A review on novel architectures and technology readiness},
  author={Zhang, Jiaqiang and Keramat, Farhad and Yu, Xianjia and Hern{\'a}ndez, Daniel Montero and Queralta, Jorge Pe{\~n}a and Westerlund, Tomi},
  booktitle={Proc. 2022 Seventh Int. Conf. on Fog and Mobile Edge Computing ({FMEC})},
  pages={1--8},
  year={2022},
  publisher={IEEE}
}

@article{lee2025probabilistic,
  title={Probabilistic latency analysis of the {Data Distribution Service} in {ROS} 2},
  author={Lee, Sanghoon and Park, Hyung-Seok and Chae, Jiyeong and Park, Kyung-Joon},
  journal={arXiv preprint arXiv:2508.10413},
  year={2025}
}

@article{lee2025optimizing,
  title={Optimizing {ROS} 2 communication for wireless robotic systems},
  author={Lee, Sanghoon and Kim, Taehun and Chae, Jiyeong and Park, Kyung-Joon},
  journal={arXiv preprint arXiv:2508.11366},
  year={2025}
}

@inproceedings{lee2025poster,
  title={Poster: How to send large data in {ROS} 2},
  author={Lee, Sanghoon and Kim, Taehun and Chae, Jiyeong and Park, Kyung-Joon},
  booktitle={Proc. 2025 {IEEE} 33rd Int. Conf. on Network Protocols ({ICNP})},
  pages={1--3},
  year={2025},
  publisher={IEEE}
}

@inproceedings{sang2025service,
  title={Service discovery-based hybrid network middleware for efficient communication in distributed robotic systems},
  author={Sang, Shiyao and Ling, Yinggang},
  booktitle={Proc. 2025 {IEEE/RSJ} Int. Conf. on Intelligent Robots and Systems ({IROS})},
  pages={1357--1364},
  year={2025},
  publisher={IEEE}
}

@inproceedings{pohnl2022middleware,
  title={A middleware journey from microcontrollers to microprocessors},
  author={P{\"o}hnl, Michael and Tamisier, Alban and Blass, Tobias},
  booktitle={Proc. 2022 Design, Automation \& Test in Europe Conf. \& Exhibition ({DATE})},
  pages={282--286},
  year={2022},
  publisher={IEEE}
}

@inproceedings{kang2020study,
  title={A study of publish/subscribe middleware under different {IoT} traffic conditions},
  author={Kang, Zhuangwei and Canady, Robert and Dubey, Abhishek and Gokhale, Aniruddha and Shekhar, Shashank and Sedlacek, Matous},
  booktitle={Proc. Int. Workshop on Middleware and Applications for the Internet of Things},
  pages={7--12},
  year={2020}
}

@inproceedings{kang2021comprehensive,
  title={A comprehensive performance evaluation of different {Kubernetes} {CNI} plugins for edge-based and containerized publish/subscribe applications},
  author={Kang, Zhuangwei and An, Kyoungho and Gokhale, Aniruddha and Pazandak, Paul},
  booktitle={Proc. 2021 {IEEE} Int. Conf. on Cloud Engineering ({IC2E})},
  pages={31--42},
  year={2021},
  publisher={IEEE}
}

@inproceedings{peeroo2022exploring,
  title={Exploring the effects of multicast communication on {DDS} performance},
  author={Peeroo, Kaleem and Popov, Peter and Stankovic, Vladimir},
  booktitle={Proc. Student Forum of the 18th European Dependable Computing Conf. ({EDCC})},
  year={2022}
}

@techreport{ros2_rmw_report_2023,
  author={{Open Robotics}},
  title={{ROS} 2 alternative middleware report},
  institution={Open Robotics},
  year={2023},
  month={Sep.},
  url={https://discourse.ros.org/t/ros-2-alternative-middleware-report/33771},
  note={Accessed: 2026-03-11}
}

@article{diluoffo2018robot,
  title={{Robot Operating System} 2: The need for a holistic security approach to robotic architectures},
  author={DiLuoffo, Vincenzo and Michalson, William R and Sunar, Berk},
  journal={Int. J. of Advanced Robotic Systems},
  volume={15},
  number={3},
  pages={1--11},
  year={2018},
  publisher={SAGE Publications}
}

@inproceedings{fu2025iort,
  title={{IoRT} {ROS} 2 applications: Evaluating {Zenoh} and {VPN} for robotic networking in the edge-cloud continuum},
  author={Fu, Lei and Kapoor, Gaurav and Militano, Leonardo and Carughi, Giovanni Toffetti and Bohnert, Thomas Michael},
  booktitle={Proc. 2025 {IEEE} Symp. on Computers and Communications ({ISCC})},
  pages={1--6},
  year={2025},
  publisher={IEEE}
}

@inproceedings{park2025analytical,
  title={An analytical latency model of the {Data Distribution Service} in {ROS} 2},
  author={Park, Hyung-Seok and Lee, Sanghoon and Um, Doosik and Ryu, Hyunho and Park, Kyung-Joon},
  booktitle={Proc. {IEEE} {INFOCOM} 2025 -- {IEEE} Conf. on Computer Communications},
  pages={1--10},
  year={2025},
  publisher={IEEE}
}

@article{zhang2024comparison,
  title={Comparison of middlewares in edge-to-edge and edge-to-cloud communication for distributed {ROS} 2 systems},
  author={Zhang, Jiaqiang and Yu, Xianjia and Ha, Sier and Pe{\~n}a Queralta, Jorge and Westerlund, Tomi},
  journal={J. of Intelligent \& Robotic Systems},
  volume={110},
  number={4},
  pages={1--20},
  year={2024},
  publisher={Springer}
}

@article{liang2023performance,
  title={A performance study on the throughput and latency of {Zenoh}, {MQTT}, {Kafka}, and {DDS}},
  author={Liang, Wen-Yew and Yuan, Yuyuan and Lin, Hsiang-Jui},
  journal={arXiv preprint arXiv:2303.09419},
  year={2023}
}

@article{lee2025optimizing_drones,
  title={Optimizing {Data Distribution Service} discovery for swarm unmanned aerial vehicles through preloading and network awareness},
  author={Lee, HyeonGyu and Kim, Doyoon and Moon, SungTae},
  journal={Drones},
  volume={9},
  number={8},
  pages={1--20},
  year={2025},
  publisher={MDPI}
}

@phdthesis{kim2026mitigating,
  title={Mitigating {DDS} discovery bursts in {ROS} 2 multi-robot system},
  author={Kim, Taehyun and others},
  year={2026},
  school={DGIST}
}

@inproceedings{pardo2003omg,
  title={{OMG} {Data Distribution Service}: Architectural overview},
  author={Pardo-Castellote, Gerardo},
  booktitle={Proc. 23rd Int. Conf. on Distributed Computing Systems Workshops},
  pages={200--206},
  year={2003},
  publisher={IEEE}
}

@article{castillo2024swarm,
  title={Swarm robot communications in {ROS} 2: An experimental study},
  author={Castillo-S{\'a}nchez, Jos{\'e}-Borja and Gonz{\'a}lez-Parada, Eva and Cano-Garc{\'\i}a, Jos{\'e}-Manuel},
  journal={{IEEE} Access},
  volume={12},
  pages={142930--142943},
  year={2024},
  publisher={IEEE}
}

@article{lee2025dependency,
  title={Dependency chain analysis of {ROS} 2 {DDS} {QoS} policies: From lifecycle tutorial to static verification},
  author={Lee, Sanghoon and Kang, Junha and Park, Kyung-Joon},
  journal={arXiv preprint arXiv:2509.03381},
  year={2025}
}

@inproceedings{bode2023systematic,
  title={Systematic analysis of {DDS} implementations},
  author={Bode, Vincent and Buettner, David and Preclik, Tobias and Trinitis, Carsten and Schulz, Martin},
  booktitle={Proc. 24th Int. Middleware Conf.},
  pages={234--246},
  year={2023}
}

@article{sciangula2023bounding,
  title={Bounding the data-delivery latency of {DDS} messages in real-time applications},
  author={Sciangula, Gerlando and Casini, Daniel and Biondi, Alessandro and Scordino, Claudio and Di Natale, Marco and others},
  journal={Leibniz Int. Proc. in Informatics},
  volume={262},
  year={2023},
  publisher={Schloss Dagstuhl--Leibniz-Zentrum f{\"u}r Informatik}
}

@inproceedings{luo2023modeling,
  title={Modeling and analysis of inter-process communication delay in {ROS} 2},
  author={Luo, Xiantong and Jiang, Xu and Guan, Nan and Liang, Haochun and Liu, Songran and Yi, Wang},
  booktitle={Proc. 2023 {IEEE} Real-Time Systems Symp. ({RTSS})},
  pages={198--209},
  year={2023},
  publisher={IEEE}
}

@article{paul2024performance,
  title={Performance evaluation of {ROS} 2--{DDS} middleware implementations facilitating cooperative driving in autonomous vehicle},
  author={Paul, Sumit and Lephuoc, Danh and Hauswirth, Manfred},
  journal={arXiv preprint arXiv:2412.07485},
  year={2024}
}

@article{sanchez2011bloom,
  title={Bloom filter-based discovery protocol for {DDS} middleware},
  author={Sanchez-Monedero, Javier and Povedano-Molina, Javier and Lopez-Vega, Jose M and Lopez-Soler, Juan M},
  journal={J. of Parallel and Distributed Computing},
  volume={71},
  number={10},
  pages={1305--1317},
  year={2011},
  publisher={Elsevier}
}

@article{nwadiugwu2023mad,
  title={{MAD-DDS}: Memory-efficient automatic discovery data distribution service for large-scale distributed control network},
  author={Nwadiugwu, Williams-Paul and Kim, Dong-Seong and Ejaz, Waleed and Anpalagan, Alagan},
  journal={{IET} Communications},
  volume={17},
  number={12},
  pages={1432--1446},
  year={2023},
  publisher={Wiley}
}

@article{park2020real,
  title={Real-time characteristics of {ROS} 2.0 in multiagent robot systems: An empirical study},
  author={Park, Jaeho and Delgado, Raimarius and Choi, Byoung Wook},
  journal={{IEEE} Access},
  volume={8},
  pages={154637--154651},
  year={2020},
  publisher={IEEE}
}

@phdthesis{chen2019performance,
  title={Performance analysis of {ROS} 2 networks using variable {Quality of Service} and security constraints for autonomous systems},
  author={Chen, Zhaolin},
  year={2019},
  school={Naval Postgraduate School}
}

@inproceedings{maruyama2016exploring,
  title={Exploring the performance of {ROS} 2},
  author={Maruyama, Yuya and Kato, Shinpei and Azumi, Takuya},
  booktitle={Proc. 13th Int. Conf. on Embedded Software},
  pages={1--10},
  year={2016}
}

@article{gutierrez2018towards,
  title={Towards a distributed and real-time framework for robots: Evaluation of {ROS} 2.0 communications for real-time robotic applications},
  author={Guti{\'e}rrez, Carlos San Vicente and Juan, Lander Usategui San and Ugarte, Irati Zamalloa and Vilches, V{\'\i}ctor Mayoral},
  journal={arXiv preprint arXiv:1809.02595},
  year={2018}
}

@article{kluner2026modern,
  title={Modern middlewares for automated vehicles: A tutorial},
  author={Kl{\"u}ner, David Philipp and Molz, Marius and Kampmann, Alexandru and Kowalewski, Stefan and Alrifaee, Bassam},
  journal={{IEEE} Intelligent Transportation Systems Magazine},
  year={2026},
  publisher={IEEE}
}

@inproceedings{dey2023novel,
  title={A novel {ROS} 2 {QoS} policy-enabled synchronizing middleware for co-simulation of heterogeneous multi-robot systems},
  author={Dey, Emon and Walczak, Mikolaj and Anwar, Mohammad Saeid and Roy, Nirmalya and Freeman, Jade and Gregory, Timothy and Suri, Niranjan and Busart, Carl},
  booktitle={Proc. 2023 32nd Int. Conf. on Computer Communications and Networks ({ICCCN})},
  pages={1--10},
  year={2023},
  publisher={IEEE}
}

@article{he2025faster,
  title={A faster and more reliable middleware for autonomous driving systems},
  author={He, Yuankai and Shi, Weisong},
  journal={arXiv preprint arXiv:2510.11448},
  year={2025}
}

@article{casini2025survey,
  title={A survey of real-time support, analysis, and advancements in {ROS} 2},
  author={Casini, Daniel and Chen, Jian-Jia and Li, Jing and Reghenzani, Federico and Teper, Harun},
  journal={Leibniz Trans. on Embedded Systems ({LITES})},
  volume={11},
  number={1},
  pages={1:1--1:37},
  year={2026},
  publisher={Schloss Dagstuhl--Leibniz-Zentrum f{\"u}r Informatik}
}

@inproceedings{li2025ater,
  title={{ATER}: Adaptive task execution rate regulation for enhanced real-time performance in {ROS} 2},
  author={Li, Ruoxiang and Song, Ziwei and Lv, Mingsong and Wu, Jen-Ming and Xue, Chun Jason and Wang, Jianping and Guan, Nan},
  booktitle={Proc. 2025 {IEEE} 31st Int. Conf. on Embedded and Real-Time Computing Systems and Applications ({RTCSA})},
  pages={90--101},
  year={2025},
  publisher={IEEE}
}

@article{lin2024intelligent,
  title={An intelligent product-driven manufacturing system using {Data Distribution Service}},
  author={Lin, Chin-Te and Lu, Hong-Ji},
  journal={{IEEE} Access},
  volume={12},
  pages={16447--16461},
  year={2024},
  publisher={IEEE}
}

@inproceedings{blass2021automatic,
  title={Automatic latency management for {ROS} 2: Benefits, challenges, and open problems},
  author={Blass, Tobias and Hamann, Arne and Lange, Ralph and Ziegenbein, Dirk and Brandenburg, Bj{\"o}rn B},
  booktitle={Proc. 2021 {IEEE} 27th Real-Time and Embedded Technology and Applications Symp. ({RTAS})},
  pages={264--277},
  year={2021},
  publisher={IEEE}
}

@book{desbiens2023building,
  title={Building enterprise {IoT} solutions with {Eclipse IoT} technologies: An open source approach to edge computing},
  author={Desbiens, Fr{\'e}d{\'e}ric},
  year={2023},
  publisher={Springer}
}

@article{ginting2021chord,
  title={{CHORD}: Distributed data-sharing via hybrid {ROS} 1 and 2 for multi-robot exploration of large-scale complex environments},
  author={Ginting, Muhammad Fadhil and Otsu, Kyohei and Edlund, Jeffrey A and Gao, Jay and Agha-Mohammadi, Ali-Akbar},
  journal={{IEEE} Robotics and Automation Letters},
  volume={6},
  number={3},
  pages={5064--5071},
  year={2021},
  publisher={IEEE}
}

@inproceedings{kim2025cros,
  title={{CRoS-RT}: Cross-layer priority scheduling for predictable inter-process communication in {ROS} 2},
  author={Kim, Sohyun and Song, Juho and Lee, Kilho and Oh, Sangeun and Chwa, Hoon Sung},
  booktitle={Proc. 2025 {IEEE} 31st Real-Time and Embedded Technology and Applications Symp. ({RTAS})},
  pages={202--214},
  year={2025},
  publisher={IEEE}
}

@article{jones2022dots,
  title={{DOTS}: An open testbed for industrial swarm robotic solutions},
  author={Jones, Simon and Milner, Emma and Sooriyabandara, Mahesh and Hauert, Sabine},
  journal={arXiv preprint arXiv:2203.13809},
  year={2022}
}

@misc{dust2022dynamic,
  title={Dynamic connection handling for scalable robotic systems using {ROS} 2},
  author={Dust, Lukas Johannes and Persson, Emil},
  year={2022}
}

@inproceedings{jalil2022efficacy,
  title={Efficacy of local cache for performance improvement of reliable data transmission in aggregated robot processing architecture},
  author={Jalil, Abdul and Kobayashi, Jun},
  booktitle={Proc. 2022 22nd Int. Conf. on Control, Automation and Systems ({ICCAS})},
  pages={1339--1344},
  year={2022},
  publisher={IEEE}
}

@inproceedings{teper2022end,
  title={End-to-end timing analysis in {ROS} 2},
  author={Teper, Harun and G{\"u}nzel, Mario and Ueter, Niklas and von der Br{\"u}ggen, Georg and Chen, Jian-Jia},
  booktitle={Proc. 2022 {IEEE} Real-Time Systems Symp. ({RTSS})},
  pages={53--65},
  year={2022},
  publisher={IEEE}
}

@mastersthesis{blancoenhancing,
  title={Enhancing communication security in {ROS} 2},
  author={Blanco Romero, Francisco Javier},
  year={2024},
  school={Universidad Miguel Hern{\'a}ndez de Elche}
}

@article{xiong2010evaluating,
  title={Evaluating the performance of publish/subscribe platforms for information management in distributed real-time and embedded systems},
  author={Xiong, Ming and Parsons, Jeff and Edmondson, James and Nguyen, Hieu and Schmidt, Douglas C},
  journal={omgwiki.org/dds},
  year={2010}
}

@inproceedings{thulasiraman2020evaluation,
  title={Evaluation of the {Robot Operating System} 2 in lossy unmanned networks},
  author={Thulasiraman, Preetha and Chen, Zhaolin and Allen, B and Bingham, Brian},
  booktitle={Proc. 2020 {IEEE} Int. Systems Conf. ({SysCon})},
  pages={1--8},
  year={2020},
  publisher={IEEE}
}

@inproceedings{chen2025fogros2,
  title={{FogROS2-PLR}: Probabilistic latency-reliability for cloud robotics},
  author={Chen, Kaiyuan and Tian, Nan and Juette, Christian and Qiu, Tianshuang and Ren, Liu and Kubiatowicz, John and Goldberg, Ken},
  booktitle={Proc. 2025 {IEEE} Int. Conf. on Robotics and Automation ({ICRA})},
  pages={16290--16297},
  year={2025},
  publisher={IEEE}
}

@article{jo2025generation,
  title={Generation of critical information and sharing mechanism for multi-robot mission success},
  author={Jo, Daeil and Kwon, Yongjin},
  journal={{IEEE} Access},
  year={2025},
  publisher={IEEE}
}

@article{macenski2023impact,
  title={Impact of {ROS} 2 node composition in robotic systems},
  author={Macenski, Steve and Soragna, Alberto and Carroll, Michael and Ge, Zhenpeng},
  journal={{IEEE} Robotics and Automation Letters},
  volume={8},
  number={7},
  pages={3996--4003},
  year={2023},
  publisher={IEEE}
}

@article{randolph2021improving,
  title={Improving the predictability of event chains in {ROS} 2},
  author={Randolph, Charles},
  journal={Ph.D. dissertation, Delft Univ. of Technology},
  year={2021}
}

@article{sperling2026low,
  title={Low latency communication of large data objects by subscriber-centric selective data transfer},
  author={Sperling, Nora and Ernst, Rolf},
  journal={{ACM} Trans. on Cyber-Physical Systems},
  year={2026},
  publisher={ACM}
}

@inproceedings{abaza2024managing,
  title={Managing end-to-end timing jitters in {ROS} 2 computation chains},
  author={Abaza, Hazem and Roy, Debayan and Trach, Bohdan and Chang, Wanli and Saidi, Selma and Motakis, Antonios and Ren, Wei and Liu, Yutao},
  booktitle={Proc. 32nd Int. Conf. on Real-Time Networks and Systems},
  pages={229--241},
  year={2024}
}

@article{wytrkebowicz2021messaging,
  title={Messaging protocols for {IoT} systems --- a pragmatic comparison},
  author={Wytr{\k{e}}bowicz, Jacek and Cabaj, Krzysztof and Krawiec, Jerzy},
  journal={Sensors},
  volume={21},
  number={20},
  pages={6904},
  year={2021},
  publisher={MDPI}
}

@article{ranjan2026meta,
  title={{Meta-ROS}: A next-generation middleware architecture for adaptive and scalable robotic systems},
  author={Ranjan, Anshul and Damodar, Anoosh and Chougule, Neha and Nayak, Dhruva S and others},
  journal={arXiv preprint arXiv:2601.21011},
  year={2026}
}

@article{perez2015modeling,
  title={Modeling the {QoS} parameters of {DDS} for event-driven real-time applications},
  author={P{\'e}rez, H{\'e}ctor and Guti{\'e}rrez, J Javier},
  journal={J. of Systems and Software},
  volume={104},
  pages={126--140},
  year={2015},
  publisher={Elsevier}
}

@inproceedings{asjad2024neuroreality,
  title={{NeuroReality}: A {Data Distribution Service}-based inter-process communication middleware},
  author={Asjad, Syed and Harber, Evan and Santos, Veronica and Tyler, Dustin},
  booktitle={Proc. 2024 {IEEE} Conf. on Telepresence},
  pages={168--171},
  year={2024},
  publisher={IEEE}
}

@mastersthesis{hietasalo2024overview,
  title={Overview and performance analysis of robotic software framework {Flexbot}},
  author={Hietasalo, Lari},
  year={2024},
  school={Tampere University}
}

@article{koksal2017obstacles,
  title={Obstacles in {Data Distribution Service} middleware: A systematic review},
  author={K{\"o}ksal, {\"O}mer and Tekinerdogan, Bedir},
  journal={Future Generation Computer Systems},
  volume={68},
  pages={191--210},
  year={2017},
  publisher={Elsevier}
}

@article{wu2021oops,
  title={Oops! it's too late. Your autonomous driving system needs a faster middleware},
  author={Wu, Tianze and Wu, Baofu and Wang, Sa and Liu, Liangkai and Liu, Shaoshan and Bao, Yungang and Shi, Weisong},
  journal={{IEEE} Robotics and Automation Letters},
  volume={6},
  number={4},
  pages={7301--7308},
  year={2021},
  publisher={IEEE}
}

@phdthesis{fasano2025optimizing,
  title={Optimizing inter-process communication for robotics applications},
  author={Fasano, Andrea},
  year={2025},
  school={Politecnico di Torino}
}

@inproceedings{de2024orchestration,
  title={Orchestration-aware optimization of {ROS} 2 communication protocols},
  author={De Marchi, Mirco and Bombieri, Nicola},
  booktitle={Proc. 2024 Design, Automation \& Test in Europe Conf. \& Exhibition ({DATE})},
  pages={1--6},
  year={2024},
  publisher={IEEE}
}

@article{chovet2025performance,
  title={Performance comparison of {ROS} 2 middlewares for multi-robot mesh networks in planetary exploration},
  author={Chovet, Lo{\"\i}ck Pierre and Garcia, Gabriel Manuel and Bera, Abhishek and Richard, Antoine and Yoshida, Kazuya and Olivares-Mendez, Miguel Angel},
  journal={J. of Intelligent \& Robotic Systems},
  volume={111},
  number={1},
  pages={18},
  year={2025},
  publisher={Springer}
}

@article{jalil2023performance,
  title={Performance improvement of multi-robot data transmission in aggregated robot processing architecture with caches and {QoS} balancing optimization},
  author={Jalil, Abdul and Kobayashi, Jun and Saitoh, Takeshi},
  journal={Robotics},
  volume={12},
  number={3},
  pages={87},
  year={2023},
  publisher={MDPI}
}

@inproceedings{zieba2006preservation,
  title={Preservation of correctness during system reconfiguration in {Data Distribution Service} for real-time systems ({DDS})},
  author={Zieba, Bogumil and van Sinderen, Marten},
  booktitle={Proc. 26th {IEEE} Int. Conf. on Distributed Computing Systems Workshops ({ICDCSW})},
  pages={30--30},
  year={2006},
  publisher={IEEE}
}

@article{almadani2016qos,
  title={{QoS}-aware scalable video streaming using {Data Distribution Service}},
  author={Almadani, Basem and Alsaeedi, Mohammed and Al-Roubaiey, Anas},
  journal={Multimedia Tools and Applications},
  volume={75},
  number={10},
  pages={5841--5870},
  year={2016},
  publisher={Springer}
}

@inproceedings{dust2022quantitative,
  title={Quantitative analysis of communication handling for centralized multi-agent robot systems using {ROS} 2},
  author={Dust, Lukas Johannes and Persson, Emil and Ekstr{\"o}m, Mikael and Mubeen, Saad and Dean, Emmanuel},
  booktitle={Proc. 2022 {IEEE} 20th Int. Conf. on Industrial Informatics ({INDIN})},
  pages={624--629},
  year={2022},
  publisher={IEEE}
}

@inproceedings{wang2022ros,
  title={{ROS-SF}: A transparent and efficient {ROS} middleware using serialization-free message},
  author={Wang, Yu-Ping and Dong, Yuejiang and Tan, Gang},
  booktitle={Proc. 23rd {ACM/IFIP} Int. Middleware Conf.},
  pages={82--93},
  year={2022}
}

@inproceedings{ishikawa2025ros,
  title={{ROS} 2 {Agnocast}: Supporting unsized message types for true zero-copy publish/subscribe {IPC}},
  author={Ishikawa--Aso, Takahiro and Kato, Shinpei},
  booktitle={Proc. 2025 28th Int. Symp. on Real-Time Distributed Computing ({ISORC})},
  pages={1--10},
  year={2025},
  publisher={IEEE}
}

@article{macenski2022robot,
  title={{Robot Operating System} 2: Design, architecture, and uses in the wild},
  author={Macenski, Steven and Foote, Tully and Gerkey, Brian and Lalancette, Chris and Woodall, William},
  journal={Science Robotics},
  volume={7},
  number={66},
  pages={eabm6074},
  year={2022},
  publisher={AAAS}
}

@inproceedings{shih2022scalable,
  title={Scalable and bounded-time decisions on edge device network using {Eclipse Zenoh}},
  author={Shih, Chi-Sheng and Lin, Hsiang-Jui and Yuan, Yuyuan and Kuo, Yi-Hung and Liang, Wen-Yew},
  booktitle={Proc. 2022 {IEEE} 28th Int. Conf. on Embedded and Real-Time Computing Systems and Applications ({RTCSA})},
  pages={170--179},
  year={2022},
  publisher={IEEE}
}

@inproceedings{romero2013self,
  title={Self-adaptive quality-of-service in distributed middleware for robotics},
  author={Romero-Garc{\'e}s, Adri{\'a}n and Ingl{\'e}s-Romero, Juan F and Mart{\'\i}nez, J and Vicente-Chicote, Cristina},
  booktitle={Proc. Workshop on Recognition and Action for Scene Understanding ({REACTS})},
  year={2013}
}

@article{calisi2010software,
  title={Software development for networked robot systems},
  author={Calisi, Daniele and Fedi, Francesco and Leo, Alberto and Nardi, Daniele},
  journal={{IFAC} Proc. Volumes},
  volume={43},
  number={16},
  pages={605--610},
  year={2010},
  publisher={Elsevier}
}

@article{chisualițua2025stepping,
  title={Stepping towards {Zenoh} protocol in automotive scenarios},
  author={Chis{\u{a}}liț{\u{a}}, Andreea-lasmina and Korodi, Adrian},
  journal={{IEEE} Access},
  year={2025},
  publisher={IEEE}
}

@article{hakiri2014supporting,
  title={Supporting {SIP}-based end-to-end {Data Distribution Service} {QoS} in {WAN}s},
  author={Hakiri, Akram and Berthou, Pascal and Gokhale, Aniruddha and Schmidt, Douglas C and Gayraud, Thierry},
  journal={J. of Systems and Software},
  volume={95},
  pages={100--121},
  year={2014},
  publisher={Elsevier}
}

@inproceedings{wang2019tzc,
  title={{TZC}: Efficient inter-process communication for robotics middleware with partial serialization},
  author={Wang, Yu-Ping and Tan, Wende and Hu, Xu-Qiang and Manocha, Dinesh and Hu, Shi-Min},
  booktitle={Proc. 2019 {IEEE/RSJ} Int. Conf. on Intelligent Robots and Systems ({IROS})},
  pages={7805--7812},
  year={2019},
  publisher={IEEE}
}

@article{szabo2023toward,
  title={Toward the automatic network resource management of {Robot Operating System} in programmable mobile networks},
  author={Szab{\'o}, G{\'e}za},
  journal={{IEEE} Access},
  volume={11},
  pages={65934--65955},
  year={2023},
  publisher={IEEE}
}

@inproceedings{plasberg2022towards,
  title={Towards distributed real-time capable robotic control using {ROS} 2},
  author={Plasberg, Carsten and Nessau, Hendrik and Timmermann, David and Eichmann, Christian and Roennau, Arne and Dillmann, R{\"u}diger},
  booktitle={Proc. 2022 {IEEE} 18th Int. Conf. on Automation Science and Engineering ({CASE})},
  pages={2205--2210},
  year={2022},
  publisher={IEEE}
}

@inproceedings{jaiswal2023wiros,
  title={{WiROS}: A {QoS} software solution for {ROS} 2 in a {WiFi} network},
  author={Jaiswal, Bishal and Tyagi, Himanshu and Gopalan, Aditya and Sevani, Vishal},
  booktitle={Proc. 2023 15th Int. Conf. on {COMmunication} Systems \& {NETworkS} ({COMSNETS})},
  pages={216--218},
  year={2023},
  publisher={IEEE}
}

@article{hu2024wireless,
  title={Wireless multi-robot collaboration: Communications, perception, control, and planning},
  author={Hu, Kai and Zou, Longhao and Chen, Zihao and Jiang, Jun and Jiang, Fan and Tao, Xiaofeng},
  journal={{IEEE} Network},
  volume={39},
  number={4},
  pages={302--311},
  year={2024},
  publisher={IEEE}
}

@inproceedings{al2012wireless,
  title={Wireless video streaming over {Data Distribution Service} middleware},
  author={Al-madani, Basem and Al-Roubaiey, Anas and Al-shehari, Taher},
  booktitle={Proc. 2012 {IEEE} Int. Conf. on Computer Science and Automation Engineering},
  pages={263--266},
  year={2012},
  publisher={IEEE}
}

@inproceedings{corsaro2023zenoh,
  title={{Zenoh}: Unifying communication, storage and computation from the cloud to the microcontroller},
  author={Corsaro, Angelo and Cominardi, Luca and Hecart, Olivier and Baldoni, Gabriele and Avital, Julien Enoch Pierre and Loudet, Julien and Guimares, Carlos and Ilyin, Michael and Bannov, Dmitrii},
  booktitle={Proc. 2023 26th Euromicro Conf. on Digital System Design ({DSD})},
  pages={422--428},
  year={2023},
  publisher={IEEE}
}

@article{valls2012iland,
  title={{iLAND}: An enhanced middleware for real-time reconfiguration of service oriented distributed real-time systems},
  author={Valls, Marisol Garc{\'\i}a and L{\'o}pez, Iago Rodr{\'\i}guez and Villar, Laura Fern{\'a}ndez},
  journal={{IEEE} Trans. on Industrial Informatics},
  volume={9},
  number={1},
  pages={228--236},
  year={2012},
  publisher={IEEE}
}

@article{ye2023ros2,
  title={{ROS} 2 real-time performance optimization and evaluation},
  author={Ye, Yanlei and Nie, Zhenguo and Liu, Xinjun and Xie, Fugui and Li, Zihao and Li, Peng},
  journal={Chinese J. of Mechanical Engineering},
  volume={36},
  number={1},
  pages={144},
  year={2023},
  publisher={Springer}
}

@inproceedings{teper2023timing,
  title={Timing-aware {ROS} 2 architecture and system optimization},
  author={Teper, Harun and Betz, Tobias and Von Der Br{\"u}ggen, Georg and Chen, Kuan-Hsun and Betz, Johannes and Chen, Jian-Jia},
  booktitle={Proc. 2023 {IEEE} 29th Int. Conf. on Embedded and Real-Time Computing Systems and Applications ({RTCSA})},
  pages={206--215},
  year={2023},
  publisher={IEEE}
}

@article{baron2025performance,
  title={On the performance of {Zenoh} in industrial {IoT} scenarios},
  author={Baron, Miguel and Diez, Luis and Zverev, Mihail and Juarez, Jose R and Ag{\"u}ero, Ram{\'o}n},
  journal={Ad Hoc Networks},
  volume={170},
  pages={103784},
  year={2025},
  publisher={Elsevier}
}

@misc{ros_metrics_clean,
  author       = {{Open Robotics}},
  title        = {{ROS} metrics},
  year         = {2026},
  howpublished = {[Online]. Available: \url{http://metrics.ros.org/}}
}

@inproceedings{casini2019response,
  title={Response-time analysis of {ROS} 2 processing chains under reservation-based scheduling},
  author={Casini, Daniel and Bla{\ss}, Tobias and L{\"u}tkebohle, Ingo and Brandenburg, Bj{\"o}rn},
  booktitle={Proc. 31st Euromicro Conf. on Real-Time Systems},
  pages={1--23},
  year={2019},
  publisher={Schloss Dagstuhl}
}

@inproceedings{teper2024end,
  title={End-to-end timing analysis and optimization of multi-executor {ROS} 2 systems},
  author={Teper, Harun and Betz, Tobias and G{\"u}nzel, Mario and Ebner, Dominic and Von Der Br{\"u}ggen, Georg and Betz, Johannes and Chen, Jian-Jia},
  booktitle={Proc. 2024 {IEEE} 30th Real-Time and Embedded Technology and Applications Symp. ({RTAS})},
  pages={212--224},
  year={2024},
  publisher={IEEE}
}

@inproceedings{tang2023real,
  title={Real-time performance analysis of processing systems on {ROS} 2 executors},
  author={Tang, Yue and Guan, Nan and Jiang, Xu and Luo, Xiantong and Yi, Wang},
  booktitle={Proc. 2023 {IEEE} 29th Real-Time and Embedded Technology and Applications Symp. ({RTAS})},
  pages={80--92},
  year={2023},
  publisher={IEEE}
}

@article{liu2024rtex,
  title={{RTeX}: An efficient and timing-predictable multithreaded executor for {ROS} 2},
  author={Liu, Songran and Jiang, Xu and Guan, Nan and Wang, Zilong and Yu, Minghe and Yi, Wang},
  journal={{IEEE} Trans. on Computer-Aided Design of Integrated Circuits and Systems},
  volume={43},
  number={9},
  pages={2578--2591},
  year={2024},
  publisher={IEEE}
}

@inproceedings{staschulat2020rclc,
  title={The rclc executor: Domain-specific deterministic scheduling mechanisms for {ROS} applications on microcontrollers: Work-in-progress},
  author={Staschulat, Jan and L{\"u}tkebohle, Ingo and Lange, Ralph},
  booktitle={Proc. 2020 Int. Conf. on Embedded Software ({EMSOFT})},
  pages={18--19},
  year={2020},
  publisher={IEEE}
}

@inproceedings{yang2020exploring,
  title={Exploring real-time executor on {ROS} 2},
  author={Yang, Yuqing and Azumi, Takuya},
  booktitle={Proc. 2020 {IEEE} Int. Conf. on Embedded Software and Systems ({ICESS})},
  pages={1--8},
  year={2020},
  publisher={IEEE}
}

@article{an2023multi,
  title={Multi-robot systems and cooperative object transport: Communications, platforms, and challenges},
  author={An, Xing and Wu, Celimuge and Lin, Yangfei and Lin, Min and Yoshinaga, Tsutomu and Ji, Yusheng},
  journal={{IEEE} Open J. of the Computer Society},
  volume={4},
  pages={23--36},
  year={2023},
  publisher={IEEE}
}

@article{cruz2012dds,
  title={A {DDS}-based middleware for quality-of-service and high-performance networked robotics},
  author={Cruz, Jes{\'u}s Mart{\'\i}nez and Romero-Garc{\'e}s, Adri{\'a}n and Rubio, Juan Pedro Bandera and Robles, Rebeca Marfil and Rubio, Antonio Bandera},
  journal={Concurrency and Computation: Practice and Experience},
  volume={24},
  number={16},
  pages={1940--1952},
  year={2012},
  publisher={Wiley}
}

@article{al2025integrating,
  title={Integrating {Data Distribution Service} ({DDS}) in smart traffic systems: A comprehensive review},
  author={Al-Madani, Basem and Hasan, Shihab and Abualhassan, Amer and Aliyu, Farouq},
  journal={Computer},
  volume={58},
  number={2},
  pages={25--34},
  year={2025},
  publisher={IEEE}
}

@inproceedings{castillo2024novel,
  title={A novel testbed for evaluating {ROS} 2 robot swarm wireless communications},
  author={Castillo-S{\'a}nchez, Jos{\'e}-Borja and Gonz{\'a}lez-Parada, Eva and Cano-Garc{\'\i}a, Jos{\'e}-Manuel},
  booktitle={Proc. 2024 {IEEE} 22nd Mediterranean Electrotechnical Conf. ({MELECON})},
  pages={68--73},
  year={2024},
  publisher={IEEE}
}

@article{agarwal2025scalable,
  title={A scalable multi-robot framework for decentralized and asynchronous perception-action-communication loops},
  author={Agarwal, Saurav and Ribeiro, A and Kumar, V},
  journal={arXiv preprint arXiv:2309.10164},
  year={2025}
}

@inproceedings{kong2025design,
  title={Design of a multimodal control system based on {ROS} 2: A hierarchical architecture for real-time human-robot collaboration},
  author={Kong, Shiyuan and Zhang, Hanwei and Ye, Cheng and Xu, Yitong and Li, Ang},
  booktitle={Proc. 2025 7th Int. Conf. on Data-driven Optimization of Complex Systems ({DOCS})},
  pages={441--446},
  year={2025},
  publisher={IEEE}
}

@article{groshev2023edge,
  title={Edge robotics: Are we ready? An experimental evaluation of current vision and future directions},
  author={Groshev, Milan and Baldoni, Gabriele and Cominardi, Luca and De La Oliva, Antonio and Gazda, Robert},
  journal={Digital Communications and Networks},
  volume={9},
  number={1},
  pages={166--174},
  year={2023},
  publisher={Elsevier}
}

@inproceedings{naury2025communication,
  title={Communication isolation for multi-robot systems using {ROS} 2},
  author={Naury, Lucas and Gouguet, Adam and Lozenguez, Guillaume and Fabresse, Luc},
  booktitle={Proc. 40th {ACM/SIGAPP} Symp. on Applied Computing},
  pages={850--858},
  year={2025}
}

@inproceedings{tang2020response,
  title={Response time analysis and priority assignment of processing chains on {ROS} 2 executors},
  author={Tang, Yue and Feng, Zhiwei and Guan, Nan and Jiang, Xu and Lv, Mingsong and Deng, Qingxu and Yi, Wang},
  booktitle={Proc. 2020 {IEEE} Real-Time Systems Symp. ({RTSS})},
  pages={231--243},
  year={2020},
  publisher={IEEE}
}

@inproceedings{dust2023dynamic,
  title={Dynamic priority scheduling for periodic systems using {ROS} 2},
  author={Dust, Lukas and Mubeen, Saad},
  booktitle={Proc. Int. Conf. on Engineering of Computer-Based Systems},
  pages={239--243},
  year={2023},
  publisher={Springer}
}

@inproceedings{enright2024paam,
  title={{PAAM}: A framework for coordinated and priority-driven accelerator management in {ROS} 2},
  author={Enright, Daniel and Xiang, Yecheng and Choi, Hyunjong and Kim, Hyoseung},
  booktitle={Proc. 2024 {IEEE} 30th Real-Time and Embedded Technology and Applications Symp. ({RTAS})},
  year={2024},
  publisher={IEEE}
}

@inproceedings{choi2021picas,
  title={{PiCAS}: New design of priority-driven chain-aware scheduling for {ROS} 2},
  author={Choi, Hyunjong and Xiang, Yecheng and Kim, Hyoseung},
  booktitle={Proc. 2021 {IEEE} 27th Real-Time and Embedded Technology and Applications Symp. ({RTAS})},
  pages={251--263},
  year={2021},
  publisher={IEEE}
}

@inproceedings{xu20223ds,
  title={{3DS}: An efficient {DPDK}-based {Data Distribution Service} for distributed real-time applications},
  author={Xu, Tianyu and Chen, Xianzhang and Wu, Changze and Wang, Jiapin and Zheng, Rongwei and Liu, Duo and Tan, Yujuan and Ren, Ao and Li, Jian},
  booktitle={Proc. 2022 {IEEE} 24th Int. Conf. on High Performance Computing \& Communications ({HPCC/DSS/SmartCity/DependSys})},
  pages={1283--1290},
  year={2022},
  publisher={IEEE}
}

@article{qi2025novel,
  title={A novel open and efficient robot development framework based on {Data Distribution Service} orchestration for agile manufacturing},
  author={Qi, Le and Zhang, Xiaogang and Tan, Haoran and Chen, Hua and Wang, Yaonan},
  journal={Robotics and Computer-Integrated Manufacturing},
  volume={96},
  pages={103067},
  year={2025},
  publisher={Elsevier}
}

@inproceedings{baldoni2021facilitating,
  title={Facilitating distributed data-flow programming with {Eclipse Zenoh}: The {ERDOS} case},
  author={Baldoni, Gabriele and Loudet, Julien and Cominardi, Luca and Corsaro, Angelo and He, Yong},
  booktitle={Proc. 1st Workshop on Serverless Mobile Networking for 6{G} Communications},
  pages={13--18},
  year={2021}
}

@techreport{ros_metrics_2025,
  author      = {Katherine Scott and Tully Foote},
  title       = {2025 {ROS} metrics report},
  institution = {Open Source Robotics Foundation ({OSRF})},
  year        = {2025},
  url         = {https://discourse.openrobotics.org/t/2025-ros-metrics-report/52575}
}

@article{li2022autoware_perf,
  title={{Autoware\_Perf}: A tracing and performance analysis framework for {ROS} 2 applications},
  author={Li, Zihang and Hasegawa, Atsushi and Azumi, Takuya},
  journal={J. of Systems Architecture},
  volume={123},
  pages={102341},
  year={2022},
  publisher={Elsevier}
}

@article{bonci2023robot,
  title={{Robot Operating System} 2 ({ROS} 2)-based frameworks for increasing robot autonomy: A survey},
  author={Bonci, Andrea and Gaudeni, Francesco and Giannini, Maria Cristina and Longhi, Sauro},
  journal={Applied Sciences},
  volume={13},
  number={23},
  pages={12796},
  year={2023},
  publisher={MDPI}
}

@misc{ros2_diff_middleware_vendors,
  author       = {{Open Robotics}},
  title        = {Different {ROS} 2 middleware vendors},
  howpublished = {[Online]. Available: \url{https://docs.ros.org/en/kilted/Concepts/Intermediate/About-Different-Middleware-Vendors.html}},
  note         = {Accessed: Mar. 28, 2026}
}

@misc{ros2_creating_rmw_implementation,
  author       = {{Open Robotics}},
  title        = {Creating an {RMW} implementation},
  howpublished = {[Online]. Available: \url{https://docs.ros.org/en/kilted/Tutorials/Advanced/Creating-An-RMW-Implementation.html}},
  note         = {Accessed: Mar. 28, 2026}
}

@misc{ros2_middleware_implementations,
  author       = {{Open Robotics}},
  title        = {{ROS} 2 middleware implementations},
  howpublished = {[Online]. Available: \url{https://docs.ros.org/en/kilted/Concepts/Advanced/About-Middleware-Implementations.html}},
  note         = {Accessed: Mar. 28, 2026}
}

@misc{ros2_internal_interfaces,
  author       = {{Open Robotics}},
  title        = {Internal {ROS} 2 interfaces},
  howpublished = {[Online]. Available: \url{https://docs.ros.org/en/kilted/Concepts/Advanced/About-Internal-Interfaces.html}},
  note         = {Accessed: Mar. 28, 2026}
}

@misc{ros2_changes_ros1_ros2,
  author       = {{Open Robotics}},
  title        = {Changes between {ROS} 1 and {ROS} 2},
  howpublished = {[Online]. Available: \url{https://design.ros2.org/articles/changes.html}},
  note         = {Accessed: Mar. 28, 2026}
}

@misc{ros2_executors,
  author       = {{Open Robotics}},
  title        = {Executors},
  howpublished = {[Online]. Available: \url{https://docs.ros.org/en/kilted/Concepts/Intermediate/About-Executors.html}},
  note         = {Accessed: Mar. 28, 2026}
}

@misc{ros2_kilted_release,
  author       = {{Open Robotics}},
  title        = {{ROS} 2 {Kilted Kaiju} released},
  howpublished = {[Online]. Available: \url{https://www.openrobotics.org/blog/2025/5/23/ros-2-kilted-kaiju-released}},
  note         = {Accessed: Mar. 28, 2026}
}

@misc{zenoh_what_is_zenoh,
  author       = {{ZettaScale Technology}},
  title        = {What is {Zenoh}?},
  howpublished = {[Online]. Available: \url{https://zenoh.io/docs/overview/what-is-zenoh/}},
  note         = {Accessed: Mar. 28, 2026}
}

@misc{zenoh_abstractions,
  author       = {{ZettaScale Technology}},
  title        = {{Zenoh} abstractions},
  howpublished = {[Online]. Available: \url{https://zenoh.io/docs/manual/abstractions/}},
  note         = {Accessed: Mar. 28, 2026}
}

@misc{rmw_zenoh_github,
  author       = {{Open Robotics}},
  title        = {rmw\_zenoh},
  howpublished = {[Online]. Available: \url{https://github.com/ros2/rmw_zenoh/tree/kilted}},
  note         = {GitHub repository, Accessed: Mar. 28, 2026}
}

@misc{ros2_topics_services_actions,
  author       = {{Open Robotics}},
  title        = {Topics vs services vs actions},
  howpublished = {[Online]. Available: \url{https://docs.ros.org/en/kilted/How-To-Guides/Topics-Services-Actions.html}},
  note         = {Accessed: Mar. 28, 2026}
}

@misc{ros2_understanding_topics,
  author       = {{Open Robotics}},
  title        = {Understanding topics},
  howpublished = {[Online]. Available: \url{https://docs.ros.org/en/kilted/Tutorials/Beginner-CLI-Tools/Understanding-ROS2-Topics/Understanding-ROS2-Topics.html}},
  note         = {Accessed: Mar. 28, 2026}
}

@misc{ros2_understanding_services,
  author       = {{Open Robotics}},
  title        = {Understanding services},
  howpublished = {[Online]. Available: \url{https://docs.ros.org/en/kilted/Tutorials/Beginner-CLI-Tools/Understanding-ROS2-Services/Understanding-ROS2-Services.html}},
  note         = {Accessed: Mar. 28, 2026}
}

@misc{ros2_understanding_actions,
  author       = {{Open Robotics}},
  title        = {Understanding actions},
  howpublished = {[Online]. Available: \url{https://docs.ros.org/en/kilted/Tutorials/Beginner-CLI-Tools/Understanding-ROS2-Actions/Understanding-ROS2-Actions.html}},
  note         = {Accessed: Mar. 28, 2026}
}

@misc{ros2_interfaces,
  author       = {{Open Robotics}},
  title        = {Interfaces},
  howpublished = {[Online]. Available: \url{https://docs.ros.org/en/kilted/Concepts/Basic/About-Interfaces.html}},
  note         = {Accessed: Mar. 28, 2026}
}

@article{bernstein1996middleware,
  title={Middleware: A model for distributed system services},
  author={Bernstein, Philip A},
  journal={Communications of the {ACM}},
  volume={39},
  number={2},
  pages={86--98},
  year={1996},
  publisher={ACM}
}

@inproceedings{emmerich2000software,
  title={Software engineering and middleware: A roadmap},
  author={Emmerich, Wolfgang},
  booktitle={Proc. Conf. on the Future of Software Engineering},
  pages={117--129},
  year={2000}
}

@article{corsaro2014data,
  title={The {Data Distribution Service} tutorial},
  author={Corsaro, Angelo},
  journal={Technical Report 4.0},
  year={2014},
  publisher={PrismTech}
}

@article{stadnik2024overview,
  title={Overview analysis of {Micro-ROS} system as an embedded solution for microcontrollers in automatics and robotics applications},
  author={Stadnik, Bart{\l}omiej and Wymys{\l}owski, Artur},
  journal={Przegl{\k{a}}d Elektrotechniczny},
  volume={100},
  year={2024}
}

@incollection{pohnl2023shared,
  title={Shared-memory-based lock-free queues: The key to fast and robust communication on safety-critical edge devices},
  author={P{\"o}hnl, Michael and Eltzschig, Christian and Blass, Tobias},
  booktitle={Proc. Cyber-Physical Systems and Internet of Things Week 2023},
  pages={179--184},
  year={2023}
}

@incollection{parmar2020syntactic,
  title={Syntactic interoperability in real-time systems, {ROS} 2, and adaptive {AUTOSAR} using {Data Distribution Services}: An approach},
  author={Parmar, Navrattan and Ranga, Virender and Simhachalam Naidu, B},
  booktitle={Inventive Communication and Computational Technologies: Proc. {ICICCT} 2019},
  pages={257--274},
  year={2020},
  publisher={Springer}
}

@inproceedings{kouril2024performance,
  title={Performance evaluation of a {ROS} 2 based automated driving system},
  author={Kouril, Jorin and Sch{\"a}ufele, Bernd and Radusch, Ilja and Schnor, Bettina},
  booktitle={Proc. 10th Int. Conf. on Vehicle Technology and Intelligent Transport Systems ({VEHITS})},
  pages={52--63},
  year={2024},
  publisher={SciTePress}
}

@inproceedings{peng2022scheduling,
  title={Scheduling performance evaluation framework for {ROS} 2 applications},
  author={Peng, Bo and Hasegawa, Atsushi and Azumi, Takuya},
  booktitle={Proc. 2022 {IEEE} 24th Int. Conf. on High Performance Computing \& Communications ({HPCC/DSS/SmartCity/DependSys})},
  pages={2031--2038},
  year={2022},
  publisher={IEEE}
}

@article{putra2015node,
  title={Node discovery scheme of {DDS} for combat management system},
  author={Putra, Handityo Aulia and Kim, Dong-Seong},
  journal={Computer Standards \& Interfaces},
  volume={37},
  pages={20--28},
  year={2015},
  publisher={Elsevier}
}

@article{maggi2022security,
  title={A security analysis of the {Data Distribution Service} ({DDS}) protocol},
  author={Maggi, Federico and Vosseler, Rainer and Cheng, Mars and Kuo, Patrick and Toyama, Chizuru and Yen, T and Vilches, E Boasson V},
  journal={Trend Micro Research},
  pages={15--20},
  year={2022}
}

@phdthesis{caruso2024autonomous,
  title={Autonomous navigation for quadruped robots: Development and optimization on the {Unitree Go1} platform},
  author={Caruso, Gabriele},
  year={2024},
  school={Politecnico di Torino}
}

@mastersthesis{zanni2024efficient,
  title={Efficient support for deep sleeping modes in embedded systems: The case of {Zenoh-pico}},
  author={Zanni, Andrea},
  year={2024},
  school={University of Bologna}
}

@inproceedings{michaud2018attacking,
  title={Attacking {OMG} {Data Distribution Service} ({DDS}) based real-time mission critical distributed systems},
  author={Michaud, Michael James and Dean, Thomas and Leblanc, Sylvain P},
  booktitle={Proc. 2018 13th Int. Conf. on Malicious and Unwanted Software ({MALWARE})},
  pages={68--77},
  year={2018},
  publisher={IEEE}
}

@article{tijero2012schedulability,
  title={On the schedulability of a data-centric real-time distribution middleware},
  author={Tijero, H{\'e}ctor P{\'e}rez and Guti{\'e}rrez, J Javier},
  journal={Computer Standards \& Interfaces},
  volume={34},
  number={1},
  pages={203--211},
  year={2012},
  publisher={Elsevier}
}

@article{scordino2022hardware,
  title={Hardware acceleration of {Data Distribution Service} ({DDS}) for automotive communication and computing},
  author={Scordino, Claudio and Mari{\~n}o, Angela Gonzalez and Fons, Francesc},
  journal={{IEEE} Access},
  volume={10},
  pages={109626--109651},
  year={2022},
  publisher={IEEE}
}

@manual{omg2022rtps,
  title         = {The real-time publish-subscribe protocol {DDS} interoperability wire protocol ({DDSI-RTPS}) specification, version 2.5},
  author        = {{Object Management Group}},
  organization  = {Object Management Group},
  month         = {Apr.},
  year          = {2022},
  note          = {{OMG} Document Number: formal/2022-04-01},
  url           = {https://www.omg.org/spec/DDSI-RTPS/2.5}
}

@misc{desbiens2021zenoh,
  title        = {{Zenoh}: A next-generation protocol for {IoT} and edge computing},
  author       = {Desbiens, Fr{\'e}d{\'e}ric},
  year         = {2021},
  month        = {Sep.},
  howpublished = {Presentation at Open Source Summit + Embedded Linux Conf. ({OSS} + {ELC}) 2021},
  organization = {Eclipse Foundation}
}

@article{hakiri2013supporting,
  title={Supporting end-to-end quality of service properties in {OMG} {Data Distribution Service} publish/subscribe middleware over wide area networks},
  author={Hakiri, Akram and Berthou, Pascal and Gokhale, Aniruddha and Schmidt, Douglas C and Gayraud, Thierry},
  journal={J. of Systems and Software},
  volume={86},
  number={10},
  pages={2574--2593},
  year={2013},
  publisher={Elsevier}
}

@inproceedings{fernandez2020performance,
  title={Performance study of the {Robot Operating System} 2 with {QoS} and cyber security settings},
  author={Fernandez, J and Allen, B and Thulasiraman, Preetha and Bingham, Brian},
  booktitle={Proc. 2020 {IEEE} Int. Systems Conf. ({SysCon})},
  pages={1--6},
  year={2020},
  publisher={IEEE}
}

@article{barcis2020information,
  title={Information distribution in multi-robot systems: Utility-based evaluation model},
  author={Barci{\'s}, Micha{\l} and Barci{\'s}, Agata and Hellwagner, Hermann},
  journal={Sensors},
  volume={20},
  number={3},
  pages={710},
  year={2020},
  publisher={MDPI}
}

@article{pardo2005introduction,
  title={An introduction to {DDS} and data-centric communications},
  author={Pardo-Castellote, Gerardo and Farabaugh, Bert and Warren, Rick},
  journal={Real-Time Innovations},
  volume={61},
  year={2005}
}

@online{zettascale_zenoh_storage,
  author = {{ZettaScale Technology}},
  title  = {{Zenoh} plugin storage manager},
  howpublished = {[Online]. Available: \url{https://zenoh.io/docs/manual/plugin-storage-manager/}},
  year   = {2024},
  note   = {Accessed: 2026-03-30}
}

@online{zettascale_zenoh_deployment,
  author = {{ZettaScale Technology}},
  title  = {{Zenoh} deployment guide},
  howpublished = {[Online]. Available: \url{https://zenoh.io/docs/getting-started/deployment/}},
  year   = {2024},
  note   = {Accessed: 2026-03-30}
}

@online{corsaro2021reliability,
  author       = {Corsaro, Angelo},
  title        = {{Zenoh} reliability: How it works?},
  year         = {2021},
  month        = {Jun.},
  howpublished = {[Online]. Available: \url{https://zenoh.io/blog/2021-06-14-zenoh-reliability/}},
  organization = {ZettaScale Technology}
}

@online{zettascale2023charmander,
  author = {{ZettaScale Technology}},
  title  = {{Zenoh} for {ROS} 2: {Charmander} and beyond},
  year   = {2023},
  month  = {Jun.},
  howpublished = {[Online]. Available: \url{https://zenoh.io/blog/2023-06-05-charmander2/}}
}

@online{zenoh_rust_liveliness,
  author = {{ZettaScale Technology}},
  title  = {Module zenoh::liveliness --- {Rust} documentation},
  howpublished = {[Online]. Available: \url{https://docs.rs/zenoh/latest/zenoh/liveliness/index.html}},
  year   = {2024},
  note   = {Official Rust API Reference}
}

@online{zenoh_rust_session,
  author = {{ZettaScale Technology}},
  title  = {Module zenoh::session --- {Rust} documentation},
  howpublished = {[Online]. Available: \url{https://docs.rs/zenoh/latest/zenoh/session/index.html}},
  year   = {2024},
  note   = {Official Rust API Reference}
}

@inproceedings{krinkin2018data,
  title={{Data Distribution Services} performance evaluation framework},
  author={Krinkin, Kirill and Filatov, Antoni and Filatov, Artyom and Kurishev, Oleg and Lyanguzov, Alexander},
  booktitle={Proc. 2018 22nd Conf. of Open Innovations Association ({FRUCT})},
  pages={94--100},
  year={2018},
  publisher={IEEE}
}

@article{al2024ros,
  title={{ROS} 2 in a nutshell: A survey},
  author={Al-Batati, Abdulrahman and Koubaa, Anis and Gabr, Khaled and Abdelkader, Mohamed and Aloquaily, Hamad},
  journal={{ACM} Computing Surveys},
  year={2025},
  publisher={ACM}
}

@article{jiang2020message,
  title={Message passing optimization in {Robot Operating System}},
  author={Jiang, Ziyue and Gong, Yifan and Zhai, Jidong and Wang, Yu-Ping and Liu, Wei and Wu, Hao and Jin, Jiangming},
  journal={Int. J. of Parallel Programming},
  volume={48},
  number={1},
  pages={119--136},
  year={2020},
  publisher={Springer}
}

@article{peeroo2023survey,
  title={A survey on experimental performance evaluation of {Data Distribution Service} ({DDS}) implementations},
  author={Peeroo, Kaleem and Popov, Peter and Stankovic, Vladimir},
  journal={arXiv preprint arXiv:2310.16630},
  year={2023}
}

@inproceedings{an2014content,
  title={Content-based filtering discovery protocol ({CFDP}): Scalable and efficient {OMG} {DDS} discovery protocol},
  author={An, Kyoungho and Gokhale, Aniruddha and Schmidt, Douglas and Tambe, Sumant and Pazandak, Paul and Pardo-Castellote, Gerardo},
  booktitle={Proc. 8th {ACM} Int. Conf. on Distributed Event-Based Systems},
  pages={130--141},
  year={2014}
}

@inproceedings{dehnavi2021compros,
  title={{CompROS}: A composable {ROS} 2 based architecture for real-time embedded robotic development},
  author={Dehnavi, Saeid and Koedam, Martijn and Nelson, Andrew and Goswami, Dip and Goossens, Kees},
  booktitle={Proc. 2021 {IEEE/RSJ} Int. Conf. on Intelligent Robots and Systems ({IROS})},
  pages={6449--6455},
  year={2021},
  publisher={IEEE}
}

@article{li2026fault,
  title={Fault tolerance {DDS} communications in {Time-Sensitive Networking} using {IEEE} 802.1 {CB} redundancy mechanisms},
  author={Li, Binqi and Zhu, Yuan and Zhao, Yingqi and Cui, Ke and Zhong, Xu and Lu, Ke},
  journal={J. of Systems Architecture},
  pages={103756},
  year={2026},
  publisher={Elsevier}
}

@inproceedings{thulasiraman2022evaluation,
  title={Evaluation of the {Data Distribution Service} for a lossy autonomous hybrid system},
  author={Thulasiraman, Preetha and Cheng, Yu Kheng Denny and Allen, Bruce},
  booktitle={Proc. 2022 {IEEE} Int. Systems Conf. ({SysCon})},
  pages={1--8},
  year={2022},
  publisher={IEEE}
}

@manual{omgdds1.4,
  title        = {{Data Distribution Service} ({DDS}), version 1.4},
  author       = {{Object Management Group}},
  organization = {Object Management Group},
  month        = {Apr.},
  year         = {2015},
  note         = {formal/2015-04-10},
  url          = {https://www.omg.org/spec/DDS/1.4}
}

@inproceedings{toffetti2023ros,
  title={{ROS}-based robotic applications orchestration in the compute continuum: Challenges and approaches},
  author={Toffetti, Giovanni and Militano, Leonardo and Tharaka, Ratnayake and Straub, Mark},
  booktitle={Proc. {IEEE/ACM} 16th Int. Conf. on Utility and Cloud Computing},
  pages={1--6},
  year={2023}
}

@article{portugal2025inquiring,
  title={Inquiring the {Robot Operating System} community on the state of adoption of the {ROS} 2 robotics middleware},
  author={Portugal, David and Rocha, Rui P and Castilho, Jo{\~a}o P},
  journal={Int. J. of Intelligent Robotics and Applications},
  volume={9},
  number={2},
  pages={454--479},
  year={2025},
  publisher={Springer}
}

@misc{rti2014datacentric_misc,
  author       = {{Real-Time Innovations}},
  title        = {Data-centric middleware: Modular software infrastructure to monitor, control and collect real-time equipment analytics},
  howpublished = {Whitepaper. [Online]. Available: \url{https://www.rti.com/hubfs/docs/RTI_Data_Centric_Middleware.pdf}},
  year         = {2014},
  note         = {Accessed: 2026-04-02}
}

@misc{eprosima_discovery_server,
  author       = {{eProsima}},
  title        = {{ROS} 2 discovery server --- {Fast DDS} documentation},
  howpublished = {[Online]. Available: \url{https://fast-dds.docs.eprosima.com/en/latest/fastdds/ros2/discovery_server/ros2_discovery_server.html}},
  year         = {2024},
  note         = {Accessed: 2026-04-03}
}

@misc{rti_routing_service,
  author       = {{Real-Time Innovations (RTI)}},
  title        = {{RTI} routing service --- {RTI Connext DDS Professional} 7.3.0 documentation},
  howpublished = {[Online]. Available: \url{https://community.rti.com/static/documentation/connext-dds/7.3.0/doc/manuals/connext_dds_professional/services/routing_service/index.html}},
  year         = {2024},
  note         = {Accessed: 2026-04-03}
}

@inproceedings{hoffert2007qos,
  title={A {QoS} policy configuration modeling language for publish/subscribe middleware platforms},
  author={Hoffert, Joe and Schmidt, Douglas and Gokhale, Aniruddha},
  booktitle={Proc. 2007 Inaugural Int. Conf. on Distributed Event-Based Systems},
  pages={140--145},
  year={2007}
}

@article{an2013model,
  title={Model-driven generative framework for automated {OMG} {DDS} performance testing in the cloud},
  author={An, Kyoungho and Kuroda, Takayuki and Gokhale, Aniruddha and Tambe, Sumant and Sorbini, Andrea},
  journal={{ACM} SIGPLAN Notices},
  volume={49},
  number={3},
  pages={179--182},
  year={2013},
  publisher={ACM}
}

@article{jalil2023qos,
  title={{QoS} balancing optimization in aggregated robot processing architecture: Rate and buffer},
  author={Jalil, Abdul and Kobayashi, Jun and Saitoh, Takeshi},
  journal={J. of Advances in Artificial Life Robotics},
  volume={3},
  number={4},
  pages={209--213},
  year={2023},
  publisher={ALife Robotics Corporation Ltd}
}

@inproceedings{parra2021specifying,
  title={Specifying {QoS} requirements and capabilities for component-based robot software},
  author={Parra, Samuel and Schneider, Sven and Hochgeschwender, Nico},
  booktitle={Proc. 2021 {IEEE/ACM} 3rd Int. Workshop on Robotics Software Engineering ({RoSE})},
  pages={29--36},
  year={2021},
  publisher={IEEE}
}

@article{kim2025dynamic,
  title={A dynamic bridge architecture for efficient interoperability between {AUTOSAR} adaptive and {ROS} 2},
  author={Kim, Suhong and Choi, Hyeongju and Lee, Suhaeng and Kim, Minseo and Shin, Hyunseo and Moon, Changjoo},
  journal={Electronics},
  volume={14},
  number={18},
  pages={3635},
  year={2025},
  publisher={MDPI}
}

@article{sim2021data,
  title={{Data Distribution Service} converter based on the {OPC UA} publish--subscribe protocol},
  author={Sim, Woongbin and Song, ByungKwen and Shin, Junho and Kim, Taehun},
  journal={Electronics},
  volume={10},
  number={20},
  pages={2524},
  year={2021},
  publisher={MDPI}
}

@misc{ros2_domain_bridge_design,
  author       = {{Open Robotics}},
  title        = {{ROS} 2 domain bridge design and implementation},
  year         = {2024},
  howpublished = {[Online]. Available: \url{https://docs.ros.org/en/kilted/p/domain_bridge/doc/design.html}},
  note         = {Accessed: Apr. 6, 2026}
}

@article{yoshino2025high,
  title={High communication performance {Internet of Robotic Things} systems based on {RT-Middleware}},
  author={Yoshino, Daishi and Watanobe, Yutaka and Naruse, Keitaro},
  journal={{IEEE} Access},
  volume={13},
  pages={207527--207540},
  year={2025},
  publisher={IEEE}
}

@article{aragao2016middleware,
  title={Middleware interoperability for robotics: A {ROS}--{YARP} framework},
  author={Arag{\~a}o, Miguel and Moreno, Plinio and Bernardino, Alexandre},
  journal={Frontiers in Robotics and {AI}},
  volume={3},
  pages={64},
  year={2016},
  publisher={Frontiers Media SA}
}

@article{song2024ros,
  title={{ROS} gateway: Enhancing {ROS} availability across multiple network environments},
  author={Song, Byoung-Youl and Choi, Hoon},
  journal={Sensors},
  volume={24},
  number={19},
  pages={6297},
  year={2024},
  publisher={MDPI}
}

@misc{zenoh_bridge_dds_doc,
  author       = {{Eclipse Zenoh}},
  title        = {{Zenoh} bridge for {DDS}},
  year         = {2024},
  howpublished = {[Online]. Available: \url{https://docs.ros.org/en/kilted/p/zenoh_bridge_dds/}},
  note         = {Accessed: Apr. 6, 2026}
}

@inproceedings{moon2020gazebo,
  title={A {Gazebo}/{ROS}-based communication-realistic simulator for networked {SUAS}},
  author={Moon, Sangwoo and Bird, John J and Borenstein, Steve and Frew, Eric W},
  booktitle={Proc. 2020 Int. Conf. on Unmanned Aircraft Systems ({ICUAS})},
  pages={1819--1827},
  year={2020},
  publisher={IEEE}
}

@article{shi2023adapting,
  title={Adapting wireless network configuration from simulation to reality via deep learning-based domain adaptation},
  author={Shi, Junyang and Ma, Aitian and Cheng, Xia and Sha, Mo and Xi, Peng},
  journal={{IEEE/ACM} Trans. on Networking},
  volume={32},
  number={3},
  pages={1983--1998},
  year={2023},
  publisher={IEEE}
}

@inproceedings{richart2022cocosim,
  title={{CoCoSim}: A tool for co-simulation of mobile cooperative robots},
  author={Richart, Mat{\'\i}as and Vel{\'a}zquez, Felipe and Ciuffardi, Federico and Visca, Jorge and Baliosian, Javier},
  booktitle={Proc. Int. Conf. on Software Engineering and Formal Methods},
  pages={258--268},
  year={2022},
  publisher={Springer}
}

@inproceedings{acharya2020cornet,
  title={{CorNet}: A co-simulation middleware for robot networks},
  author={Acharya, Srikrishna and Bharadwaj, Amrutur and Simmhan, Yogesh and Gopalan, Aditya and Parag, Parimal and Tyagi, Himanshu},
  booktitle={Proc. 2020 Int. Conf. on {COMmunication} Systems \& {NETworkS} ({COMSNETS})},
  pages={245--251},
  year={2020},
  publisher={IEEE}
}

@article{calvo2021ros,
  title={{ROS-NetSim}: A framework for the integration of robotic and network simulators},
  author={Calvo-Fullana, Miguel and Mox, Daniel and Pyattaev, Alexander and Fink, Jonathan and Kumar, Vijay and Ribeiro, Alejandro},
  journal={{IEEE} Robotics and Automation Letters},
  volume={6},
  number={2},
  pages={1120--1127},
  year={2021},
  publisher={IEEE}
}

@article{agarwal2025sim,
  title={Sim-to-real transfer for estimation over wireless networks},
  author={Agarwal, Shivangi and Asija, Adi and Kaul, Sanjit K and Anand, Saket},
  journal={{ACM} Trans. on Cyber-Physical Systems},
  volume={9},
  number={4},
  pages={1--25},
  year={2025},
  publisher={ACM}
}

@article{zhao2026distributed,
  title={Distributed state estimation for discrete-time {LTI} systems: The design trilemma and a novel framework},
  author={Zhao, Ruixuan and Yang, Guitao and Fleming, James and Chen, Boli},
  journal={arXiv preprint arXiv:2603.20144},
  year={2026}
}

@inproceedings{hoffert2008dqml,
  title={{DQML}: A modeling language for configuring distributed publish/subscribe {Quality of Service} policies},
  author={Hoffert, Joe and Schmidt, Douglas and Gokhale, Aniruddha},
  booktitle={Proc. {OTM} Confederated Int. Conf. on the Move to Meaningful Internet Systems},
  pages={515--534},
  year={2008},
  publisher={Springer}
}

@article{sciangula2025end,
  title={End-to-end response-time analysis of {DDS}-based real-time applications},
  author={Sciangula, Gerlando and Casini, Daniel and Biondi, Alessandro and Scordino, Claudio and Di Natale, Marco},
  journal={Internet of Things},
  pages={101853},
  year={2025},
  publisher={Elsevier}
}

@article{almadani2017qos,
  title={{QoS} adaptation for publish/subscribe middleware in real-time dynamic environments},
  author={Almadani, Basem and Abudalfa, Shadi and others},
  journal={Int. Arab J. of Information Technology ({IAJIT})},
  volume={14},
  number={2},
  year={2017}
}

@inproceedings{hakiri2013supportinginternet,
  title={Supporting end-to-end internet {QoS} for {DDS}-based large-scale distributed simulation},
  author={Hakiri, Akram and Berthou, Pascal and Slim Abdellatif, Slim and Diaz, Michel and Gayraud, Thierry},
  booktitle={Proc. 1st {ACM} {SIGSIM} Conf. on Principles of Advanced Discrete Simulation},
  pages={397--402},
  year={2013}
}

@article{hoffert2011timely,
  title={Timely autonomic adaptation of publish/subscribe middleware in dynamic environments},
  author={Hoffert, Joe and Gokhale, Aniruddha and Schmidt, Douglas C},
  journal={Int. J. of Adaptive, Resilient and Autonomic Systems ({IJARAS})},
  volume={2},
  number={4},
  pages={1--24},
  year={2011},
  publisher={IGI Global}
}

@inproceedings{bosk2024towards,
  title={Towards domain-specific time-sensitive information-centric networking architecture},
  author={Bosk, Marcin and Ott, J{\"o}rg},
  booktitle={Proc. 2024 {IFIP} Networking Conf. ({IFIP} Networking)},
  pages={720--725},
  year={2024},
  publisher={IEEE}
}

@inproceedings{balasubramanian2009adaptive,
  title={Adaptive failover for real-time middleware with passive replication},
  author={Balasubramanian, Jaiganesh and Tambe, Sumant and Lu, Chenyang and Gokhale, Aniruddha and Gill, Christopher and Schmidt, Douglas C},
  booktitle={Proc. 2009 15th {IEEE} Real-Time and Embedded Technology and Applications Symp.},
  pages={118--127},
  year={2009},
  publisher={IEEE}
}

@inproceedings{orsini2016cloudaware,
  title={{CloudAware}: A context-adaptive middleware for mobile edge and cloud computing applications},
  author={Orsini, Gabriel and Bade, Dirk and Lamersdorf, Winfried},
  booktitle={Proc. 2016 {IEEE} 1st Int. Workshops on Foundations and Applications of Self* Systems ({FAS*W})},
  pages={216--221},
  year={2016},
  publisher={IEEE}
}

@inproceedings{bellavista2008context,
  title={Context-aware middleware for reliable multi-hop multi-path connectivity},
  author={Bellavista, Paolo and Corradi, Antonio and Giannelli, Carlo},
  booktitle={Proc. {IFIP} Int. Workshop on Software Technologies for Embedded and Ubiquitous Systems},
  pages={66--78},
  year={2008},
  publisher={Springer}
}

@article{pitoura2002data,
  title={Data consistency in intermittently connected distributed systems},
  author={Pitoura, Evaggelia and Bhargava, Bharat},
  journal={{IEEE} Trans. on Knowledge and Data Engineering},
  volume={11},
  number={6},
  pages={896--915},
  year={2002},
  publisher={IEEE}
}

@article{szabo2025enhancing,
  title={Enhancing vertical application and network co-design: A solution for multipath channel switching in {ROS} 2 with {3GPP} integration},
  author={Szab{\'o}, G{\'e}za},
  journal={{IEEE} Access},
  year={2025},
  publisher={IEEE}
}

@inproceedings{garcia2009topology,
  title={Topology-aware group communication middleware for {MANET}s},
  author={Garcia Lopez, Pedro and Gracia, Ra{\'u}l and Espelt, Marc and Paris, Gerard and Arrufat, Marcel and Messeguer, Roc},
  booktitle={Proc. 4th Int. {ICST} Conf. on {COMmunication} System {softWAre} and {middlewaRE}},
  pages={1--10},
  year={2009}
}

@inproceedings{balasubramanian2008towards,
  title={Towards middleware for fault-tolerance in distributed real-time and embedded systems},
  author={Balasubramanian, Jaiganesh and Gokhale, Aniruddha and Schmidt, Douglas C and Wang, Nanbor},
  booktitle={Proc. {IFIP} Int. Conf. on Distributed Applications and Interoperable Systems},
  pages={72--85},
  year={2008},
  publisher={Springer}
}

@article{david2013dds,
  title={A {DDS}-based middleware for scalable tracking, communication and collaboration of mobile nodes},
  author={David, Lincoln and Vasconcelos, Rafael and Alves, Lucas and Andr{\'e}, Rafael and Endler, Markus},
  journal={J. of Internet Services and Applications},
  volume={4},
  number={1},
  pages={16},
  year={2013},
  publisher={Springer}
}

@inproceedings{gomes2015federated,
  title={A federated discovery service for the {Internet of Things}},
  author={Gomes, Porfirio and Cavalcante, Everton and Rodrigues, Taniro and Batista, Thais and Delicato, Flavia C and Pires, Paulo F},
  booktitle={Proc. 2nd Workshop on Middleware for Context-Aware Applications in the {IoT}},
  pages={25--30},
  year={2015}
}

@article{ebadi2011new,
  title={A new distributed and hierarchical mechanism for service discovery in a grid environment},
  author={Ebadi, Saeed and Khanli, Leyli Mohammad},
  journal={Future Generation Computer Systems},
  volume={27},
  number={6},
  pages={836--842},
  year={2011},
  publisher={Elsevier}
}

@inproceedings{harbird2004adaptive,
  title={Adaptive resource discovery for ubiquitous computing},
  author={Harbird, Rae and Hailes, Stephen and Mascolo, Cecilia},
  booktitle={Proc. 2nd Workshop on Middleware for Pervasive and Ad-Hoc Computing},
  pages={155--160},
  year={2004}
}

@article{kurte2024decentralised,
  title={Decentralised global service discovery for the {Internet of Things}},
  author={Kurte, Ryan and Salcic, Zoran and Wang, Kevin I-Kai},
  journal={Sensors},
  volume={24},
  number={7},
  pages={2196},
  year={2024},
  publisher={MDPI}
}

@inproceedings{tang2019design,
  title={Design of high availability service discovery for microservices architecture},
  author={Tang, Weilun and Wang, Li and Xue, Guangtao},
  booktitle={Proc. 2019 3rd Int. Conf. on Management Engineering, Software Engineering and Service Sciences},
  pages={253--257},
  year={2019}
}

@article{jung2008efficient,
  title={Efficient service discovery mechanism for wireless sensor networks},
  author={Jung, Jaehoon and Lee, Seunghak and Kim, Namgi and Yoon, Hyunsoo},
  journal={Computer Communications},
  volume={31},
  number={14},
  pages={3292--3298},
  year={2008},
  publisher={Elsevier}
}

@inproceedings{song2020integration,
  title={Integration of {Data Distribution Service} into partitioned real-time embedded systems},
  author={Song, Boyang and Hu, Xiaoguang and Xiao, Jin and Zhang, Guofeng and Wang, Shuo and Zhou, Qing},
  booktitle={Proc. 2020 15th {IEEE} Conf. on Industrial Electronics and Applications ({ICIEA})},
  pages={1606--1611},
  year={2020},
  publisher={IEEE}
}

@article{boari2008middleware,
  title={Middleware for automatic dynamic reconfiguration of context-driven services},
  author={Boari, Maurelio and Lodolo, Enrico and Monti, Stefano and Pasini, Samuele},
  journal={Microprocessors and Microsystems},
  volume={32},
  number={3},
  pages={145--158},
  year={2008},
  publisher={Elsevier}
}

@inproceedings{mokadem2010resource,
  title={Resource discovery service while minimizing maintenance overhead in hierarchical {DHT} systems},
  author={Mokadem, Riad and Hameurlain, Abdelkader and Tjoa, A Min},
  booktitle={Proc. 12th Int. Conf. on Information Integration and Web-based Applications \& Services},
  pages={630--638},
  year={2010}
}

@inproceedings{detzner2023sola,
  title={{SOLA}: A decentralized communication middleware developed with {ns-3}},
  author={Detzner, Peter and G{\"o}deke, Jana and T{\"o}nning, Lars and Laskowski, Patrick and H{\"o}rstrup, Maximilian and Stolz, Oliver and Brehler, Marius and Kerner, S{\"o}ren},
  booktitle={Proc. 2023 Workshop on {ns-3}},
  pages={78--85},
  year={2023}
}

@inproceedings{wang2024formal,
  title={A formal analysis of {Data Distribution Service} security},
  author={Wang, Binghan and Li, Hui and Guan, Jingjing},
  booktitle={Proc. 19th {ACM} Asia Conf. on Computer and Communications Security},
  pages={716--727},
  year={2024}
}

@inproceedings{patel2022analyzing,
  title={Analyzing security vulnerability and forensic investigation of {ROS} 2: A case study},
  author={Patel, Yash and Rughani, Parag H and Desai, Dhruvil},
  booktitle={Proc. 8th Int. Conf. on Robotics and Artificial Intelligence},
  pages={6--12},
  year={2022}
}

@inproceedings{aartsen2022analyzing,
  title={Analyzing interoperability and security overhead of {ROS} 2 {DDS} middleware},
  author={Aartsen, Max and Banga, Kanta and Talko, Konrad and Touw, Dustin and Wisman, Bertus and Me{\"\i}nsma, Daniel and Bj{\"o}rkqvist, Mathias},
  booktitle={Proc. 2022 30th Mediterranean Conf. on Control and Automation ({MED})},
  pages={976--981},
  year={2022},
  publisher={IEEE}
}

@inproceedings{sandoval2019cyber,
  title={Cyber security assessment of the {Robot Operating System} 2 for aerial networks},
  author={Sandoval, Sergio and Thulasiraman, Preetha},
  booktitle={Proc. 2019 {IEEE} Int. Systems Conf. ({SysCon})},
  pages={1--8},
  year={2019},
  publisher={IEEE}
}

@inproceedings{wagner2024dds,
  title={{DDS} {Security+}: Enhancing the {Data Distribution Service} with {TPM}-based remote attestation},
  author={Wagner, Paul Georg and Birnstill, Pascal and Beyerer, J{\"u}rgen},
  booktitle={Proc. 19th Int. Conf. on Availability, Reliability and Security},
  pages={1--11},
  year={2024}
}

@article{du2022formal,
  title={Formal safety assessment and improvement of {DDS} protocol for industrial {Data Distribution Service}},
  author={Du, Jinze and Gao, Chengtai and Feng, Tao},
  journal={Future Internet},
  volume={15},
  number={1},
  pages={24},
  year={2022},
  publisher={MDPI}
}

@inproceedings{xia2025investigating,
  title={Investigating security threats in multi-tenant {ROS} 2 systems},
  author={Xia, Lichen and Gao, Xing and Shi, Weisong},
  booktitle={Proc. 2025 {IEEE} Int. Conf. on Robotics and Automation ({ICRA})},
  pages={16441--16448},
  year={2025},
  publisher={IEEE}
}

@inproceedings{deng2022security,
  title={On the (in)security of secure {ROS} 2},
  author={Deng, Gelei and Xu, Guowen and Zhou, Yuan and Zhang, Tianwei and Liu, Yang},
  booktitle={Proc. 2022 {ACM} {SIGSAC} Conf. on Computer and Communications Security},
  pages={739--753},
  year={2022}
}

@inproceedings{takemoto2019performance,
  title={Performance evaluation of {CAESAR} authenticated encryption on {SROS} 2},
  author={Takemoto, Shu and Nishida, Kanata and Nozaki, Yusuke and Yoshikawa, Masaya and Honda, Shinya and Kurachi, Ryo},
  booktitle={Proc. 2019 2nd Artificial Intelligence and Cloud Computing Conf.},
  pages={168--172},
  year={2019}
}

@inproceedings{white2018procedurally,
  title={Procedurally provisioned access control for robotic systems},
  author={White, Ruffin and Christensen, Henrik I and Caiazza, Gianluca and Cortesi, Agostino},
  booktitle={Proc. 2018 {IEEE/RSJ} Int. Conf. on Intelligent Robots and Systems ({IROS})},
  pages={1--9},
  year={2018},
  publisher={IEEE}
}

@article{kim2018security,
  title={Security and performance considerations in {ROS} 2: A balancing act},
  author={Kim, Jongkil and Smereka, Jonathon M and Cheung, Calvin and Nepal, Surya and Grobler, Marthie},
  journal={arXiv preprint arXiv:1809.09566},
  year={2018}
}

@inproceedings{wang2024support,
  title={Support remote attestation for decentralized {Robot Operating System} ({ROS}) using trusted execution environment},
  author={Wang, Qian and Lee, Brian and Qiao, Yuansong},
  booktitle={Proc. 2024 {IEEE} Int. Conf. on Blockchain and Cryptocurrency ({ICBC})},
  pages={693--695},
  year={2024},
  publisher={IEEE}
}

@article{basem2017data,
  title={{Data Distribution Service} ({DDS}) based implementation of smart grid devices using {ANSI} {C12.19} standard},
  author={Basem, AL-Madani and Ali, Hassan},
  journal={Procedia Computer Science},
  volume={110},
  pages={394--401},
  year={2017},
  publisher={Elsevier}
}

@article{almadani2015performance,
  title={Performance evaluation of {DDS}-based middleware over wireless channel for reconfigurable manufacturing systems},
  author={Almadani, Basem and Bajwa, Muhammad Naseer and Yang, Shuang-Hua and Saif, Abdul-Wahid A},
  journal={Int. J. of Distributed Sensor Networks},
  volume={11},
  number={7},
  pages={863123},
  year={2015},
  publisher={SAGE Publications}
}

@inproceedings{casesbenchmarking,
  title={Benchmarking publish/subscribe middleware for radar applications},
  author={Rhoades, Andrew S and Schrader, Glenn and Poulin, Paul},
  booktitle={Proc. Eleventh Annual {HPEC} Workshop},
  year={2007}
}

@inproceedings{lee2014router,
  title={Router design for {DDS}: Architecture and performance evaluation},
  author={Lee, Kyu-haeng and Kim, Chong-kwon and Kim, Kyeong Tae and Kim, Won-tae},
  booktitle={Proc. 2014 Int. Conf. on Big Data and Smart Computing ({BIGCOMP})},
  pages={250--254},
  year={2014},
  publisher={IEEE}
}

@inproceedings{youssef2015dds,
  title={{DDS}-based interoperability framework for smart grid testbed infrastructure},
  author={Youssef, Tarek A and Elsayed, Ahmed T and Mohammed, Osama A},
  booktitle={Proc. 2015 {IEEE} 15th Int. Conf. on Environment and Electrical Engineering ({EEEIC})},
  pages={219--224},
  year={2015},
  publisher={IEEE}
}

@article{youssef2017dds,
  title={A {DDS}-based energy management framework for small microgrid operation and control},
  author={Youssef, Tarek A and El Hariri, Mohamad and Elsayed, Ahmed T and Mohammed, Osama A},
  journal={{IEEE} Trans. on Industrial Informatics},
  volume={14},
  number={3},
  pages={958--968},
  year={2017},
  publisher={IEEE}
}

@online{zenoh_longwang_2026,
  title={Zenoh Longwang Release},
  organization={Zenoh},
  url={https://zenoh.io/blog/2026-04-16-zenoh-longwang/},
  urldate={2026-05-27}
}

@article{lee2026harness,
  title={Harness engineering for {Physical AI}: Robot middleware is the harness layer},
  author={Lee, Sanghoon and Chae, Jiyeong and Park, Kyung-Joon},
  journal={arXiv preprint arXiv:2606.09416},
  year={2026}
}

@article{yu2026ros2probe,
  title={{ros2probe}: Non-intrusive, kernel-selective observability for {Robot Operating System} 2 middleware},
  author={Yu, Jisang and Lee, Sanghoon and Choi, Yeonwoo and Park, Kyung-Joon},
  journal={arXiv preprint arXiv:2606.10746},
  year={2026}
}

\end{document}